%% file: main.tex
\RequirePackage{fix-cm}
\documentclass{fairastra}

\usepackage{lmodern}
\usepackage{newtxtext}
\usepackage{microtype}
\usepackage{graphicx}
\usepackage{listings}
\lstdefinestyle{prompt}{basicstyle=\ttfamily\footnotesize,breaklines=true,
  columns=fullflexible,keepspaces=true,showstringspaces=false,
  literate={—}{{---}}1 {–}{{--}}1 {’}{{'}}1 {…}{{...}}1 {≤}{{$\leq$}}1}
\usepackage{booktabs}
\usepackage{multirow}
\usepackage{capt-of}
\usepackage{float}
\usepackage{placeins}
\usepackage{afterpage}
\usepackage{needspace}
\usepackage{array}
\usepackage{amsmath}
\usepackage{amssymb}
\usepackage{bm}
\usepackage{algorithm}
\usepackage{algpseudocode}
\usepackage{xcolor}
\usepackage{colortbl}
\usepackage{longtable}
\usepackage{tikz}
\usetikzlibrary{arrows.meta,positioning,patterns}
\usepackage{tcolorbox}
\tcbuselibrary{breakable,skins,listings}
\usepackage[scaled=.95]{zi4}

\usepackage[sf,scaled=.90]{noto}
\newcommand{\agentfont}{\fontencoding{T1}\fontfamily{NotoSans-TLF}\linespread{1.10}\selectfont}

\usepackage{soul}
\definecolor{hlmem}{RGB}{226,222,247}\definecolor{hlwarn}{RGB}{252,232,205}\definecolor{hlok}{RGB}{214,240,226}
\definecolor{barmem}{RGB}{74,58,167}\definecolor{barwarn}{RGB}{179,86,15}\definecolor{barok}{RGB}{27,140,96}
\definecolor{figblue}{HTML}{3A6EA5}\definecolor{figochre}{HTML}{D08A3E}\definecolor{figteal}{HTML}{4E9A8A}\definecolor{figgrey}{HTML}{8C8C8C}\definecolor{figred}{HTML}{C9694A}
\definecolor{barjudge}{RGB}{42,120,214}\definecolor{insightbg}{RGB}{245,243,252}
\newenvironment{apptab}[1]{\renewcommand{\arraystretch}{1.18}\begin{tabular}{#1}\toprule}{\end{tabular}}
\newcommand{\thd}[1]{\textbf{#1}}

\AtBeginEnvironment{table}{\setlength{\belowcaptionskip}{7pt}\setlength{\abovecaptionskip}{4pt}}
\newcommand{\hlm}[1]{\begingroup\sethlcolor{hlmem}\hl{#1}\endgroup}   
\newcommand{\hlw}[1]{\begingroup\sethlcolor{hlwarn}\hl{#1}\endgroup}  
\newcommand{\hlo}[1]{\begingroup\sethlcolor{hlok}\hl{#1}\endgroup}    
\newtcolorbox{ledger}[4][barmem]{enhanced,breakable,colback=white,colframe=gray!40,boxrule=0.4pt,arc=2pt,
  borderline west={2.2pt}{0pt}{#1},left=7pt,right=6pt,top=6pt,bottom=5pt,before skip=6pt,after skip=7pt,
  title={\small\textbf{#2}\hspace{1.2em plus 1fill}{\footnotesize\normalfont #4}},fonttitle=\rmfamily,coltitle=black,
  colbacktitle=gray!10,toptitle=2pt,bottomtitle=2pt,after title={\par\vspace{2pt}\footnotesize\itshape #3},fontupper=\small\agentfont}
\newtcolorbox{insight}{enhanced,breakable,colback=insightbg,colframe=insightbg,boxrule=0pt,arc=1.5pt,
  borderline west={2.2pt}{0pt}{barmem},left=8pt,right=6pt,top=3pt,bottom=3pt,before skip=2pt,after skip=9pt,fontupper=\small}
\lstdefinestyle{promptbody}{basicstyle=\small\fontfamily{zi4}\selectfont,breaklines=true,breakindent=0pt,
  columns=fullflexible,keepspaces=true,showstringspaces=false,frame=none,
  aboveskip=0pt,belowskip=0pt,xleftmargin=0pt,
  literate={—}{{---}}1 {–}{{--}}1 {’}{{'}}1 {…}{{...}}1 {≤}{{$\leq$}}1 {°}{{$^\circ$}}1}

\hypersetup{colorlinks=true,linkcolor=astracolor,citecolor=astracolor,urlcolor=astracolor}
\usepackage{url}
\DeclareTextFontCommand{\texttt}{\rmfamily}

\newcommand{\tool}[1]{\textrm{#1}}   
\newcommand{\CERT}{\textsc{certified}}
\newcommand{\NOTDONE}{\textsc{not done}}
\newcommand{\UNRES}{\textsc{unresolved}}
\newcommand{\CLAIMED}{\textsc{claimed}}

\definecolor{draftplaceholder}{HTML}{808080}
\definecolor{draftprompted}{HTML}{237A45}
\newif\ifdraftcolors
\draftcolorsfalse 
\ifdraftcolors\else\hypersetup{allcolors=astracolor}\fi
\newcommand{\draftstatus}[1]{\ifdraftcolors\color{#1}\hypersetup{linkcolor=#1,citecolor=#1,urlcolor=#1}\else\color{black}\fi}
\newenvironment{aiplaceholder}{\begingroup\draftstatus{draftplaceholder}}{\par\endgroup}
\newenvironment{humanprompted}{\begingroup\draftstatus{draftprompted}}{\par\endgroup}
\newenvironment{humanconfirmed}{\begingroup\draftstatus{black}}{\par\endgroup}

\title{NavHarness: Towards Lifelong Embodied Navigation}

\author{\parbox{\linewidth}{\raggedright
Xunyi Zhao\textsuperscript{1,2,*}\quad
Jian Zhou\textsuperscript{1,*}\quad
Sihao Lin\textsuperscript{1,2}\quad
Gengze Zhou\textsuperscript{1}\quad
Zerui Li\textsuperscript{1}\\[1pt]
Xinyu Yan\textsuperscript{1,2}\quad
Jiajun Liu\textsuperscript{2,3}\quad
Anton van den Hengel\textsuperscript{1,2}\quad
Qi Wu\textsuperscript{1,2,\ensuremath{\dagger}}
}}
\renewcommand{\affiliationlist}{{\small\sffamily\bfseries
\mbox{\textsuperscript{1}Adelaide University}\quad
\mbox{\textsuperscript{2}Responsible AI Research Centre, AIML}\quad
\mbox{\textsuperscript{3}CSIRO Data61}\par}}
\contribution[*]{Equal contribution}
\contribution[\dagger]{Corresponding author}
\abstract{\input{sections/abstract}}
\date{September 28, 2026}
\astradata[Project Page]{\url{https://billzhao1030.github.io}}
\astradata[GitHub]{\url{https://github.com/billzhao1030/NavHarness}}
\hypersetup{pdftitle={NavHarness: Towards Lifelong Embodied Navigation},
  pdfauthor={Xunyi Zhao, Jian Zhou, Sihao Lin, Gengze Zhou, Zerui Li, Xinyu Yan, Jiajun Liu, Anton van den Hengel, Qi Wu}}

\begin{document}
\noindent\makebox[\linewidth][s]{%
\makebox[0.31\linewidth][c]{\raisebox{-0.5\height}{\includegraphics[height=1.55cm,viewport=36 359 1882 1081,clip]{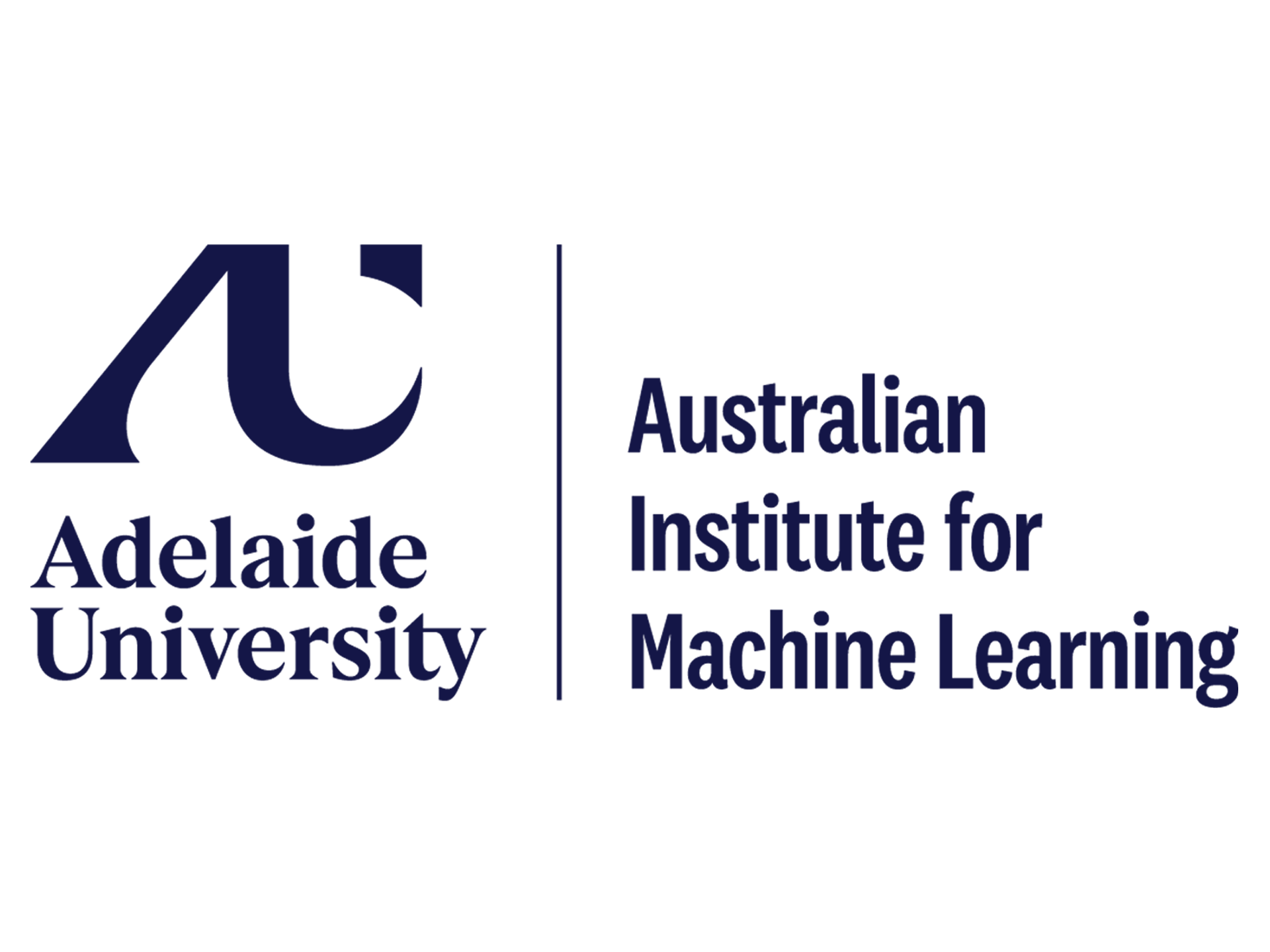}}}\hfill
\makebox[0.31\linewidth][c]{\raisebox{-0.5\height}{\includegraphics[height=0.95cm,viewport=104 520 1816 921,clip]{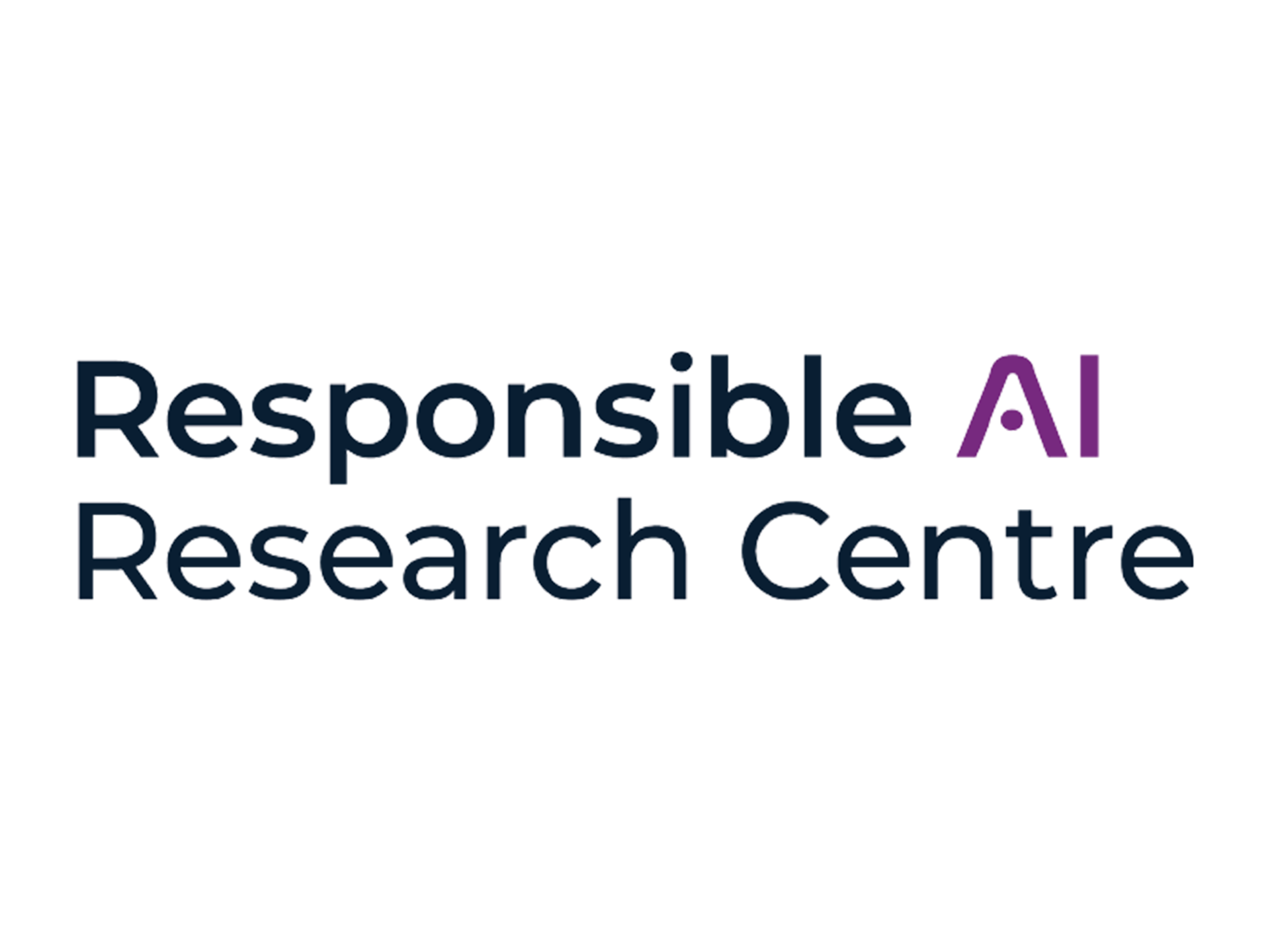}}}\hfill
\makebox[0.31\linewidth][c]{\raisebox{-0.5\height}{\includegraphics[height=1.55cm]{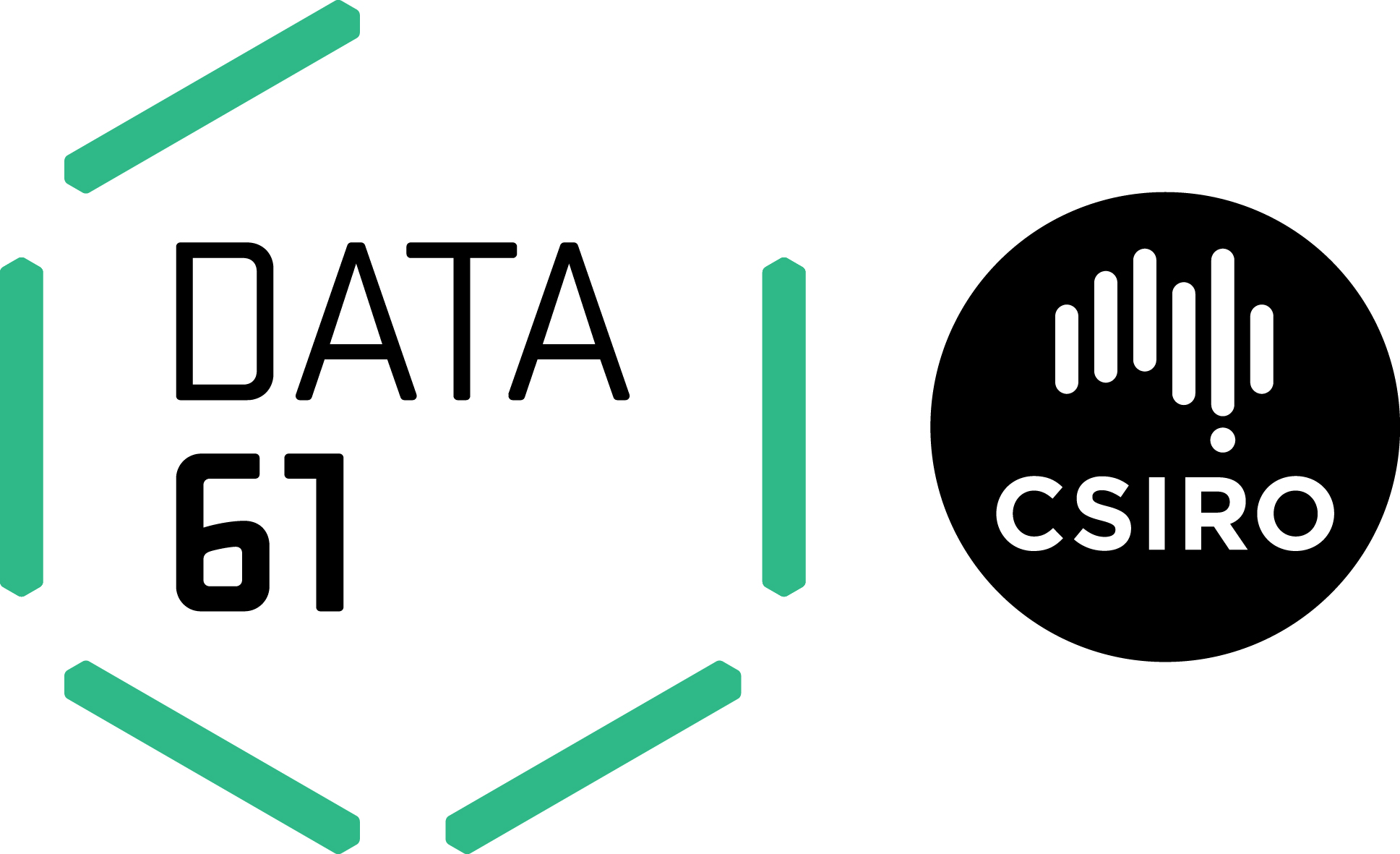}}}%
}\par\vspace{10pt}
\maketitle

\afterpage{%
\begin{figure}[!t]
\draftstatus{draftprompted}
\centering
\includegraphics[width=\linewidth]{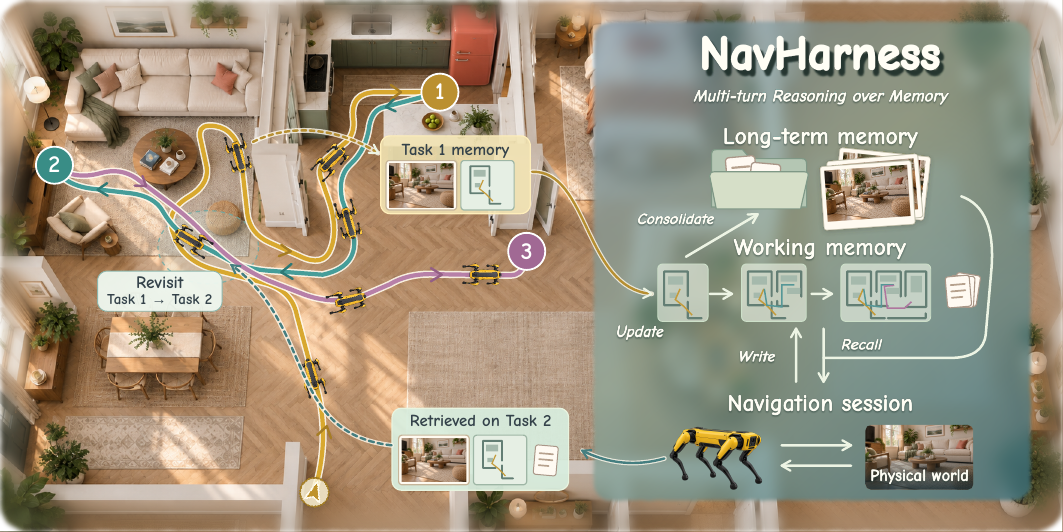}
\caption{\textbf{NavHarness} orchestrates long-horizon embodied navigation
through multi-round multimodal reasoning over working memory and house
knowledge, retaining experience for successive tasks.}
\label{fig:main}
\end{figure}
}
\begin{humanprompted}
\input{sections/intro}
\input{sections/related}
\end{humanprompted}
\begin{humanconfirmed}
\input{sections/harness}
\end{humanconfirmed}
\begin{aiplaceholder}
\input{sections/experiments}
\input{sections/discussion}
\end{aiplaceholder}

\clearpage
\bibliography{refs}

\clearpage
\appendix
\raggedbottom
\begin{center}
{\Large\bfseries Appendix}
\end{center}
\noindent
The appendix complements the main paper with implementation details and
a closer examination of the experimental results.
Appendix~\ref{app:protocol} presents the full algorithm, six-component harness
specification, and transition rules. Appendix~\ref{app:benchmarks} describes the evaluation
settings, statistical procedures, and supplementary controls, while
Appendix~\ref{app:results} reports component results and diagnostics.
Appendix~\ref{app:analysis} examines extended deployments and case studies of
memory use, and Appendix~\ref{app:limitations} discusses limitations and
future directions.
\vspace{6pt}
\section{Harness Specification}
\label{app:protocol}
\input{sections/app_algorithm}
\input{sections/app_protocol}

\section{Benchmarks and Evaluation Protocol}
\label{app:benchmarks}
\input{sections/app_benchmarks}

\input{sections/app_results}
\input{sections/app_analysis}
\input{sections/app_limitations}

\end{document}

%% file: sections/abstract.tex
Frontier models can now perform well on individual embodied navigation tasks through
multi-round multimodal reasoning with simple tools. Across successive
tasks, however, an agent must also rely on an evolving map and earlier
search records, both of which may be incomplete or conflict with new
observations. We present \textbf{NavHarness}, a training-free embodied
harness towards lifelong navigation that makes memory processing part
of the navigation loop. During navigation, its multi-round agentic session
draws on maps, task records, and house knowledge, checking them against
observations and recording corrections to guide its actions. NavHarness
preserves this experience across fresh conversations for new tasks or recovery
attempts, while outcome verification and run-end summaries support its later reuse.
On GOAT-Bench, NavHarness improves
s-SR over context-only independent sessions by 18.6 points with Astra
and 22.6 with Opus 5. Using SLAM-estimated poses, NavHarness with GPT-6 Astra
achieves state-of-the-art task success of 83.7 s-SR with 36.9 e-SR on
GOAT-Bench and 85.9 s-SR on IR2R-CE. To understand these gains, we examine
how experience is carried between sessions and find that structured recovery
handovers outperform length-matched summaries. In extended deployments across houses,
consolidation improves navigation beyond retaining maps and task records,
with case studies showing how agents use earlier experience to interpret
new goals, investigate unresolved questions, and resume failed searches.
We suggest that progress towards lifelong navigation depends on how
successive reasoning sessions build on prior experience, alongside
improvements in single-task capability.

%% file: sections/intro.tex
\section{Introduction}
\label{sec:intro}

Consider a household robot navigating a home throughout the day, moving
from one request to the next, revisiting familiar rooms, and finding its
way through places it has yet to explore. Advances in multi-round
multimodal reasoning bring this vision closer, enabling frontier models
to navigate effectively within a single task using a basic coding harness
and simple observation and movement tools, without
navigation-specific training \citep{mip}. For a robot that keeps working,
each journey also reveals more of the home and leaves experience that
could help with a later request. Extending this capability towards lifelong
navigation requires the robot to draw on that experience when deciding
where to go and what to inspect next.

In continuous household navigation, each task starts where the previous
one ended, so a new request arrives in a home the robot has already begun
to explore. Restarting the conversation leaves the robot where it is,
so the next reasoning session needs the map and search evidence accumulated
so far. Yet a record that an object was not found in a room does not
establish that every relevant part was inspected, and a stored description
may conflict with the current view. To choose where to search next, the
navigator must interpret these records against current observations and
revise them when the evidence changes.

Long-horizon navigation has made substantial progress toward this goal
by retaining scene knowledge across tasks \citep{overnav}, retrieving
past observations for exploration \citep{threedmem,ssmgnav,gsmem}, and
learning policies that query accumulated experience \citep{lmee}.
Building on this progress, we study how a general-purpose reasoning
session can use these different sources together as it navigates.
For example, a house note may suggest a room and the map a route to it,
while a failed-search record points to an overlooked corner. On reaching
the room, the robot can compare these leads with what it sees to decide
whether to continue searching there, consult another record, or correct
the note. This makes memory retrieval and revision part of the ongoing
interaction with the environment.

Multi-round agentic sessions offer a practical starting point because
memory and coding agents already use tools to inspect external records
and pursue further queries based on what they find
\citep{memgpt,contextengineering,longrunningharness}. In navigation, this
interaction lets a session read a search record, compare it with the
current view, and consult the map before choosing its next action,
without training a separate memory policy. Keeping these records outside
the conversation also allows a fresh session to inherit evidence about
inspected places, supported exclusions, and unresolved searches. Since
later searches depend on these records, outcome checks help prevent
incorrect completion claims from becoming accepted experience.

We introduce \textbf{NavHarness}, an embodied harness towards lifelong
navigation centered on memory processing without additional model
training. Each multi-round agentic session reasons over current
observations, spatial and task memory, and longer-term house knowledge,
while an outer loop retains these memories as new tasks or recovery
attempts start fresh conversations. The loop also checks task outcomes
and consolidates the resulting records into house notes for later runs.
On GOAT-Bench, NavHarness improves s-SR by 18.6 points with Astra and
22.6 with Opus 5 compared with the same models using a fresh,
context-only session for each task, without external memory
(Section~\ref{sec:exp:gap}). Using SLAM-estimated poses rather than simulator
ground-truth poses, NavHarness with GPT-6 Astra reaches 83.7\% s-SR and
36.9\% e-SR on GOAT-Bench and 85.9\% s-SR on IR2R-CE, exceeding previous
state-of-the-art results. Extended deployments show how retained experience
redirects search, prompts observations to resolve uncertainty, and lets
failed attempts inform later decisions.

\Needspace{5\baselineskip}
To summarize, our contributions are as follows:
\nopagebreak[4]
\begin{enumerate}
\item We show that how experience is used matters beyond retaining it.
Structured handovers outperform length-matched summaries.
NavHarness brings this evidence transfer
into a navigation loop where fresh reasoning sessions continue from the
robot's current state using evidence from earlier searches.
\item We improve all four backbones and achieve state-of-the-art GOAT-Bench
and IR2R-CE performance using SLAM-estimated poses rather than simulator
ground-truth poses, with gains in task success and path efficiency over
same-backbone independent sessions.
\clubpenalty=0
\makeatletter\@clubpenalty=0\makeatother
\item Continuous deployment shows that consolidation improves navigation
beyond retaining maps and task records. Case studies reveal how agents
reuse failed searches and revise earlier accounts, informing the design
of future long-horizon embodied agents.
\end{enumerate}

%% file: sections/related.tex
\section{Related Work}
\label{sec:related}

\paragraph{Long-horizon and lifelong navigation.}
IR2R-CE and GOAT-Bench evaluate successive goals in shared environments,
where accumulated experience can improve later navigation
\citep{ivln,goatbench}. OVER-NAV, 3D-Mem, SSMG-Nav, and GSMem retain spatial knowledge
through structured objects, visual snapshots, semantic graphs, and
renderable memories \citep{overnav,threedmem,ssmgnav,gsmem}, helping agents
connect past observations to places they can revisit. Beyond scene
representations, SeqWalker uses hierarchical planning and
trajectory correction \citep{seqwalker}, MemoryExplorer trains active
retrieval with single-round tool invocation \citep{lmee}, and Uni-Walker
and AllDayNav pursue continued policy learning \citep{uniwalker,alldaynav}.
We share this goal, focusing on how general-purpose multi-round sessions
consult and revise spatial memory, task records, and house knowledge
without navigation-specific training.

\paragraph{Embodied harnesses and agentic control.}
MIP shows that multi-round multimodal reasoning can control navigation
through observation and movement tools in a basic coding harness
\citep{mip}. Navigation interfaces and spatial tools extend this approach
\citep{agenticnav,spacevln,navmcp}, while embodied harnesses connect
reasoning to robot controllers and reusable skills
\citep{harnessvla,showharness,embodiedharness,rho}.
Persistent experience also supports skill and harness evolution
\citep{voyager,zetta,shaper}. HarnessVLN uses event memory and a
spatiotemporal graph to validate navigation proposals
\citep{harnessvln}. RoboHarness uses execution memories
to select heterogeneous policies and prepare the physical state for
policy handoffs \citep{roboharness}.
NavHarness instead studies what must cross a reasoning-session boundary.
A fresh conversation starts from the robot's current physical state and
needs to know which places were inspected, what those observations exclude,
and where search remains incomplete. We test how retrieving, handing over,
and consolidating this evidence supports later navigation.

\paragraph{Memory processing in long-horizon agents.}
External memory lets agents retain experience beyond a context window
without encoding each new experience in model parameters. Generative
Agents and A-MEM turn observations into retrievable, revisable records
\citep{generativeagents,amem}, while Reflexion and ACE preserve lessons
from earlier attempts \citep{reflexion,ace}. To use a growing memory
within a limited context window, agents employ compression, selective
eviction, and agent-directed context management
\citep{acon,beyondcompaction,selfgc,acm}. MemGPT lets a model move
information between memory tiers, and CoALA places memory access within
the agent's decision process \citep{memgpt,coala}. In coding harnesses,
multi-round tool interaction lets agents read and update persistent
files to carry progress between sessions
\citep{sweagent,openhands,contextengineering,longrunningharness,appharness}.
NavHarness builds on these practices for physical search, where checking
records may require revisits. Fixed-retrieval and ordinary-summary controls
test the value of agent-directed access and structured search handovers.

%% file: sections/harness.tex
\section{NavHarness}
\label{sec:harness}
\begin{humanprompted}
\raggedbottom

\par\smallskip
\begingroup
\small
\setlength{\tabcolsep}{5pt}
\renewcommand{\arraystretch}{1.08}
\noindent\begin{tabular}{@{}>{\raggedright\arraybackslash}p{2.7cm}>{\raggedright\arraybackslash}p{\dimexpr\linewidth-2.7cm-2\tabcolsep\relax}@{}}
\toprule
\textbf{Term} & \textbf{Meaning and relation} \\
\midrule
Run (deployment) & A sequence of $K$ tasks in one house. \\
Task & Pursuit of goal $g_k$, possibly through several attempts sharing one task budget. \\
Attempt & One continuous search for the current goal, supported by one navigation session. \\
Session (navigation) & The multi-turn multimodal conversation that serves as the reasoning core for an attempt. \\
Hop & A bounded block of model turns and tool calls within a session. \\
\bottomrule
\end{tabular}
\par\endgroup\smallskip

\paragraph{Task setup.}
Unlike single-episode navigation evaluation
\citep{r2r}, we study successive navigation goals with reusable
experience. Each goal is specified by language, an object category, or an image.
Within a continuous sequence, each task starts at the previous task's
endpoint.
\emph{Recovery} replaces the conversation while retaining physical state,
external memory, and remaining budget for another attempt at the same goal.
\emph{Task closure} records the outcome before the next goal, including after
budget exhaustion. At run end, \emph{consolidation} summarizes task records for future runs.

The outer \emph{orchestrator} manages session boundaries, handles recovery and
completion requests, and retains memory. It advances the session in hops without
clearing the conversation, tracking the task budget and robot state between them.
Figures~\ref{fig:main} and~\ref{fig:task-flow} show memory reuse and execution,
respectively. Appendix~\ref{app:protocol} gives the complete execution loop
(Algorithm~\ref{alg:navharness-full}), the six-component specification
\citep{agentharness}, and detailed transition rules.

\begin{figure}[!t]
\centering
\includegraphics[width=0.94\linewidth]{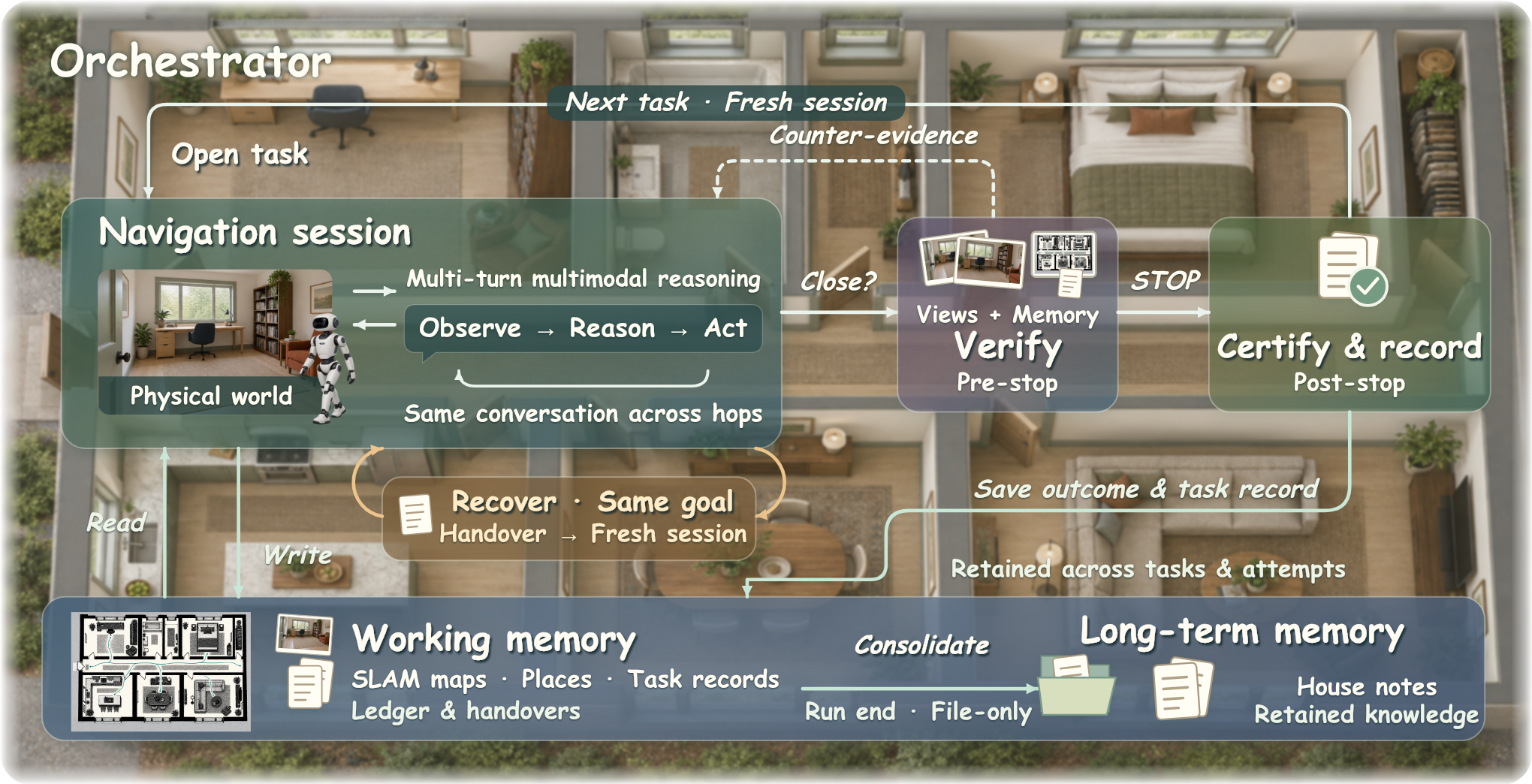}
\caption{\textbf{NavHarness execution flow.} The orchestrator manages
navigation sessions and persistent memory. Verify and Certify denote
pre-stop verification and post-stop certification, respectively.}
\label{fig:task-flow}
\end{figure}

\subsection{Starting a task with retained experience}
\label{sec:memory}\label{sec:context}

On receiving a new goal, the orchestrator sets model-turn and movement-step
budgets and opens a fresh coding-agent conversation with access to persistent
working memory, instructing it to read the referenced files before moving.
When available, these references point to the previous task's handover,
the run ledger, and the house notes. Their contents enter the conversation
through file reads rather than being inserted wholesale into the opening
prompt, and the session can consult further records as the search develops.
A \emph{handover} preserves search experience for a later session.
At task closure, a \emph{task handover} summarizes the goal, searched
places, outcome evidence, and unfinished searches. During recovery,
a \emph{recovery note} instead passes evidence and remaining options
to a fresh attempt at the same goal (Section~\ref{sec:hooks}).
The \emph{ledger} is a compact index with one entry per closed task,
listing its goal, recorded status, and named places. It helps later
sessions find relevant tasks and open their detailed records through
file tools, without carrying over earlier conversations.
The navigation session writes its own task handover and recovery notes,
while the orchestrator writes the ledger and verdict files. It can read
earlier task records but cannot overwrite them.

The navigation session accesses three memory tiers. The \emph{active
context} holds the current attempt's goal, reasoning, observations, and
retrieved information. \emph{Working memory} persists outside the
conversation across tasks and attempts, pairing task records with spatial
memory comprising occupancy maps built using SLAM-estimated poses,
named places and photographs, and room and floor connections. Task
records, which form the journal, include the ledger, task handovers, recovery notes, and
completion-check results (Section~\ref{sec:task}).
\emph{Long-term memory} comprises four files, together called \emph{house notes},
containing an index, a house overview, room notes, and navigation skills. They cover
room connections, landmarks, useful routes,
failed searches, places ruled out by evidence, and unresolved questions.
Navigation sessions read these files to guide searches, but only consolidation
updates them (Section~\ref{sec:consolidation}). Saved maps
and views persist separately from these textual notes for reuse in later runs
(Appendix~\ref{app:protocol}).
The tiers therefore describe how memory is used and how long it remains
available, rather than three disjoint stores. A map shared by tasks in
the current run becomes part of the retained house memory when saved
for a later run, without being converted into a textual summary.

\subsection{Navigating by consulting and updating memory}
\label{sec:spatial}\label{sec:tools}\label{sec:loop}

Starting from this context, the navigation session interleaves observation,
reasoning, action, and further memory access \citep{mip}. A house note can
suggest a room, a task record can explain what was already searched,
and a map query can show how to reach that room. Tool responses bring retrieved
information into the conversation for comparison with new observations,
helping the model decide whether to follow an earlier account or query further.
Following prior work, we use the standard navigation action space of
forward movement (0.25\,m), left/right turns (15$^{\circ}$), and STOP.

Reaching a remembered place also requires locating the robot and planning
through explored space. ORB-SLAM3 \citep{orbslam3} estimates the robot's
position and orientation (pose) from color and depth (RGB-D) observations.
A mapper uses these estimates to build an occupancy grid for each floor,
distinguishing free, occupied, and unexplored space. As the robot moves, the mapper updates the
grids while pose tracking is available, pausing updates if tracking is
lost. The navigation session queries this growing map to inspect explored
space and preview routes, and can mark named places with saved views
for later reference (Appendix~\ref{app:interfaces}).
The grid is built by projecting depth observations through the estimated
poses, rather than using the tracker's point cloud directly. Route queries
run A$^*$ over known free cells and return a path and its length for the
session to consider. Named markers link positions in this grid to saved
views, allowing a later session to compare a remembered place with what
it currently sees.

Closure and recovery tools pass requests to the orchestrator, which
handles them when the hop returns. It checks the remaining budget and
tracking status. If the environment has already ended the task, the
orchestrator records its outcome. Otherwise, it handles a pending
closure request before considering recovery. If neither requires a transition, it resumes
the same conversation with its context intact. The coding backend manages
the transcript within that conversation, while NavHarness opens fresh
sessions at task and recovery boundaries. Separating transcript management
from these boundaries lets different reasoning cores use the same memory rules
(Appendix~\ref{app:context}).
NavHarness adds no image-eviction window or transcript-summary policy
inside a running session. The task's turn and movement budgets span all
its hops and attempts, so replacing a conversation does not grant a new
search budget.

\subsection{Recovering with a fresh conversation}
\label{sec:hooks}

When a session repeatedly follows an unproductive plan, recovery lets a
fresh conversation reconsider the search using retained evidence.
Recovery begins with a
session request or when the attempt reaches a configured turn threshold.
The orchestrator checks
exploration progress, tracking status, and previous attempts before
approving a restart and deciding whether location also needs reassessment.
Before replacement, the outgoing session writes a \emph{recovery note},
a handover to the next attempt at the same goal. With movement disabled,
it records searched places, evidence-supported
exclusions, a reliable landmark, untried options, and uncertainty.
Distinguishing \emph{searched} from \emph{ruled out} is important because
visiting a room does not establish that every relevant object was seen.
The note names the observation supporting each exclusion, so that the
next attempt can distinguish evidence against a location from an
unfinished search there.
The fresh session reads the note and resumes the same goal from the current
location, retaining the map, task records, house notes, and remaining budget
without inheriting the old conversation.

The recovery \emph{scope} specifies what the new attempt should reconsider,
from its search plan to its location, route, floor, or understanding of
the house, without deleting stored memory. When location needs reassessment,
a separate \emph{wake-up session} compares surrounding views with saved
place photographs and gives the new navigation session a short account
of its likely location.
Without saved photographs, the navigation session reasons about its
location from the supplied views. This visual recognition helps orient
the search, while SLAM estimates the robot's metric pose
(Appendix~\ref{app:hooks}).
The scope changes the beliefs to reconsider, not the robot's position
or the stored observations. Unlike recovery, a prescribed relocation
between task sequences moves the robot before the next task begins.
In that case, the new navigation session receives surrounding views
and uses retained memory to identify its current surroundings.

\subsection{Checking completion and passing on the task record}
\label{sec:task}

When the navigation session decides to stop searching, it writes a
task handover of its search, outcome, and remaining uncertainty, then requests
closure with its own assessment of completion. Because later tasks may rely on this
account, a separate \emph{judge session} performs \emph{verification}
using the goal, four views at the stopping location, sampled earlier frames, the map,
and execution records. Without access to tools, the navigation conversation,
or the navigator's self-assessment, the judge returns an evidence-backed verdict of
\emph{complete}, \emph{incomplete}, or \emph{unknown}.
The visual evidence includes up to twelve frames, prioritizing the four
stopping views and sampling the remainder from earlier observations.

For \emph{pre-stop verification}, the orchestrator compares the judge's
independent verdict with the navigator's completion claim before issuing
STOP. If the first checked request yields a definite verdict that
contradicts the claim, the orchestrator returns evidence to the current navigation session,
giving it a chance to continue searching. An unknown verdict does not
block closure, and a repeated request is honored.
This one-time check belongs to the task and is not reset by recovery,
preventing repeated vetoes from indefinitely delaying closure.
\emph{Post-stop certification} records the judge's final assessment without changing the task's
score. It reuses the earlier verdict if the action-step count is unchanged and
otherwise checks the final evidence (Appendix~\ref{app:closure}).

The orchestrator saves the verdict, appends the ledger entry, and retains
spatial memory for the next task's conversation, including when the task
ends through budget exhaustion. Claims and verdicts remain separate because a
completion check does not validate every spatial assertion in a handover.

\subsection{Consolidating experience for the next run}
\label{sec:consolidation}

To spare later sessions from inspecting every past task account, a separate
\emph{consolidation session} reads the journal and existing house notes at
run end, then rewrites the index, house overview, room notes, and navigation
skills. It adds a skill only when supported by the journal. Using file tools
only, it integrates experience and corrections while distinguishing checked
outcomes from navigator claims and preserving source records
(Appendix~\ref{app:memory}).
Its write access is restricted to the four house-note files, so
consolidation neither changes the map geometry nor overwrites the
underlying task accounts. Maps, place records, and task experience can
already persist before this summary is written.
On a later run in the same house, the navigation session reads these notes
alongside saved maps and views to connect past experience to its current
surroundings, including after a prescribed relocation between sequences.
We evaluate this cross-run reuse in Section~\ref{sec:exp:day}, whereas benchmark
runs in Sections~\ref{sec:exp:gap} and~\ref{sec:exp:organs} start with empty memory.

\end{humanprompted}

%% file: sections/experiments.tex
\begin{humanconfirmed}
\section{Experiments}
\label{sec:experiments}
\definecolor{scoreblue}{HTML}{E3F0FA}

\subsection{Experimental setting}
\label{sec:exp:setting}
\begin{humanprompted}
We evaluate NavHarness on the validation-unseen splits of GOAT-Bench
\citep{goatbench} and IR2R-CE \citep{ivln}. On GOAT-Bench, we evaluate
all 2,669 subtasks in 360 episodes across 36 scenes. Our reasoning cores are
Qwen3.8-27B \citep{qwen38}, GPT-4o \citep{gpt4o}, Opus 5 \citep{claudeopus5}, and GPT-6 Astra
\citep{gpt6astra}, all provided with the same navigation and memory tools.
Comparisons within each model use the same evaluation tasks and task
budgets. We report individual-task success (s-SR) and the fraction of
sequences in which every task succeeds (e-SR) \citep{ssmgnav}, together
with SPL for path efficiency \citep{spl} and IR2R-CE's t-nDTW for
trajectory fidelity \citep{ivln,ndtw}. Appendix~\ref{app:benchmarks}
provides the backend configurations and detailed evaluation settings.
Memory carries from one subtask to the next within a GOAT-Bench episode,
and between tasks within an IR2R-CE tour. Each new episode or tour starts
with empty maps, task records, and house notes, without inheriting memory
from the previous one.
\end{humanprompted}

\subsection{Comparison with state-of-the-art methods}
\label{sec:exp:gap}
\begin{table}[!ht]
\draftstatus{black}
\setlength{\belowcaptionskip}{3pt}
\caption{Comparison with SOTA methods on GOAT-Bench Val-Unseen.
GT pose: simulator ground-truth pose. $^{\dagger}$: 278-subtask evaluation.
Our scores: mean $\pm$ s.d.\ over three seeds.}
\label{tab:goat-main}\label{tab:landscape}
\centering\fontsize{8.5}{10}\selectfont\setlength{\tabcolsep}{2.5pt}
\renewcommand{\arraystretch}{0.92}
\begin{tabular*}{\linewidth}{@{\extracolsep{\fill}}lllcccc@{}}
\toprule
\textbf{Method} & \textbf{Base Model} & \textbf{Type} & \textbf{GT pose} & \textbf{s-SR} & \textbf{SPL} & \textbf{e-SR} \\
\midrule
MemoryExplorer$^{\dagger}$ \citep{lmee} & Qwen2.5-VL-7B & trained & \checkmark & 46.4 & 28.0 & -- \\
SSMG-Nav \citep{ssmgnav} & Qwen-VL-Plus & zero-shot & \checkmark & 46.5 & 34.1 & 8.6 \\
EvoMemNav \citep{evomemnav} & Qwen3-VL-8B & zero-shot & \checkmark & 59.6 & 38.9 & -- \\
ReEXplore$^{\dagger}$ \citep{reexplore} & GPT-4o & zero-shot & \checkmark & 59.8 & 42.5 & -- \\
AstraNav-Memory \citep{astranavmem} & Qwen2.5-VL-3B & trained & \checkmark & 62.7 & 56.9 & -- \\
STEGNav \citep{stegnav} & GPT-5.4-mini & zero-shot & \checkmark & 66.3 & 39.7 & -- \\
GSMem \citep{gsmem} & GPT-4o & zero-shot & \checkmark & 67.2 & 46.9 & -- \\
3D-Mem$^{\dagger}$ \citep{threedmem} & GPT-4o & zero-shot & \checkmark & 69.1 & 48.9 & -- \\
MetaNav \citep{metanav} & GPT-4o & zero-shot & \checkmark & 71.4 & 51.8 & -- \\
HGR$^{\dagger}$ \citep{hgr} & GPT-4o & zero-shot & \checkmark & 72.4 & 56.2 & -- \\
ObsGraph$^{\dagger}$ \citep{obsgraph} & GPT-4o & zero-shot & \checkmark & 72.7 & 51.5 & -- \\
HIMM$^{\dagger}$ \citep{himm} & GPT-4o & zero-shot & \checkmark & 72.8 & 56.1 & -- \\
\midrule
\multicolumn{7}{@{}l}{\textit{Ours (same-backbone comparisons)}} \\
Independent sessions & \multirow{3}{*}{GPT-4o} & \multirow{3}{*}{agentic} & \multirow{3}{*}{$\times$} & 43.5\,$\pm$\,0.6 & 30.2\,$\pm$\,0.5 & 6.4\,$\pm$\,0.6 \\
NavHarness (subset)$^{\dagger}$ &  &  &  & \cellcolor{scoreblue}77.6\,$\pm$\,1.0 & \cellcolor{scoreblue}56.4\,$\pm$\,0.8 & \cellcolor{scoreblue}25.0\,$\pm$\,2.8 \\
NavHarness &  &  &  & \cellcolor{scoreblue}78.3\,$\pm$\,0.5 & \cellcolor{scoreblue}57.1\,$\pm$\,0.4 & \cellcolor{scoreblue}26.9\,$\pm$\,0.8 \\
\addlinespace[2pt]
Independent sessions & \multirow{2}{*}{Qwen3.8-27B} & \multirow{2}{*}{agentic} & \multirow{2}{*}{$\times$} & 41.4\,$\pm$\,0.6 & 27.5\,$\pm$\,0.5 & 3.1\,$\pm$\,0.5 \\
NavHarness &  &  &  & \cellcolor{scoreblue}71.7\,$\pm$\,0.5 & \cellcolor{scoreblue}48.2\,$\pm$\,0.4 & \cellcolor{scoreblue}14.4\,$\pm$\,0.7 \\
\addlinespace[2pt]
Independent sessions & \multirow{2}{*}{Opus 5} & \multirow{2}{*}{agentic} & \multirow{2}{*}{$\times$} & 58.9\,$\pm$\,0.9 & 34.9\,$\pm$\,0.8 & 5.8\,$\pm$\,0.6 \\
NavHarness &  &  &  & \cellcolor{scoreblue}81.5\,$\pm$\,0.7 & \cellcolor{scoreblue}55.0\,$\pm$\,0.6 & \cellcolor{scoreblue}28.6\,$\pm$\,0.8 \\
\addlinespace[2pt]
Independent sessions & \multirow{2}{*}{GPT-6 Astra} & \multirow{2}{*}{agentic} & \multirow{2}{*}{$\times$} & 65.1\,$\pm$\,0.6 & 40.1\,$\pm$\,0.5 & 13.6\,$\pm$\,0.7 \\
\textbf{NavHarness} &  &  &  & \cellcolor{scoreblue}\textbf{83.7\,$\boldsymbol{\pm}$\,0.4} & \cellcolor{scoreblue}\textbf{62.3\,$\boldsymbol{\pm}$\,0.3} & \cellcolor{scoreblue}\textbf{36.9\,$\boldsymbol{\pm}$\,0.9} \\
\bottomrule
\end{tabular*}
\vspace{-6pt}
\end{table}

\begin{humanprompted}
With each reasoning model held fixed, NavHarness improves GOAT-Bench
s-SR by 30.3 points for Qwen3.8-27B, 34.8 for GPT-4o, 22.6 for Opus 5, and 18.6 for
GPT-6 Astra over independent sessions (Table~\ref{tab:goat-main}).
These MIP-style controls \citep{mip} use only the current task's context,
so the gains reflect the full harness, including mapping, recovery, and
completion checks. Sections~\ref{sec:exp:organs} and~\ref{sec:exp:day}
isolate memory's contribution. Astra also improves SPL from 40.1 to 62.3,
while its 17.5-point s-SR gain on IR2R-CE is close to its 18.6-point
gain on GOAT-Bench.

With the same GPT-4o backbone and the 278-subtask protocol, NavHarness
reaches 77.6 s-SR, exceeding HIMM (72.8) and 3D-Mem (69.1) without
simulator poses \citep{himm,threedmem}. With Astra, NavHarness reaches
83.7\% s-SR and 62.3 SPL on GOAT-Bench, exceeding the strongest prior
scores in Table~\ref{tab:goat-main} by 10.9 and 5.4 points, respectively
\citep{himm,astranavmem}. On IR2R-CE (Table~\ref{tab:ivln-main}),
it reaches 85.9\% s-SR and 76.1 SPL, compared with SeqWalker's 36\% and 34
\citep{seqwalker}. The open 27B Qwen executor also reaches 71.7\% s-SR
on GOAT-Bench, 2.6 points above the published 3D-Mem score
\citep{threedmem}.
NavHarness obtains these results without simulator
poses, a global navmesh, auxiliary perception models, or navigation-specific
learned policies.
\end{humanprompted}

\begin{table}[H]
\draftstatus{black}
\setlength{\belowcaptionskip}{3pt}
\caption{IR2R-CE Val-Unseen. GT pose: simulator pose. Ours: three-seed mean $\pm$ s.d.}
\label{tab:ivln-main}
\centering\fontsize{8.5}{10}\selectfont\setlength{\tabcolsep}{3pt}
\renewcommand{\arraystretch}{0.82}
\begin{tabular*}{\linewidth}{@{\extracolsep{\fill}}llccccc@{}}
\toprule
\textbf{Method} & \textbf{Type} & \textbf{GT pose} & \textbf{s-SR} & \textbf{SPL} & \textbf{e-SR} & \textbf{t-nDTW} \\
\midrule
CMA \citep{ivln} & trained & $\times$ & 19 & 18 & -- & 38 \\
MAP-CMA \citep{ivln} & trained & \checkmark & 35 & 32 & -- & 47 \\
ETPNav \citep{seqwalker} & trained & \checkmark & 28 & 27 & -- & 41 \\
HNR \citep{seqwalker} & trained & \checkmark & 30 & 28 & -- & 44 \\
OVER-NAV \citep{overnav} & hybrid & \checkmark & 35 & 33 & -- & 50 \\
SeqWalker \citep{seqwalker} & trained & \checkmark & 36 & 34 & -- & 52 \\
\midrule
\multicolumn{7}{@{}l}{\textit{Ours (same-backbone comparisons)}} \\
Independent sessions (Qwen3.8-27B) & agentic & $\times$ & $29.0\pm0.9$ & $21.0\pm1.1$ & $0.0\pm0.0$ & $36.9\pm1.2$ \\
NavHarness (Qwen3.8-27B) & agentic & $\times$ & \cellcolor{scoreblue}$45.1\pm1.1$ & \cellcolor{scoreblue}$32.0\pm0.9$ & \cellcolor{scoreblue}$0.0\pm0.0$ & \cellcolor{scoreblue}$43.2\pm1.2$ \\
\addlinespace[1pt]
Independent sessions (GPT-4o) & agentic & $\times$ & $37.0\pm1.3$ & $29.0\pm1.2$ & $2.8\pm2.8$ & $43.1\pm0.9$ \\
NavHarness (GPT-4o) & agentic & $\times$ & \cellcolor{scoreblue}$57.8\pm1.2$ & \cellcolor{scoreblue}$45.2\pm1.0$ & \cellcolor{scoreblue}$13.9\pm2.8$ & \cellcolor{scoreblue}$52.9\pm1.1$ \\
\addlinespace[1pt]
Independent sessions (Opus 5) & agentic & $\times$ & $59.3\pm1.1$ & $32.8\pm1.2$ & $2.8\pm0.0$ & $45.1\pm1.3$ \\
NavHarness (Opus 5) & agentic & $\times$ & \cellcolor{scoreblue}$78.2\pm0.9$ & \cellcolor{scoreblue}$56.9\pm1.3$ & \cellcolor{scoreblue}$16.7\pm2.8$ & \cellcolor{scoreblue}$58.1\pm1.0$ \\
\addlinespace[1pt]
Independent sessions (GPT-6 Astra) & agentic & $\times$ & $68.4\pm1.4$ & $53.6\pm0.8$ & $9.3\pm1.6$ & $56.8\pm1.2$ \\
\textbf{NavHarness (GPT-6 Astra)} & agentic & $\times$ & \cellcolor{scoreblue}\textbf{85.9\,$\boldsymbol{\pm}$\,0.8} & \cellcolor{scoreblue}\textbf{76.1\,$\boldsymbol{\pm}$\,0.9} & \cellcolor{scoreblue}\textbf{27.8\,$\boldsymbol{\pm}$\,2.8} & \cellcolor{scoreblue}\textbf{64.2\,$\boldsymbol{\pm}$\,1.1} \\
\bottomrule
\end{tabular*}
\vspace{-6pt}
\end{table}

\begin{humanprompted}
For lifelong navigation, these improvements must carry through a whole
sequence of goals. With Astra, GOAT-Bench e-SR rises from 13.6\% in
independent sessions to 36.9\% with NavHarness, meaning that over a
third of tours finish with every subtask completed. The strongest prior
e-SR reported in Table~\ref{tab:goat-main} is 8.6\% for SSMG-Nav
\citep{ssmgnav}. On IR2R-CE, Astra's e-SR rises from 9.3\% to 27.8\%.
\end{humanprompted}

\subsection{What carries navigation across tasks and attempts?}
\label{sec:exp:organs}\label{sec:exp:insights}
\label{sec:exp:triggers}\label{sec:exp:verification}
We use Opus 5 on all 2,669 GOAT-Bench Val-Unseen subtasks with matched task budgets.
Table~\ref{tab:organs-ablation} compares session policies (A), cross-task
memory (B), recovery handovers (C), and completion checks (D).
Recovery shares the original task's budget
(Appendices~\ref{app:defensive} and~\ref{app:stats}).

\begin{table}[!htbp]
\draftstatus{black}
\caption{Opus 5 comparisons on the full GOAT-Bench Val-Unseen split.
Scores are mean $\pm$ s.d.\ over three seeds (\%).
$\Delta$s-SR denotes the paired change from full.}
\label{tab:organs-ablation}\label{tab:recovery}\label{tab:organs-ci}
\centering\fontsize{8.5}{10}\selectfont\setlength{\tabcolsep}{3pt}
\begin{tabular}{@{}lrrrr@{}}
\toprule
Configuration / intervention & s-SR & SPL & $\Delta$s-SR & 95\% CI \\
\midrule
NavHarness (full reference) & $81.5\pm0.7$ & $55.0\pm0.6$ & -- & -- \\
\addlinespace
\multicolumn{5}{@{}l}{\textbf{A. Session policies}} \\
Independent sessions & $58.9\pm0.9$ & $34.9\pm0.8$ & $-22.6$ & $[-26.3,-19.0]$ \\
Single session & $54.9\pm1.1$ & $32.8\pm0.9$ & $-26.6$ & $[-30.5,-22.8]$ \\
Independent sessions + pre-stop verification & $62.4\pm0.8$ & $36.5\pm0.7$ & $-19.1$ & $[-22.4,-16.4]$ \\
\addlinespace
\multicolumn{5}{@{}l}{\textbf{B. Cross-task memory}} \\
Clear map after each task (keep records) & $68.9\pm0.8$ & $39.8\pm0.7$ & $-12.6$ & $[-15.6,-9.7]$ \\
Hide earlier task records (keep map) & $71.8\pm0.7$ & $41.8\pm0.6$ & $-9.7$ & $[-12.5,-7.0]$ \\
No cross-task memory & $66.8\pm0.9$ & $37.4\pm0.7$ & $-14.7$ & $[-16.9,-12.4]$ \\
\addlinespace
\multicolumn{5}{@{}l}{\textbf{C. Recovery handovers}} \\
No recovery (same task budget) & $71.1\pm0.9$ & $46.1\pm0.8$ & $-10.4$ & $[-13.2,-7.7]$ \\
Recovery + empty handover & $72.9\pm0.8$ & $44.7\pm0.7$ & $-8.6$ & $[-11.3,-6.0]$ \\
Recovery + another task's handover & $68.5\pm1.0$ & $44.8\pm0.8$ & $-13.0$ & $[-16.1,-10.0]$ \\
Recovery + matched-length summary & $73.2\pm0.7$ & $47.0\pm0.6$ & $-8.3$ & $[-11.0,-5.7]$ \\
\addlinespace
\multicolumn{5}{@{}l}{\textbf{D. Completion checks}} \\
No pre-stop verification & $76.1\pm0.8$ & $47.0\pm0.7$ & $-5.4$ & $[-7.8,-3.1]$ \\
No post-stop certification & $79.0\pm0.6$ & $48.2\pm0.6$ & $-2.5$ & $[-4.5,-0.6]$ \\
One stop view instead of four & $78.6\pm0.7$ & $47.9\pm0.6$ & $-2.9$ & $[-5.0,-0.9]$ \\
\bottomrule
\end{tabular}
\end{table}

\begin{humanprompted}
\paragraph{A. Session policies.}
A single session continues the conversation across tasks, using backend
compaction only when the context window fills. Yet it performs worse
than independent sessions (54.9\% versus 58.9\%), which start afresh
for each task with only the current conversation and no external memory.
The longer conversation also retains outdated goals and observations,
which may compete with information relevant to the current task.
NavHarness gives each task a fresh conversation while keeping prior
experience available through external memory.
Adding pre-stop verification to independent sessions still leaves a
19.1-point gap to NavHarness. The remaining comparisons examine the
memory and recovery mechanisms behind this gap.
\end{humanprompted}

\paragraph{B. Cross-task memory.}
All B controls retain within-task mapping, recovery, and verification.
Clearing the map and markers preserves earlier task records, whereas hiding
those records, including the ledger and handovers, preserves the map.
No cross-task memory keeps these task-local mechanisms but removes both forms of carryover,
isolating cross-task reuse with a 14.7-point s-SR loss.
Hiding records costs 9.7 s-SR points and increases steps by 25\%, while
clearing spatial memory costs 12.6 points and increases steps by 32\%.
These interventions leave later tasks to reconstruct routes or repeat
earlier searches, adding exploration that retained experience could avoid.

\paragraph{C. Recovery handovers.}
To separate restarting from passing on experience, we keep recovery
but remove or replace its handover. An empty handover reaches 72.9\%
success, compared with 71.1\% without recovery, while another task's
note performs worse. Resetting the conversation removes its history,
but without a useful handover the new attempt also loses the evidence
needed to choose a different search. A matched-length ordinary summary
still trails the structured handover by 8.3 points with unchanged
triggers and budgets. The handover distinguishes places that were visited
from those where the target was ruled out by evidence, and retains untried
options. This gives the next attempt a basis for choosing where to search.

\paragraph{D. Completion checks.}
Unlike memory used throughout a search, completion checks intervene
when the agent proposes to stop. Pre-stop verification can prolong the
search, whereas post-stop certification only changes the record that
later tasks inherit. This helps explain their smaller s-SR effects
(5.4 and 2.5 points). Certification also reaches 90.7\% record accuracy
against 82.6\% for accepting every claim, a benefit not captured by
success rate alone (Appendix~\ref{app:closure-diag}).

\subsection{Continuous deployment}
\label{sec:exp:day}
\begin{humanprompted}
To study memory reuse beyond the benchmark horizon, we concatenate ten
GOAT-Bench tours per house across simulated days. At each tour boundary,
the robot is placed at the next prescribed start, modeling resumption
after a shutdown. It must recover its bearings using SLAM and stored
house knowledge before reusing earlier experience. Detailed case studies
are provided in Appendix~\ref{app:selfcorrect}.

To test what consolidation adds beyond retained maps and task records,
we compare NavHarness with \emph{No consolidation} across all 36 houses
using Opus 5. Both retain maps and task records across tasks and use the
same recovery, verification, and task budgets. The control disables only
run-end consolidation, so it does not build house notes for later runs.
Consolidating experience into long-term memory raises pooled s-SR from
72.8\% to 80.5\% and SPL from 36.5 to 44.3,
with paired gains of 7.7 and 7.8 points
(Figure~\ref{fig:deployment-progress-preview}). Beyond these quantitative
results, we examine the deployment records for qualitative insights into
how the agent uses memory to interpret goals, guide its search, and
reconsider earlier conclusions.
\end{humanprompted}

\begin{figure}[!htbp]
\draftstatus{black}
\centering
\includegraphics[width=\linewidth]{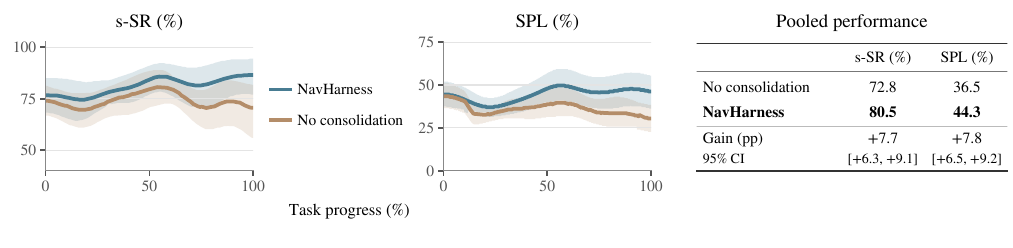}
\vspace{-24pt}
\caption{Continuous deployment across 36 houses with Opus 5, varying only
consolidation. The table reports pooled scores, paired gains, and 95\% CIs.}
\label{fig:deployment-progress-preview}
\end{figure}

\paragraph{Interpreting goals through experience.}
In a repeated dresser task, the phrase ``above the staircase handrail''
sends the agent without consolidation upstairs, although the target is in the
freezer room. NavHarness reads a house note connecting that
same wording to the dresser and searches in the correct room.
The stored association lets the session identify the intended object
even when the wording alone suggests a different search location.
It reads this note before moving or querying the map, resolving which
floor and object the instruction refers to before planning a route.

These associations need not be settled when first recorded. The house
notes initially leave open whether two refrigerator descriptions refer
to one appliance or two. After completing a later task, the session
returns to this question, inspects both faces of the refrigerator, and
records its answer for consolidation. When a subsequent task supplies a dark photograph,
NavHarness recognizes the appliance, while the agent without consolidation
searches for a second refrigerator in the basement.
The session has gathered information beyond what its current goal
requires, prompted by an unresolved question in memory. That answer
then helps a future session interpret a new goal.

\paragraph{Reusing failed attempts and revising memory.}
Useful experience also comes from tasks that did not succeed. In the
mirror search, an earlier attempt runs out of steps but records a wall
left unchecked behind an open door. A later session initially follows
an incomplete house note to the wrong mirror. After recovery, it
compares the goal image with the note and searches the journal,
where it finds that earlier suggestion.
The failed attempt has preserved a concrete place to resume, even
though the consolidated notes do not contain it. Keeping the detailed
record lets the session check what was searched and what remained
unchecked when the house summary proves insufficient. It can then
choose its next search from the earlier observations rather than relying
entirely on what consolidation retained.

Sometimes only part of a record needs revision. A stored note describes
a corner with two plants but advises avoiding it. When a new goal
photograph asks for one of those plants, the agent uses the description
to locate the corner, finds a suitable viewpoint, and records a correction
that consolidation incorporates into the note. It can keep the spatial
knowledge while revising the advice attached to it. Similarly, after a
restart changes the SLAM frame, landmark descriptions remain useful
while coordinates are re-established. Across these cases, the reasoning
session determines which parts of memory apply to the current task
and supplies observations that later sessions can reuse.

\begin{humanprompted}
Memory reliability remains a limitation. An incorrect completion accepted
by the judge can propagate into house notes, and correcting one record
does not necessarily update its copies or validate other spatial claims.
Appendix~\ref{app:limitations} discusses these remaining risks.
\end{humanprompted}

\end{humanconfirmed}

%% file: sections/discussion.tex
\begin{humanprompted}

\FloatBarrier
\section{Conclusion}

NavHarness improves navigation across four reasoning models by letting
multi-round sessions use maps, search records, and house knowledge without
additional training. Fresh attempts benefit from structured search evidence
that survives the previous conversation, allowing the robot to reconsider
its plan without discarding what it has learned about the search. In extended
deployments, this experience helps interpret ambiguous goals, investigate
unresolved questions, and resume failed searches, while new observations
lead to corrections that later sessions can use. These results support
building lifelong navigation around successive reasoning sessions that
inherit useful experience and continue to check and revise it as they act.
\end{humanprompted}

%% file: sections/app_algorithm.tex
\begingroup
\setlength{\intextsep}{8pt}
\begin{algorithm}[H]
\caption{NavHarness over a sequence of $K$ tasks}
\label{alg:navharness-full}
\fontsize{8.5}{10}\selectfont
\begin{algorithmic}[1]
\Require goals $g_{1:K}$, navigation and memory tools, recovery controller, judge,
task limits, and hop limit $H$
\State mount spatial memory, task records, and any existing house notes
\For{each goal $g_k$}
  \State write the goal and prepare instructions referring to available memory files
  \If{the robot was relocated since the previous task}
    \State attach surrounding views and an instruction to recognize its current location
  \EndIf
  \State open a fresh navigation session and initialize the task's counters and closing claim
  \While{the task is open and its execution budget remains}
    \State advance the session for up to $H$ model turns, including tool calls
    \State read execution state and pending requests from the tool bridge
    \If{the environment has ended the task}
      \State \textbf{break}
    \ElsIf{the session has requested task closure}
      \If{the task handover is missing and has not yet been requested}
        \State ask the session to write its handover; \textbf{continue}
      \EndIf
      \If{this is the first closure request to reach verification}
        \State obtain a judge verdict from the goal, views, map, and execution records
        \If{the verdict is definite and contradicts the navigator's closing claim}
          \State return counter-evidence to the same session; \textbf{continue}
        \EndIf
      \EndIf
      \State issue STOP and close the task
    \Else
      \State let the recovery controller assess any request or enabled duration fallback
      \If{recovery is approved}
        \State let the outgoing session write a recovery note in a final, file-only hop
        \State prepare the selected scope's reassessment and any location brief
        \State retain the goal, physical state, map, task records, and remaining budget
        \State replace the conversation with a fresh session instructed to read the recovery note
      \Else
        \State resume the same session unless persistent inactivity requires termination
      \EndIf
    \EndIf
  \EndWhile
  \State ensure the task has ended and record any budget or inactivity termination
  \State certify the final outcome, reusing the pre-stop verdict if the action-step count is unchanged
  \State collect the claim, verdict, available task handover, and named places into a task record
  \State write the verdict and ledger entry, update place memory, and save the run checkpoint
  \State release the acting conversation before opening the next task's session
\EndFor
\State close the acting session
\If{cross-run consolidation is enabled}
  \State a separate file-only session consolidates the journal into the four house-note files
\EndIf
\end{algorithmic}
\end{algorithm}
\endgroup

%% file: sections/app_protocol.tex
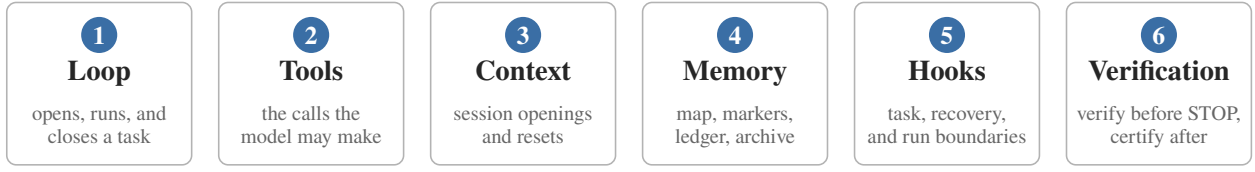
\begin{figure}[H]
\centering
\resizebox{\linewidth}{!}{\input{figures/fig-six}}
\caption{The six components of NavHarness and their responsibilities.}
\label{fig:six}
\end{figure}

NavHarness comprises six components: the loop, the tool registry, the
context policy, the memory, the hooks that fire at boundaries, and the
verification of a claimed completion (Figure~\ref{fig:six}). This is the
decomposition used by \citet{agentharness}. \citet{agentsystemsurvey}
also distinguishes observation, action, and governance interfaces.
We use the six component names to organize
the implementation. NavHarness specifies the state retained at navigation
boundaries and the handover between attempts, as detailed below.

\subsection{The loop}
\label{app:full-algorithm}

A coding-agent session is the reasoning core of NavHarness. It uses
multi-round multimodal dialogue to interpret observations, consult memory,
and invoke navigation and file tools. The outer orchestrator manages the
sequence of tasks and decides when that session should continue, close,
or hand its experience to a fresh attempt. Algorithm~\ref{alg:navharness-full}
describes this lifecycle. Tool execution remains inside the coding
session, so the orchestrator need not implement another loop for
model inference and tool execution.

For a deployment in house $h$, we write the state at the closure of task $k$ as
\begin{equation}
X_k=(C_k,\,M_k,\,L_k,\,D_h),\qquad
M_k=\big(\{G^{(f)}\}_f,\;\mathcal P_k,\;\Gamma_k\big),\qquad
L_k=(r_1,\ldots,r_k),
\label{eq:f-state}
\end{equation}
where $C_k$ is the active session context, $G^{(f)}$ the occupancy grid
of floor $f$, $\mathcal P_k$ the named places and their views,
$\Gamma_k$ the place graph, $L_k$ the task records, and $D_h$ the
persistent house notes. Within an attempt, ordinary interactions extend
the conversation as
\begin{equation}
C_{k,a,t+1}=C_{k,a,t}\,\|\,\big(y_t,\;\mathcal A_t,\;o_{t+1}\big),
\qquad (y_t,\mathcal A_t)=\pi_\theta(C_{k,a,t},\mathcal T),
\label{eq:f-loop}
\end{equation}
with reply $y_t$, tool calls $\mathcal A_t$, and returned observations
$o_{t+1}$. Here $a$ indexes attempts and $t$ indexes interactions.
NavHarness does not rewrite the active conversation, and it opens a fresh
session for every task, so no transcript persists across a task boundary.
The single-session comparison is the only configuration that keeps one
conversation across tasks.

Before task $k$ closes, only $L_{k-1}$ is available as prior task history,
since $r_k$ is appended at closure. The opening in Equation~\ref{eq:f-open}
therefore refers to $L_{k-1}$.

The orchestrator advances the session in \emph{hops}. Each hop allows up
to $H$ model turns with their tool calls and returns the resulting
execution state. Several hops may belong to the same conversation.
On return, an environment termination takes priority, followed by a
pending closure request and then recovery. Otherwise the orchestrator
resumes the session with a continuation instruction. Closure and recovery
tools record requests and ask the agent to yield, allowing these
transitions to be handled between hops. The task retains its turn and
physical-step limits $(B,b)$ across attempts. Budget exhaustion or
persistent inactivity also ends a task, even without a closing claim.
The implementation defaults to $H=25$ model turns per hop and an earliest
requested recovery at 12 attempt turns. Three consecutive hops without
a tool call end an inactive task. Three hops with no model turn instead
raise a provider-unavailable error rather than recording navigation
failure. Task budgets and enabled fallback settings are given in
Appendix~\ref{app:benchmarks}.

A session adapter exposes opening, advancing, and closing a conversation
to the orchestrator. Its tool bridge connects the coding loop to the
environment and the memory workspace. This separation lets a different
reasoning model use the same memory and boundary rules, provided its
backend supports the session and tool interfaces. Acting, judging, and
consolidation have separate conversations and tool permissions
(Appendices~\ref{app:context} and~\ref{app:closure}).

\subsection{Tool registry}
\label{app:interfaces}

The model acts only through the calls in Table~\ref{tab:interfaces}. Movement,
mapping, and route queries are provided by the navigator. The harness adds
file access and requests for closure and recovery. The localisation tool that the spatial
calls draw on is described after the table.

\begin{table}[h]
\caption{The tool registry. Only \tool{step} consumes the task's physical
budget, while free calls cost no steps.}
\label{tab:interfaces}
\centering
\small
\setlength{\tabcolsep}{5pt}
\begin{apptab}{l>{\raggedright\arraybackslash}p{3.4cm}>{\raggedright\arraybackslash}p{6.4cm}}
\thd{Call} & \thd{Arguments} & \thd{Returns or effect} \\
\midrule
\tool{observe} & none & the forward RGB view and remaining step budget \\
\tool{step} & a list of actions & executes them and reports motion, collision, and budget status \\
\addlinespace
\tool{get\_goal} & none & the current goal, as text or as the goal image \\
\addlinespace
\tool{get\_map} & floor, zoom & the top-down occupancy map with the markers on it \\
\tool{mark} & a place name & drops or moves a named marker at the current position \\
\tool{preview\_path} & a map point & the length of a route over known free space; nothing moves \\
\addlinespace
\tool{list\_files} & a directory & lists the available memory files \\
\tool{read\_file} & a file path & reads a task record or house note \\
\tool{write\_file} & a file path, text & writes a file assigned to this session \\
\addlinespace
\tool{request\_recovery} & scope, reason, confidence & proposes a recovery; the harness decides \\
\tool{close\_task} & status, evidence & captures stop views and requests checked closure \\
\bottomrule
\end{apptab}
\end{table}

\paragraph{Sensing and motion.}
\tool{observe} returns the forward camera image. \tool{step} takes a list
of discrete actions, forward 0.25\,m, turn left 15$^{\circ}$, or turn right
15$^{\circ}$, and executes them in order. Every action costs one step of the
task's physical budget, and the session also has a model-turn limit. The
model is told that turning in place is a cheap way to look around and that it
should alternate looking and moving.

A \tool{step} call accepts up to twelve actions $\alpha=(a_1,\ldots,a_n)$ over the action
set $\mathcal A=\{\text{forward }0.25\,\mathrm m,\;\text{left }15^{\circ},\;\text{right }15^{\circ}\}$,
consumes one step for each executed action, and is executed as up to $5n$
sub-motions of $0.05$\,m or $3^{\circ}$, each of which feeds one frame to the
localisation tool. Execution ends early when the environment terminates the
task. Every other agent-facing call costs no steps.

\paragraph{Action discretization and collisions.}
The simulator executes each sub-motion separately, and the macro-action
ends early if a collision is detected. Sliding is enabled in the simulator's
embodiment configuration.
This refinement supplies the intermediate observations required by SLAM.
Our component and session comparisons share this motion interface.
Benchmark goals and scoring are retained.

\paragraph{Goal.}
\tool{get\_goal} re-reads the current goal. When the goal was given as a
photograph, the call returns the photograph, so an image goal can be
consulted again after other images have entered the context.

\paragraph{Spatial memory.}
\tool{get\_map} returns the occupancy map that the navigator builds as the
body moves, with its named markers. The map persists across tasks, allowing
later searches to reuse explored space and earlier landmarks. \tool{mark}
drops a marker named for the place, not for the object sought, and the harness
attaches to it the views taken at that point, bucketed by heading into at
most four, and the navigator's pose estimate. Re-marking an existing name
moves the marker. \tool{preview\_path} draws a route over the map's known
free space to a point and reports its length without moving the body. All
three are free.

A \tool{mark} with name $a$ at time $t$ replaces any place of that name,
\begin{equation}
\mathcal P\leftarrow\big(\mathcal P\setminus\{z_i:a_i=a\}\big)\cup
\big\{\big(a,\;m_t,\;B_t\big)\big\},\qquad |B_t|\le 4,
\label{eq:f-mark}
\end{equation}
where $m_t$ is the marker at the current estimated pose and $B_t$ the views
taken there, bucketed by heading. \tool{get\_map} returns
$\mathrm{Render}\big(G^{(f)},\mathcal P^{(f)}\big)$ for the requested floor
and zoom, and \tool{preview\_path} evaluates Equation~\ref{eq:preview} on
$G^{(f)}$.

\paragraph{Localisation.}
Behind \tool{observe}, \tool{get\_map}, \tool{mark}, and \tool{preview\_path}
is one estimator, and we treat it as a tool like any other. In the reported
runs it is ORB-SLAM3 \citep{orbslam3} in RGB-D mode, run as a separate process
that the navigator feeds and queries. Nothing in the harness depends on that
choice. The tool consumes the colour and depth frame of each sub-motion and
returns a pose estimate and a status,
\begin{equation}
(\hat{\mathbf T}_t,\, s_t) = \mathrm{SLAM}\big(I_t, D_t \,\big|\, \hat{\mathbf T}_{<t}\big),
\qquad \hat{\mathbf T}_t \in SE(3),\quad s_t \in \{\text{ok}, \text{lost}\},
\label{eq:slam}
\end{equation}
with the world frame anchored at the first frame seen in the house. The
planar pose $(\hat x_t, \hat z_t, \hat\psi_t)$ read off $\hat{\mathbf T}_t$ is
the only position the navigator has. It is what a marker stores, what the map
is drawn in, and what a route is planned from. The simulator's pose is never
read for any of these.

Each coarse action, forward 0.25\,m or a turn of 15$^{\circ}$, is executed as
five sub-motions of 0.05\,m or 3$^{\circ}$, and every sub-motion yields one
tracked frame at a synthetic 10\,Hz. The split exists for the tracker, not
for the body: a single frame per 15$^{\circ}$ turn leaves too few feature matches
and loses tracking on a large share of frames, while 3$^{\circ}$ per frame
keeps inter-frame matches available during navigation. For tracking loss
within an ongoing run, when $s_t$ is lost no
pose is returned and the map is not written. Tracking status is exposed
through motion and map responses and the bridge's execution state. The tracker
opens a fresh map at its next successful initialisation and the navigator
re-anchors it to the last trusted pose, $\hat{\mathbf T} \leftarrow \mathbf A\,
\tilde{\mathbf T}$ with $\mathbf A = \hat{\mathbf T}_{\text{last}}\,
\tilde{\mathbf T}_{\text{first}}^{-1}$. After a controlled relocation the body
turns once in place while feeding the tracker to seek relocalisation
against the retained tracker map. Loading saved occupancy maps and markers
after a tracker restart does not itself recover their coordinate alignment.

\paragraph{Mapping and routes.}
The occupancy map is built from depth and the estimated pose rather than
from the tracker's own point cloud, so it is the same computation for any
estimator that returns a pose. A depth pixel $u$ with $D_t(u) \in [0.3, 4.5]$\,m
is back-projected and placed in the world, and its height over the floor
under the camera is read off,
\begin{equation}
\mathbf p = \hat{\mathbf T}_t\, \pi^{-1}\big(u, D_t(u)\big),
\qquad h(\mathbf p) = h_c - \big(p_y - y_{\mathrm{ref}}\big),
\label{eq:backproject}
\end{equation}
where the mapper uses a downward-positive $y$ axis. Here $y_{\mathrm{ref}}$
is the camera's reference world-$y$ coordinate for the floor, and $h_c$
is its height above that floor. The floor plane is therefore at
$y_{\mathrm{ref}}+h_c$, where the formula gives zero height.
A point is an obstacle when $0.15 < h < 1.4$\,m and floor when
$-0.5 < h \le 0.15$\,m. Points fall into a grid of
$\delta = 0.10$\,m cells, $c(\mathbf p) = \lfloor (p_x, p_z) / \delta \rfloor$,
with three states, unknown, free, and obstacle: an obstacle endpoint marks
its cell, the ray from the body to it carves free cells, and an obstacle is
never overwritten by free. The grid is kept per floor, and a new layer opens
when the estimated height settles more than 1\,m from every registered
floor. \tool{get\_map} renders the current layer with the markers on it. The
grid is written from the online pose, so a later loop-closure correction
inside the tracker is not propagated into cells already written, which is
one of the limitations in Appendix~\ref{app:limitations}.

\tool{preview\_path} answers a route query on the same grid. For a target
cell $c^\star$, snapped to the nearest free cell if it lands on unknown or
obstacle, it runs A$^*$ over the free cells with eight-connected moves and
reports
\begin{equation}
L(c^\star) = \min_{\pi:\, c_0 \to c^\star} \sum_i \delta\, \|c_{i+1} - c_i\|_2,
\label{eq:preview}
\end{equation}
or \emph{unreachable} when no such path exists. Nothing moves. A marker is a
name paired with the planar pose at the moment of the call,
$m_k = (\text{name}_k, (\hat x, \hat z, \hat\psi)_{t_k})$, together with the
views banked there.

\paragraph{Substituting the estimator.}
The localisation estimator supplies pose and tracking status. A separate
mapping adapter combines these with depth to obtain the world points
used by the occupancy map. Another visual or visual-inertial estimator
can use the same adapter if it supplies compatible poses. A planar lidar
would require a different range-projection and obstacle update, while
preserving the navigation tools and orchestrator interface. Thus changing
sensors can require tool-side implementation changes without changing
the session and memory lifecycle. The reported runs use ORB-SLAM3.

\paragraph{Closing a task.}
The agent writes its handover before calling
$\tool{close\_task}(c_k,\epsilon_k)$, where
$c_k\in\{\text{complete},\text{incomplete}\}$ and $\epsilon_k$ is a short
account of the stopping evidence. The tool captures surrounding views,
records the request, and asks the session to yield. It does not issue
STOP itself. At the next hop boundary, the orchestrator prompts once if
the handover file is absent, then applies the pre-stop verification of
Appendix~\ref{app:closure}. An admitted request becomes a physical STOP.
Tasks can also end through execution limits, in which case the ledger
records the termination without inventing a completion claim.

\paragraph{Recovery.}
\tool{request\_recovery} lets the model say that part of what it relies on
is no longer trustworthy, and how much of it, by naming the smallest scope
that fits (Appendix~\ref{app:hooks}). The call costs no steps and does not end
the task. The harness decides what happens: it can refuse, or act at a larger
scope than asked.

The proposal is a triple $(\hat u,\mathrm{reason},c)$: a scope
$\hat u\in\{\text{attempt},\text{place},\text{room},\text{floor},\text{house}\}$,
a textual reason, and a confidence $c\in[0,1]$. The model never proposes
\emph{continue}, which is the harness's answer as decided by
Equation~\ref{eq:controller}.

\paragraph{File access.}
The session reaches textual memory through \tool{list\_files},
\tool{read\_file}, and \tool{write\_file}. Paths are resolved within the
run workspace or its mounted house directory. An acting session can read
earlier records but can write only its assigned handover and recovery
note. The orchestrator owns the ledger and verdict files, while the
consolidation session owns the house notes. Coordinate-like expressions
are scrubbed from session-written prose so that a later map re-anchoring
does not leave stale metric coordinates in text. A read returns at most
20,000 characters and a write stores at most 6,000, with truncation
reported by the tool. These are file-access limits, separate from the
model's context window. The current read tool has no offset parameter:
an overlong file returns only its prefix. Detailed task files can be
located through directory listing and read individually, but a single
ledger read does not guarantee access to an arbitrarily long history.

\subsection{Context management}
\label{app:context}\label{app:curves}

\paragraph{One session per attempt.}
The task-boundary policy opens a fresh coding session with the same
standing system brief $\sigma$. Its first user instruction gives the
goal, names the files to read, and identifies the handover file it may
write. When available, required reading comprises the previous task's
handover, the run ledger, and the existing house notes. Their contents
enter the conversation through the session's own file reads, rather
than being pasted into the system message.

Writing $\rho_k$ for these file references, the opening is
\begin{equation}
C_{k,1,0}=\mathrm{Open}\bigl(\sigma,I(g_k,\rho_k)\bigr),
\qquad
\rho_k=\operatorname{refs}\bigl(h_{k-1},L_{k-1},D_h\bigr),
\label{eq:f-open}
\end{equation}
where nonexistent files are omitted. An image goal remains accessible
through \tool{get\_goal}, and the map through the spatial tools.
The session is instructed to read the listed records before moving and
may consult additional files as the task develops. It then continues
across hops without harness-side editing of the transcript.

\paragraph{Resets at task and attempt boundaries.}
At a task boundary, the orchestrator records the closing status and
makes the actor-written handover available to the next session.
At an approved recovery, the current session first writes a recovery
note, then a new session resumes the same task from that file.
Neither transition carries the old conversation into the new one.
Map state, task records, and house notes remain outside the session and
can be read again. Recovery preserves the physical trajectory and
remaining task budget.

\paragraph{Division of context management.}
NavHarness controls when a conversation ends, while the coding backend
controls how a running conversation is maintained. It introduces no
image-eviction window or additional transcript-summarization policy.
Keeping these responsibilities separate allows the reasoning core to be
replaced without changing the memory interface. Unchanged prefixes also
permit prompt-cache reuse where the provider supports it. The
navigation-turn budget applies to the acting session. Judging and
run-end consolidation use separate model calls.

\paragraph{Session policies and observed context use.}
The independent-session baseline opens a fresh conversation for each task
and relies only on that context, without external working or long-term memory.
Single session keeps using the same conversation after each task,
accumulating observations and tool responses until the backend's context
window fills. The backend then compacts that conversation so it can
continue. Task boundaries do not trigger compaction. NavHarness instead opens fresh conversations
at task boundaries while retaining external memory, with mean peak
prompts of approximately 20k tokens. Table~\ref{tab:organs-ablation}
compares these complete policies, while its memory interventions hold
the other harness mechanisms fixed.

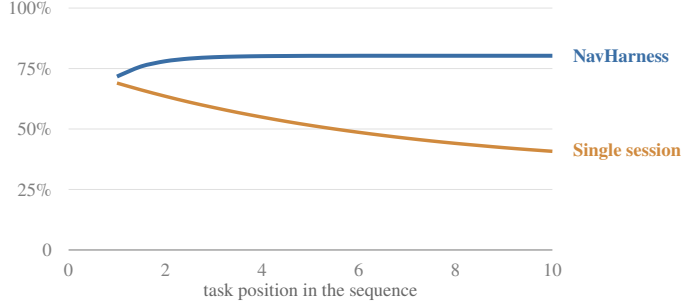
\begin{figure}[H]
\centering
\input{figures/fig-d-succ}
\caption{Success over successive tasks. Fitted trends for NavHarness and
single session over positions 1 to 10.}
\label{fig:d-succ}
\end{figure}

\paragraph{Success over successive tasks.}
Figure~\ref{fig:d-succ} compares the fitted success trends of NavHarness
and the single-session policy over positions 1 to 10.
The curves summarize performance as tasks progress within this range.
The paired system comparison and its uncertainty are reported
in Table~\ref{tab:organs-ablation}. Later positions contain fewer eligible
scenes because episode lengths differ.
Those aggregate intervals are not confidence bands for the curves.
Table~\ref{tab:position-observations} separately reports measured
position-level success from the NavHarness GOAT-Bench Val-Unseen
task-position audit. These are single-run diagnostic observations,
not averages over the three-seed component study.

\begin{table}[ht]
\centering\small
\caption{Observed task-position success in the NavHarness Val-Unseen diagnostic run.}
\label{tab:position-observations}
\begin{apptab}{@{}lrrrrrrrrrr@{}}
Position & 1 & 2 & 3 & 4 & 5 & 6 & 7 & 8 & 9 & 10 \\
\midrule
s-SR (\%) & 72.2 & 72.2 & 83.3 & 83.3 & 77.8 & 82.1 & 73.1 & 85.7 & 81.2 & 85.7 \\
\bottomrule
\end{apptab}
\end{table}

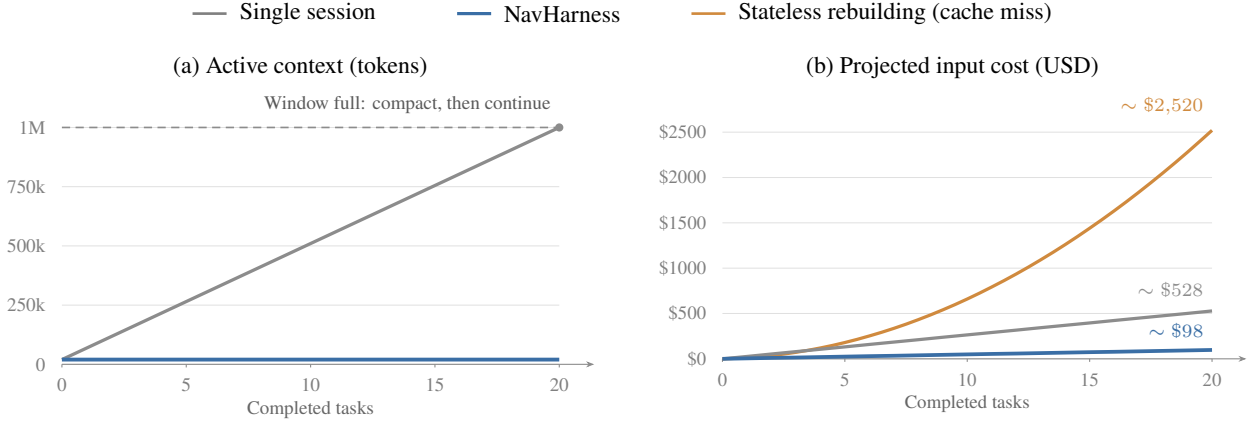
\begin{figure}[!t]
\centering
\begin{tikzpicture}
\draw[gray,line width=1pt] (0,0)--(.45,0);
\node[anchor=west,font=\small] at (.5,0) {Single session};
\draw[figblue,line width=1.2pt] (3.5,0)--(3.95,0);
\node[anchor=west,font=\small] at (4,0) {NavHarness};
\draw[figochre,line width=1pt] (6.6,0)--(7.05,0);
\node[anchor=west,font=\small] at (7.1,0) {Stateless rebuilding (cache miss)};
\end{tikzpicture}
\par\smallskip
\begin{minipage}[t]{.48\linewidth}
\centering\small (a) Active context (tokens)\par\smallskip
\resizebox{\linewidth}{!}{\input{figures/fig-d-ctx}}
\end{minipage}\hfill
\begin{minipage}[t]{.48\linewidth}
\centering\small (b) Projected input cost (USD)\par\smallskip
\resizebox{\linewidth}{!}{\input{figures/fig-d-cost}}
\end{minipage}
\caption{Context and input-cost scaling over the first 20 tasks of
a longer deployment. Orange shows stateless context rebuilding without
cache hits.}
\label{fig:d-ctx-cost}\label{fig:d-ctx}\label{fig:d-cost}
\end{figure}

\paragraph{Context footprint and input-cost scaling.}
Figure~\ref{fig:d-ctx-cost} shows the first 20 tasks of a longer deployment.
In the schematic context panel, single session reaches a roughly
1M-token window near task 20, when the backend compacts the conversation
and continues. This boundary does not end the deployment. NavHarness
instead opens task-level sessions with approximately 20k-token active
contexts. Gray denotes single session in both panels.

The cost panel distinguishes a retained session from \emph{stateless
context rebuilding}, which reconstructs the prompt on every interaction.
The orange curve represents the cache-miss case of the latter, not the
single-session baseline or the independent-session baseline. Retaining
a stable prefix permits cache reuse, whereas rebuilding a prompt can
invalidate that prefix. Actual input cost therefore depends on both
uncached and cached input volumes, $C_K=\sum_{k=1}^{K}(p_uU_k+p_cV_k)$,
where $U_k$ and $V_k$ are the respective token counts for task $k$.

The cost comparison uses an assumed input price of \$5 per million
tokens. With $2.4\times10^6 k$ input tokens at task $k$, stateless
rebuilding without cache hits costs approximately \$2{,}520 over 20
tasks. Using 0.975M input tokens per task for NavHarness gives
approximately \$98 at the same price. For retained sessions, we use a
linear reference, $C_K=528K/20$ dollars, reaching \$528 at task 20.
These calculations cover navigation input tokens only, excluding output
charges and auxiliary calls, and stop at task 20.
Benchmark runtime is reported in Appendix~\ref{app:runtime}.

\subsection{Memory}
\label{app:memory}

Memory is organized by lifetime and access, rather than by disjoint
storage locations. The coding session maintains the current conversation.
Working memory contains the sensor-built map, named places, and task
records shared across successive tasks. When retained for later runs,
maps and place records also form part of long-term house memory,
alongside textual notes and experience archives. The same map can thus
serve both roles without being copied into a textual summary.

\paragraph{The ledger and task handovers.}
We use \emph{handover} for experience passed across a session boundary.
A task handover records a task at closure, whereas a recovery note
passes search evidence to a new attempt at the same goal
(Appendix~\ref{app:hooks}). These are separate files, as listed in
Table~\ref{tab:memory-files}.
Before requesting closure, the acting session writes $h_k$, a brief
account of the goal, visited rooms and landmarks, outcome evidence, and
unsuccessful searches. The prompt requests fewer than 120 words.
The orchestrator reads this file at the task boundary and combines it
with the claim, checking verdict, and named places,
\begin{equation}
r_k=\big(g_k,\;c_k,\;v_k,\;h_k,\;p_k\big),\qquad
\mathrm{status}(r_k)=\begin{cases}
\CERT{} & v_k=\text{complete},\\
\NOTDONE{} & v_k=\text{incomplete},\\
\UNRES{} & v_k=\text{unknown},\\
\CLAIMED{} & \text{no certification ran}.
\end{cases}
\label{eq:f-row}
\end{equation}
The full run record also retains the claim and verdict evidence, searched
regions, step count, and termination reason. Environment scores are
stored separately for evaluation and are not rendered in agent-facing
memory. A certified completion concerns the goal predicate, so the
actor's route descriptions and spatial interpretations remain its own
account.
\CLAIMED{} denotes the implementation's unchecked status, not a claim of
success. When a task ends without a model closing claim, the record uses
\textnormal{stop} and retains the termination reason. This remains
distinguishable from an explicit \textnormal{complete} claim even when
certification is disabled.

The ledger file is a compact index with one line per closed task,
containing its status, a shortened goal, and up to three marker names.
Detailed handovers and verdicts remain separate files. A later session
reads the ledger to orient its search and can open the relevant task
directory for details. This keeps the account of an individual task
addressable without copying every account into a new conversation.
The complete task record $r_k$ includes the handover and verdict,
while the ledger file provides its short index entry.
The ledger index itself grows with the number of tasks.

\begin{table}[ht]
\centering\small
\caption{Textual memory in the workspace. Each file has a designated writer.}
\label{tab:memory-files}
\setlength{\tabcolsep}{4pt}
\begin{apptab}{>{\raggedright\arraybackslash}p{4.6cm}>{\raggedright\arraybackslash}p{3.3cm}>{\raggedright\arraybackslash}p{5.2cm}}
\thd{Path} & \thd{Writer} & \thd{Content} \\
\midrule
\textnormal{run/ledger.md} & Orchestrator & Task-status index \\
\textnormal{run/tasks/$k$/goal.md} & Orchestrator & Goal and modality \\
\textnormal{run/tasks/$k$/handover.md} & Acting session & Task experience \\
\textnormal{run/tasks/$k$/lambda.md} & Acting session & Recovery handover \\
\textnormal{run/tasks/$k$/verdict.json} & Orchestrator & Judge's assessment \\
\textnormal{house/} & Consolidation session & House notes: index, overview, room notes, skills \\
\bottomrule
\end{apptab}
\end{table}

The house directory is mounted from persistent storage, whereas the run
directory holds the current sequence. Repeated recoveries receive
distinct note filenames. The permission checks of
Appendix~\ref{app:interfaces} prevent an acting session from overwriting
the ledger, judge's verdict, or consolidated house notes.

\paragraph{Places.}
A place is a marker, at most four views taken at it and bucketed by heading,
and the navigator's estimate of where it was taken. The views are
observations, the estimate is an estimate, and the name is a claim; the three
are kept apart. During spatial recovery, current views can be compared with these banked
observations to identify a familiar place.
Formally a named place is
\begin{equation}
z_i=\left(a_i,m_i,\{(I_j,\hat\xi_j)\}_{j\in B_i}\right),
\qquad |B_i|\leq4.
\label{eq:place}
\end{equation}
with $a_i$ the label, $m_i$ the marker, and $B_i$ its banked views with
their SLAM pose estimates $\hat\xi_j$.

The places form a graph,
\begin{equation}
\begin{aligned}
\Gamma&=\big(V_{\mathrm{place}},\;E_{\mathrm{conn}}\big),\\
E_{\mathrm{conn}}&=\big\{(z_i,z_j):\ z_i,z_j\ \text{marked consecutively},\ f(z_i)\neq f(z_j)\big\},
\end{aligned}
\label{eq:f-graph}
\end{equation}
where each place record carries a room label and floor association $f(z)$.
The room label defaults to the marker's name. These are attributes of
places, rather than a separate, fully connected room-and-floor graph.
Room connections described in house notes remain textual knowledge.
The graph is updated from task-boundary
marker records and the reported map layer. An edge between consecutively
registered places on different layers is a candidate connector for
planning, rather than a directly observed staircase. Graph addresses are
labels and relations, with metric geometry retained in the spatial map.

\paragraph{The house archive.}
Long-term memory is the persistent knowledge maintained for a house,
not a single file. Its textual part, the house notes, comprises four files
containing an index, a house overview, room notes, and navigation skills,
denoted together by $D_h$ in the equations. All four are written only by
the consolidation session. Saved maps, place
records, and experience archives retain their own representations.

Persistent records need not wait for consolidation. At task closure,
the orchestrator updates the enabled place and experience stores and
saves spatial state when checkpointing is enabled. Consolidation instead
summarizes what happened over a sequence of tasks. It reads the existing
house notes, ledger, and handovers to retain useful routes, failed searches,
corrections, and unresolved questions without requiring later sessions
to reread every account. In the current implementation this separate
session runs at run closure, with the house-note update
\begin{equation}
D_h^{+}=\mathcal C_\theta(D_h,L_K),
\label{eq:f-consolidate}
\end{equation}
where $L_K$ includes the task accounts accessible through their files.
The prompt asks the model to retain the distinction between certified
outcomes and the robot's own interpretations, and to add a navigation
skill only when supported by the journal. The workspace enforces file-write
permissions, while these content requirements are instructions to the model.

Consolidation has file tools but no navigation tools, and its write
permission covers only the house notes. It can summarize descriptions
of rooms and landmarks, but does not rebuild or modify map geometry.
The house notes appear in a subsequent run's required reading, while
detailed records remain separately accessible. The summarization interval
is a scheduling choice: the same operation could be applied after a task,
a group of tasks, or a longer deployment. Run closure is the invocation
point used here, not a requirement that all long-term memory be written
at once. Storing or summarizing a claim does not certify it.

\subsection{Hooks}
\label{app:hooks}

Hooks are the orchestrator's responses to boundary events. A session can
request closure or recovery through a tool, but the orchestrator decides
and carries out the transition after the hop returns. Four events
connect the reasoning session to persistent memory.

\paragraph{Task boundary.}
After closure, the orchestrator obtains the final checking verdict,
reads the handover already written by the actor, and appends the task
record. It writes the verdict file and ledger index, updates the enabled
named-place and experience stores, and saves spatial state when checkpointing
is enabled. The next
task's instruction refers to the newly available records.
With the task-boundary session policy, the transition is
\begin{equation}
(C,\,M,\,L,\,D_h)\;\mapsto\;
\big(\varnothing,\;M^{+},\;L\,\|\,r_k,\;D_h\big),
\label{eq:f-task}
\end{equation}
where $M^{+}$ includes the updated place graph. The house notes $D_h$ are
unchanged at this boundary even when other persistent records are updated.
A task ended by its execution limit follows
the same recording path, even if it has no handover or completion claim.

\paragraph{Attempt boundary.}
Recovery begins with a model request or an enabled duration fallback.
The controller checks whether a new attempt is warranted, then selects
how much of the current interpretation to reconsider. Every scope keeps
the map, markers, ledger, and archive. Broader scopes add a reassessment
of location or route to the conversation reset
(Table~\ref{tab:scopes}), so restarting does not erase the observations
that the robot has already collected.

The controller aims to choose the smallest scope consistent with the
observed failure,
\begin{equation}
u_t^*=\min_{u\in\mathcal U}
\{u:I(u)\text{ covers the beliefs implicated by the observed failure}\},
\label{eq:scope}
\end{equation}
where $I(u)$ denotes the beliefs to reconsider. The rules implement this
objective using measurable execution signals, rather than an oracle for
the cause of failure. At each hop they read
\[
z=(t_a,\,\Delta A,\,\Delta E,\,\mathrm{lost},\,\mathrm{mismatch},
\,\mathrm{first},\,u_{\mathrm{last}},\,n_r).
\]
Here $t_a$ counts turns in the attempt, $\Delta A$ is newly explored area,
and $\Delta E=\lfloor\Delta n_{\mathrm{frames}}/4\rfloor+
\Delta n_{\mathrm{markers}}$ is an observation-count proxy since the
previous checkpoint. The remaining signals indicate tracking loss, a
reported floor mismatch, a first visit, the previous recovery scope, and
the number of recoveries already used. Progress is defined as
$\mathrm{prog}(z)=[\Delta A\ge1.5\,\mathrm m^2]\vee[\Delta E\ge1]$.
New frame counts need not imply new scene content, so this test is used
as a practical guard against premature recovery.

Let $f$ indicate that an enabled fallback threshold $T_f$ has been
reached, and let $N$ be the task's recovery cap. Given a proposal $\hat u$,
the controller applies the following rules in order. A
\emph{continue} decision returns immediately.
\begin{align}
u&=\text{continue}
 &&\text{if no proposal and }\neg f,\ \text{or }n_r\ge N,\notag\\
u&=\text{continue}
 &&\text{if proposed, }\neg f,\ \text{and }t_a<t_{\min},\notag\\
u&=\text{continue}
 &&\text{if proposed, }\neg f,\ \mathrm{prog}(z)
       \wedge\neg\mathrm{lost}\wedge\neg\mathrm{mismatch},\notag\\
u&=\hat u\text{ (proposed) or attempt (fallback)},\notag\\
u&\leftarrow\max(u,\text{place})
 &&\text{if lost},\notag\\
u&\leftarrow\max(u,\text{floor})
 &&\text{if mismatch},\notag\\
u&\leftarrow\text{room}
 &&\text{if }u=\text{house and first},\notag\\
u&\leftarrow u_{\mathrm{last}}^{+}
 &&\text{if escalation is enabled, }n_r>0,\ u\le u_{\mathrm{last}},\notag\\
u&\leftarrow\min(u,u_{\max}).
\label{eq:controller}
\end{align}
The configuration supplies $T_f$, $t_{\min}$, $N$, and the maximum scope
$u_{\max}$. The notation $u^+$ denotes one rung above $u$.
Proposal confidence is recorded but is not a threshold in these rules.
When $N=1$, repeated-recovery escalation is inactive.
Floor and house scopes use the same retained map and place-graph brief.
Their distinction is the breadth of reassessment requested in the
recovery prompt, not a different physical reset.

\begin{figure}[htb]
\centering
\input{figures/fig-recovery-scopes}
\caption{\textbf{Recovery scopes and retained state.} Broader recovery
reconsiders additional beliefs while preserving accumulated observations
and task records.}
\label{fig:recovery-scopes}
\end{figure}

\begin{table}[h]
\caption{Recovery scopes. All retain the map, markers, ledger, and archive.}
\label{tab:scopes}
\centering\small
\begin{apptab}{l>{\raggedright\arraybackslash}p{7.6cm}}
\thd{Scope} & \thd{State reconsidered} \\
\midrule
\textit{continue} & No reset \\
\textit{attempt} & Active conversation \\
\textit{place} & Conversation and current location \\
\textit{room} & Conversation, location, and route \\
\textit{floor} & Conversation, location, route, and floor association \\
\textit{house} & The same beliefs as floor, with house-level reassessment \\
\bottomrule
\end{apptab}
\end{table}

Before the session is replaced, it receives one final hop with file tools
only. It writes a JSON handover at its assigned recovery-note path,
\begin{equation}
\lambda_{k,a}=\mathrm{WriteNote}_{\theta}(C_{k,a})
=\big(S,\;R,\;a^\star,\;U,\;\widetilde U,\;\hat u\big),
\label{eq:f-note}
\end{equation}
covering searched places, evidence-supported exclusions, the last reliable
anchor, untried options, uncertainties, and a suggested scope. The prompt
explicitly separates \emph{searched} from \emph{ruled out} and asks the
agent to name the observation supporting each exclusion. The session
that made the attempt therefore selects what its successor should know.
These are the actor's evidence-backed interpretations, not independently
verified exclusions. The recovery prompt discourages repeating them,
so a mistaken exclusion can also be carried into the next attempt.

The orchestrator applies the selected scope and opens a new conversation,
\begin{equation}
C_{k,a+1,0}=\mathrm{Open}\bigl(\sigma,
I_{\mathrm{recover}}(g_k,\operatorname{ref}(\lambda_{k,a}))\bigr).
\label{eq:f-reopen}
\end{equation}
The recovery instruction names the scope, any location brief, and the
note to read first. The map and other files remain accessible through
their usual tools, while the robot retains the same goal, position, and
remaining task budget. Reconsidering a pose or route belief does not
delete the stored map or reset the physical trajectory.

For scopes above \emph{attempt}, the harness collects four surrounding
views. When stored place photographs are available, a separate wake-up
session compares them with current views and returns a location brief.
Floor- and house-level recovery also provide the named-place graph.
These observations help the new attempt establish its location, while
metric tracking remains the localisation tool's responsibility.

\paragraph{Relocation.}
A relocation between tasks differs from an in-task recovery. The
environment has already moved the body to the next starting point.
The orchestrator retains the spatial state and files, collects four
surrounding views, and attaches them to the next task's opening,
\begin{equation}
C_{k,1,0}=\mathrm{Open}\bigl(\sigma,
I(g_k,\rho_k)\,\|\,I_{\mathrm{relocated}}\,\|\,
F^{\circlearrowleft}_k\bigr).
\label{eq:f-wake}
\end{equation}
This reasoning session establishes which room it is in before planning
from the retained map. No separate room-identification session is
required at this boundary. Visual recognition and the tracker's metric
relocalisation remain distinct, since recognizing a room does not by
itself determine the robot's pose.

\paragraph{Consolidation.}
At run closure, the acting session ends and a file-only consolidation
session updates the house notes by Equation~\ref{eq:f-consolidate}.
Its reading list contains the ledger, task handovers, and previous house
notes. Its write permission covers only the four house-note files, which
become available to subsequent runs.

\subsection{Verification}
\label{app:closure}

\begin{figure}[h]
\centering
\resizebox{\linewidth}{!}{\input{figures/fig-closure}}
\caption{Verification at closure. Pre-stop verification can return counter-evidence. Certification records the outcome afterward, reusing the
pre-stop verdict when the action-step count is unchanged.}
\label{fig:closure}
\end{figure}
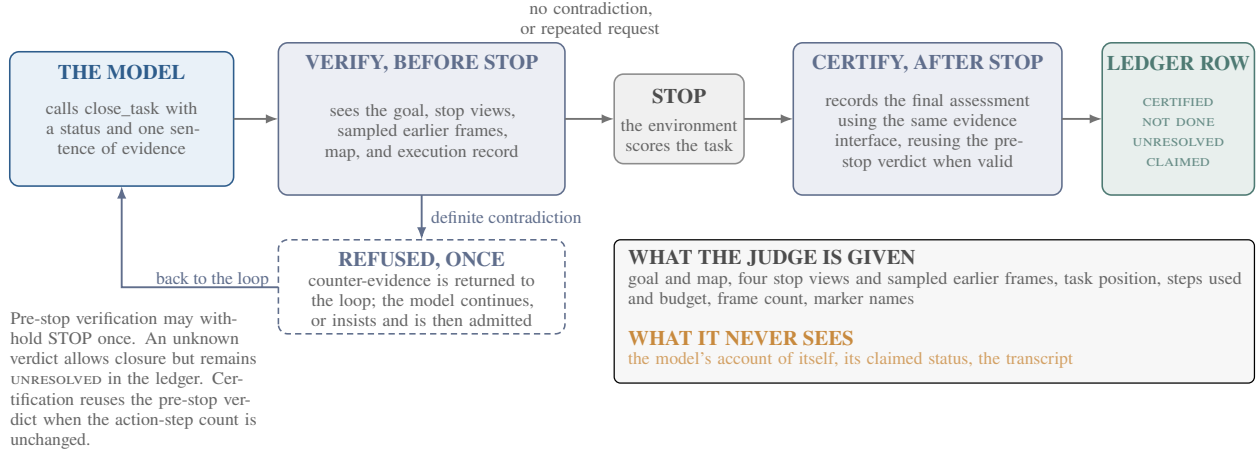

Verification is the completion-checking mechanism. It serves two roles
at closure (Figure~\ref{fig:closure}).
Pre-stop verification can return evidence to the running actor.
Post-stop certification determines the status stored in memory.
Both use the same evidence interface and, by default, the same model
as the actor, but neither receives the actor's conversation or closing
self-assessment.

\paragraph{Evidence and verdict.}
Calling \tool{close\_task} captures four instrument-selected views at the
proposed stopping point before either check runs. The judge receives the
goal, these views, a sample of earlier frames, the map, and a mechanical
record containing task position, steps used, step budget, frame count,
and marker names. The frame selector prioritizes stop views within a
twelve-frame allowance and fills the remaining slots uniformly from
the other frames, presenting the result in temporal order.
Writing $\mathcal S_{12}$ for this selector, the checks have the form
\begin{equation}
v_k^{\mathrm{pre}}=
V_\theta\bigl(g_k,\mathcal S_{12}(F_k^{\mathrm{pre}}),
M_k^{\mathrm{pre}},m_k^{\mathrm{pre}}\bigr),
\qquad
v_k=
V_\theta\bigl(g_k,\mathcal S_{12}(F_k),M_k,m_k\bigr).
\label{eq:f-verify}
\end{equation}
The superscript identifies evidence available at the first checked
request, while the second expression uses the final task evidence.
Both include the stopping views already captured by the tool.

The verdict is \emph{complete}, \emph{incomplete}, or \emph{unknown},
with an evidence statement, searched-place names, and confidence.
The judge has no action tools. If its response cannot be parsed, the
same checking conversation is prompted once for the JSON verdict alone.
A failed check or an unresolved response is recorded as unknown.
The run recorder retains this verdict separately from incomplete when
cross-tabulating verdicts with task success.

\paragraph{Pre-stop verification.}
If the handover file is missing, the orchestrator first prompts the
actor once to write it. This file check is separate from verification.
On the first request that reaches the judge, pre-stop verification withholds STOP exactly
when
\[
v_k^{\mathrm{pre}}\in\{\text{complete},\text{incomplete}\}
\quad\text{and}\quad v_k^{\mathrm{pre}}\ne c_k.
\]
It returns the evidence to the existing acting session, allowing another
observation or a revised search. An unknown verdict does not block
closure, and a subsequent request is honored without another veto.
The checked-request counter belongs to the task, survives recovery,
and resets only at the next task. The one-time missing-handover prompt
does not consume this checking opportunity.
Only the orchestrator issues the resulting STOP. This bounded check
gives the actor a chance to respond without allowing the judge to
withhold closure indefinitely.

\paragraph{Post-stop certification.}
Certification assigns the ledger status in Equation~\ref{eq:f-row}
without changing the completed task's score. If model-based pre-stop verification and
certification are both enabled and the action-step count has not changed
since pre-stop verification, the stored verdict is reused. Otherwise certification
opens a fresh checking session on the final evidence. Unknown remains
\UNRES{} in memory even though it permits the current task to close.

The two checks have separate switches because only pre-stop verification
can change the current rollout. Sharing model parameters does not make
actor and judge errors independent. Their separation instead controls
which evidence the judge sees and which records the actor may write.
Environment scores enter evaluation logs, not the model-based judge
or the agent-facing ledger. The single-view intervention in
Section~\ref{sec:exp:organs} changes the visual evidence supplied to the judge.

%% file: figures/fig-six.tex
\begin{tikzpicture}[x=1cm,y=1cm]
\definecolor{v4c}{HTML}{3A6EA5}
\draw[black!30,line width=0.5pt,rounded corners=3pt,fill=white] (0.000,0) rectangle (2.050,1.800);
\fill[v4c] (1.025,1.440) circle (0.2cm);
\node[font=\footnotesize\bfseries,text=white] at (1.025,1.440) {1};
\node[font=\small\bfseries,text=black!85,anchor=base] at (1.025,0.940) {Loop};
\node[font=\fontsize{6.5}{7.5}\selectfont,text=black!60,anchor=north,align=center] at (1.025,0.780) {opens, runs, and\\closes a task};
\draw[black!30,line width=0.5pt,rounded corners=3pt,fill=white] (2.350,0) rectangle (4.400,1.800);
\fill[v4c] (3.375,1.440) circle (0.2cm);
\node[font=\footnotesize\bfseries,text=white] at (3.375,1.440) {2};
\node[font=\small\bfseries,text=black!85,anchor=base] at (3.375,0.940) {Tools};
\node[font=\fontsize{6.5}{7.5}\selectfont,text=black!60,anchor=north,align=center] at (3.375,0.780) {the calls the\\model may make};
\draw[black!30,line width=0.5pt,rounded corners=3pt,fill=white] (4.700,0) rectangle (6.750,1.800);
\fill[v4c] (5.725,1.440) circle (0.2cm);
\node[font=\footnotesize\bfseries,text=white] at (5.725,1.440) {3};
\node[font=\small\bfseries,text=black!85,anchor=base] at (5.725,0.940) {Context};
\node[font=\fontsize{6.5}{7.5}\selectfont,text=black!60,anchor=north,align=center] at (5.725,0.780) {session openings\\and resets};
\draw[black!30,line width=0.5pt,rounded corners=3pt,fill=white] (7.050,0) rectangle (9.100,1.800);
\fill[v4c] (8.075,1.440) circle (0.2cm);
\node[font=\footnotesize\bfseries,text=white] at (8.075,1.440) {4};
\node[font=\small\bfseries,text=black!85,anchor=base] at (8.075,0.940) {Memory};
\node[font=\fontsize{6.5}{7.5}\selectfont,text=black!60,anchor=north,align=center] at (8.075,0.780) {map, markers,\\ledger, archive};
\draw[black!30,line width=0.5pt,rounded corners=3pt,fill=white] (9.400,0) rectangle (11.450,1.800);
\fill[v4c] (10.425,1.440) circle (0.2cm);
\node[font=\footnotesize\bfseries,text=white] at (10.425,1.440) {5};
\node[font=\small\bfseries,text=black!85,anchor=base] at (10.425,0.940) {Hooks};
\node[font=\fontsize{6.5}{7.5}\selectfont,text=black!60,anchor=north,align=center] at (10.425,0.780) {task, recovery,\\and run boundaries};
\draw[black!30,line width=0.5pt,rounded corners=3pt,fill=white] (11.750,0) rectangle (13.800,1.800);
\fill[v4c] (12.775,1.440) circle (0.2cm);
\node[font=\footnotesize\bfseries,text=white] at (12.775,1.440) {6};
\node[font=\small\bfseries,text=black!85,anchor=base] at (12.775,0.940) {Verification};
\node[font=\fontsize{6.5}{7.5}\selectfont,text=black!60,anchor=north,align=center] at (12.775,0.780) {verify before STOP,\\certify after};
\end{tikzpicture}

%% file: figures/fig-d-succ.tex
\begin{tikzpicture}[x=1cm,y=1cm]
\definecolor{v4c}{HTML}{3A6EA5}\definecolor{selfc}{HTML}{D08A3E}\definecolor{neverc}{HTML}{8C8C8C}
\draw[black!12,line width=0.3pt] (0,0.000) -- (6.4,0.000); \node[font=\scriptsize,text=black!50,anchor=east] at (-0.1,0.000) {0};
\draw[black!12,line width=0.3pt] (0,0.800) -- (6.4,0.800); \node[font=\scriptsize,text=black!50,anchor=east] at (-0.1,0.800) {25\%};
\draw[black!12,line width=0.3pt] (0,1.600) -- (6.4,1.600); \node[font=\scriptsize,text=black!50,anchor=east] at (-0.1,1.600) {50\%};
\draw[black!12,line width=0.3pt] (0,2.400) -- (6.4,2.400); \node[font=\scriptsize,text=black!50,anchor=east] at (-0.1,2.400) {75\%};
\draw[black!12,line width=0.3pt] (0,3.200) -- (6.4,3.200); \node[font=\scriptsize,text=black!50,anchor=east] at (-0.1,3.200) {100\%};
\draw[black!45,line width=0.4pt] (0,0) -- (6.4,0);
\node[font=\scriptsize,text=black!50,anchor=north] at (0.000,-0.05) {0};
\node[font=\scriptsize,text=black!50,anchor=north] at (1.280,-0.05) {2};
\node[font=\scriptsize,text=black!50,anchor=north] at (2.560,-0.05) {4};
\node[font=\scriptsize,text=black!50,anchor=north] at (3.840,-0.05) {6};
\node[font=\scriptsize,text=black!50,anchor=north] at (5.120,-0.05) {8};
\node[font=\scriptsize,text=black!50,anchor=north] at (6.400,-0.05) {10};
\node[font=\scriptsize,text=black!60,anchor=north] at (3.20,-0.32) {task position in the sequence};
\draw[selfc,line width=1.3pt,smooth] plot coordinates {(0.640,2.208) (0.960,2.117) (1.280,2.033) (1.599,1.955) (1.921,1.884) (2.239,1.818) (2.561,1.757) (2.880,1.700) (3.199,1.648) (3.520,1.600) (3.840,1.556) (4.160,1.515) (4.479,1.477) (4.799,1.442) (5.119,1.410) (5.439,1.381) (5.760,1.353) (6.079,1.328) (6.400,1.305)};
\node[font=\scriptsize\bfseries,text=selfc,anchor=west] at (6.55,1.305) {Single session};
\draw[v4c,line width=1.5pt,smooth] plot coordinates {(0.640,2.294) (0.960,2.428) (1.280,2.497) (1.599,2.532) (1.921,2.550) (2.239,2.559) (2.561,2.564) (2.880,2.566) (3.199,2.568) (3.520,2.568) (3.840,2.569) (4.160,2.569) (4.479,2.569) (4.799,2.569) (5.119,2.569) (5.439,2.569) (5.760,2.569) (6.079,2.569) (6.400,2.569)};
\node[font=\scriptsize\bfseries,text=v4c,anchor=west] at (6.55,2.569) {NavHarness};
\end{tikzpicture}

%% file: figures/fig-d-ctx.tex
\begin{tikzpicture}[x=1cm,y=1cm]
\definecolor{v4c}{HTML}{3A6EA5}\definecolor{sessionc}{HTML}{8C8C8C}
\foreach \y/\label in {0/0,.762/250k,1.524/500k,2.286/750k,3.048/1M}{
  \draw[black!12,line width=.3pt] (0,\y)--(6.4,\y);
  \node[font=\scriptsize,text=black!50,anchor=east] at (-.1,\y) {\label};
}
\draw[sessionc,densely dashed,line width=.6pt] (0,3.048)--(6.4,3.048);
\node[font=\scriptsize,text=black!65,anchor=south east] at (6.4,3.12) {Window full: compact, then continue};
\draw[black!45,line width=.4pt,-{Stealth[length=3pt]}] (0,0)--(6.85,0);
\foreach \k in {0,5,10,15,20}{
  \draw[black!45,line width=.4pt] ({.32*\k},0)--({.32*\k},-.04);
  \node[font=\scriptsize,text=black!50,anchor=north] at ({.32*\k},-.07) {\k};
}
\node[font=\scriptsize,text=black!60,anchor=north] at (3.2,-.36) {Completed tasks};
\draw[sessionc,line width=1.2pt] (0,0.061)--(6.4,3.048);
\fill[sessionc] (6.4,3.048) circle[radius=1.5pt];
\draw[v4c,line width=1.4pt] (0,.061)--(6.4,.061);
\end{tikzpicture}

%% file: figures/fig-d-cost.tex
\begin{tikzpicture}[x=1cm,y=1cm]
\definecolor{v4c}{HTML}{3A6EA5}\definecolor{sessionc}{HTML}{8C8C8C}\definecolor{statelessc}{HTML}{D08A3E}
\foreach \cost in {0,500,1000,1500,2000,2500}{
  \draw[black!12,line width=.3pt] (0,{\cost*3.2/2700})--(6.4,{\cost*3.2/2700});
  \node[font=\scriptsize,text=black!50,anchor=east] at (-.1,{\cost*3.2/2700}) {\$\cost};
}
\draw[black!45,line width=.4pt,-{Stealth[length=3pt]}] (0,0)--(6.85,0);
\foreach \k in {0,5,10,15,20}{
  \draw[black!45,line width=.4pt] ({.32*\k},0)--({.32*\k},-.04);
  \node[font=\scriptsize,text=black!50,anchor=north] at ({.32*\k},-.07) {\k};
}
\node[font=\scriptsize,text=black!60,anchor=north] at (3.2,-.36) {Completed tasks};
\draw[statelessc,line width=1.2pt,smooth] plot coordinates {(0.000,0.000) (0.320,0.014) (0.640,0.043) (0.960,0.085) (1.280,0.142) (1.600,0.213) (1.920,0.299) (2.240,0.398) (2.560,0.512) (2.880,0.640) (3.200,0.782) (3.520,0.939) (3.840,1.109) (4.160,1.294) (4.480,1.493) (4.800,1.707) (5.120,1.934) (5.440,2.176) (5.760,2.432) (6.080,2.702) (6.400,2.987)};
\draw[sessionc,line width=1.2pt] plot coordinates {(0.000,0.000) (0.320,0.031) (0.640,0.063) (0.960,0.094) (1.280,0.125) (1.600,0.156) (1.920,0.188) (2.240,0.219) (2.560,0.250) (2.880,0.282) (3.200,0.313) (3.520,0.344) (3.840,0.375) (4.160,0.407) (4.480,0.438) (4.800,0.469) (5.120,0.501) (5.440,0.532) (5.760,0.563) (6.080,0.594) (6.400,0.626)};
\draw[v4c,line width=1.4pt] plot coordinates {(0.000,0.000) (0.320,0.006) (0.640,0.012) (0.960,0.017) (1.280,0.023) (1.600,0.029) (1.920,0.035) (2.240,0.040) (2.560,0.046) (2.880,0.052) (3.200,0.058) (3.520,0.064) (3.840,0.069) (4.160,0.075) (4.480,0.081) (4.800,0.087) (5.120,0.092) (5.440,0.098) (5.760,0.104) (6.080,0.110) (6.400,0.116)};
\node[font=\scriptsize,text=statelessc,anchor=south east] at (6.4,3.08) {$\sim\$2{,}520$};
\node[font=\scriptsize,text=sessionc,anchor=south east] at (6.4,.70) {$\sim\$528$};
\node[font=\scriptsize,text=v4c,anchor=south east] at (6.4,.16) {$\sim\$98$};
\end{tikzpicture}

%% file: figures/fig-recovery-scopes.tex
\begin{tikzpicture}[x=1cm,y=1cm,
 every node/.style={font=\scriptsize,inner sep=1pt}]
\definecolor{scopekeep}{HTML}{E3EEE9}
\definecolor{scopereset}{HTML}{F6E4D4}
\definecolor{scopereload}{HTML}{E3E8F5}
\node[align=center] at (2.15,0.40) {Context};
\node[align=center] at (3.54,0.40) {Pose belief};
\node[align=center] at (4.93,0.40) {Route belief};
\node[align=center] at (6.32,0.40) {Floor belief};
\node[align=center] at (7.71,0.40) {Map};
\node[align=center] at (9.10,0.40) {Markers};
\node[align=center] at (10.49,0.40) {Ledger};
\node[align=center] at (11.88,0.40) {Archive};
\node[anchor=east] at (1.37,0.00) {\textit{continue}};
\draw[fill=scopekeep,draw=white] (1.49,-0.19) rectangle (2.81,0.19);
\node[text=black] at (2.15,0.00) {retain};
\draw[fill=scopekeep,draw=white] (2.88,-0.19) rectangle (4.20,0.19);
\node[text=black] at (3.54,0.00) {retain};
\draw[fill=scopekeep,draw=white] (4.27,-0.19) rectangle (5.59,0.19);
\node[text=black] at (4.93,0.00) {retain};
\draw[fill=scopekeep,draw=white] (5.66,-0.19) rectangle (6.98,0.19);
\node[text=black] at (6.32,0.00) {retain};
\draw[fill=scopekeep,draw=white] (7.05,-0.19) rectangle (8.37,0.19);
\node[text=black] at (7.71,0.00) {retain};
\draw[fill=scopekeep,draw=white] (8.44,-0.19) rectangle (9.76,0.19);
\node[text=black] at (9.10,0.00) {retain};
\draw[fill=scopekeep,draw=white] (9.83,-0.19) rectangle (11.15,0.19);
\node[text=black] at (10.49,0.00) {retain};
\draw[fill=scopekeep,draw=white] (11.22,-0.19) rectangle (12.54,0.19);
\node[text=black] at (11.88,0.00) {retain};
\node[anchor=east] at (1.37,-0.46) {\textit{attempt}};
\draw[fill=scopereset,draw=white] (1.49,-0.65) rectangle (2.81,-0.27);
\node[text=black] at (2.15,-0.46) {replace};
\draw[fill=scopekeep,draw=white] (2.88,-0.65) rectangle (4.20,-0.27);
\node[text=black] at (3.54,-0.46) {retain};
\draw[fill=scopekeep,draw=white] (4.27,-0.65) rectangle (5.59,-0.27);
\node[text=black] at (4.93,-0.46) {retain};
\draw[fill=scopekeep,draw=white] (5.66,-0.65) rectangle (6.98,-0.27);
\node[text=black] at (6.32,-0.46) {retain};
\draw[fill=scopekeep,draw=white] (7.05,-0.65) rectangle (8.37,-0.27);
\node[text=black] at (7.71,-0.46) {retain};
\draw[fill=scopekeep,draw=white] (8.44,-0.65) rectangle (9.76,-0.27);
\node[text=black] at (9.10,-0.46) {retain};
\draw[fill=scopekeep,draw=white] (9.83,-0.65) rectangle (11.15,-0.27);
\node[text=black] at (10.49,-0.46) {retain};
\draw[fill=scopekeep,draw=white] (11.22,-0.65) rectangle (12.54,-0.27);
\node[text=black] at (11.88,-0.46) {retain};
\node[anchor=east] at (1.37,-0.92) {\textit{place}};
\draw[fill=scopereset,draw=white] (1.49,-1.11) rectangle (2.81,-0.73);
\node[text=black] at (2.15,-0.92) {replace};
\draw[fill=scopereset,draw=white] (2.88,-1.11) rectangle (4.20,-0.73);
\node[text=black] at (3.54,-0.92) {reassess};
\draw[fill=scopekeep,draw=white] (4.27,-1.11) rectangle (5.59,-0.73);
\node[text=black] at (4.93,-0.92) {retain};
\draw[fill=scopekeep,draw=white] (5.66,-1.11) rectangle (6.98,-0.73);
\node[text=black] at (6.32,-0.92) {retain};
\draw[fill=scopekeep,draw=white] (7.05,-1.11) rectangle (8.37,-0.73);
\node[text=black] at (7.71,-0.92) {retain};
\draw[fill=scopekeep,draw=white] (8.44,-1.11) rectangle (9.76,-0.73);
\node[text=black] at (9.10,-0.92) {retain};
\draw[fill=scopekeep,draw=white] (9.83,-1.11) rectangle (11.15,-0.73);
\node[text=black] at (10.49,-0.92) {retain};
\draw[fill=scopekeep,draw=white] (11.22,-1.11) rectangle (12.54,-0.73);
\node[text=black] at (11.88,-0.92) {retain};
\node[anchor=east] at (1.37,-1.38) {\textit{room}};
\draw[fill=scopereset,draw=white] (1.49,-1.57) rectangle (2.81,-1.19);
\node[text=black] at (2.15,-1.38) {replace};
\draw[fill=scopereset,draw=white] (2.88,-1.57) rectangle (4.20,-1.19);
\node[text=black] at (3.54,-1.38) {reassess};
\draw[fill=scopereset,draw=white] (4.27,-1.57) rectangle (5.59,-1.19);
\node[text=black] at (4.93,-1.38) {reassess};
\draw[fill=scopekeep,draw=white] (5.66,-1.57) rectangle (6.98,-1.19);
\node[text=black] at (6.32,-1.38) {retain};
\draw[fill=scopekeep,draw=white] (7.05,-1.57) rectangle (8.37,-1.19);
\node[text=black] at (7.71,-1.38) {retain};
\draw[fill=scopekeep,draw=white] (8.44,-1.57) rectangle (9.76,-1.19);
\node[text=black] at (9.10,-1.38) {retain};
\draw[fill=scopekeep,draw=white] (9.83,-1.57) rectangle (11.15,-1.19);
\node[text=black] at (10.49,-1.38) {retain};
\draw[fill=scopekeep,draw=white] (11.22,-1.57) rectangle (12.54,-1.19);
\node[text=black] at (11.88,-1.38) {retain};
\node[anchor=east] at (1.37,-1.84) {\textit{floor}};
\draw[fill=scopereset,draw=white] (1.49,-2.03) rectangle (2.81,-1.65);
\node[text=black] at (2.15,-1.84) {replace};
\draw[fill=scopereset,draw=white] (2.88,-2.03) rectangle (4.20,-1.65);
\node[text=black] at (3.54,-1.84) {reassess};
\draw[fill=scopereset,draw=white] (4.27,-2.03) rectangle (5.59,-1.65);
\node[text=black] at (4.93,-1.84) {reassess};
\draw[fill=scopereset,draw=white] (5.66,-2.03) rectangle (6.98,-1.65);
\node[text=black] at (6.32,-1.84) {reassess};
\draw[fill=scopekeep,draw=white] (7.05,-2.03) rectangle (8.37,-1.65);
\node[text=black] at (7.71,-1.84) {retain};
\draw[fill=scopekeep,draw=white] (8.44,-2.03) rectangle (9.76,-1.65);
\node[text=black] at (9.10,-1.84) {retain};
\draw[fill=scopekeep,draw=white] (9.83,-2.03) rectangle (11.15,-1.65);
\node[text=black] at (10.49,-1.84) {retain};
\draw[fill=scopekeep,draw=white] (11.22,-2.03) rectangle (12.54,-1.65);
\node[text=black] at (11.88,-1.84) {retain};
\node[anchor=east] at (1.37,-2.30) {\textit{house}};
\draw[fill=scopereset,draw=white] (1.49,-2.49) rectangle (2.81,-2.11);
\node[text=black] at (2.15,-2.30) {replace};
\draw[fill=scopereset,draw=white] (2.88,-2.49) rectangle (4.20,-2.11);
\node[text=black] at (3.54,-2.30) {reassess};
\draw[fill=scopereset,draw=white] (4.27,-2.49) rectangle (5.59,-2.11);
\node[text=black] at (4.93,-2.30) {reassess};
\draw[fill=scopereset,draw=white] (5.66,-2.49) rectangle (6.98,-2.11);
\node[text=black] at (6.32,-2.30) {reassess};
\draw[fill=scopekeep,draw=white] (7.05,-2.49) rectangle (8.37,-2.11);
\node[text=black] at (7.71,-2.30) {retain};
\draw[fill=scopekeep,draw=white] (8.44,-2.49) rectangle (9.76,-2.11);
\node[text=black] at (9.10,-2.30) {retain};
\draw[fill=scopekeep,draw=white] (9.83,-2.49) rectangle (11.15,-2.11);
\node[text=black] at (10.49,-2.30) {retain};
\draw[fill=scopekeep,draw=white] (11.22,-2.49) rectangle (12.54,-2.11);
\node[text=black] at (11.88,-2.30) {retain};
\node[anchor=west,font=\scriptsize] at (1.49,-2.85)
{All recovery scopes preserve the physical trajectory and accumulated task budget.};
\end{tikzpicture}

%% file: figures/fig-closure.tex
\begin{tikzpicture}[x=1cm,y=1cm, every node/.style={inner sep=0pt},
  capz/.style={font=\fontsize{6.6}{6.6}\selectfont\bfseries},
  tny/.style={font=\fontsize{5.6}{6.4}\selectfont, text=black!60, align=center},
  flow/.style={-{Latex[length=1.4mm]}, black!55, line width=0.6pt},
  vflow/.style={-{Latex[length=1.4mm]}, jd, line width=0.6pt}]
\definecolor{hz}{RGB}{46,93,140}\definecolor{hzt}{RGB}{233,241,248}
\definecolor{jd}{RGB}{96,110,140}\definecolor{jdt}{RGB}{236,238,244}
\definecolor{mem}{RGB}{78,122,115}\definecolor{memt}{RGB}{232,239,237}
\definecolor{rec}{RGB}{200,138,62}
\draw[rounded corners=3pt,fill=hzt,draw=hz,line width=0.5pt] (0,2.55) rectangle (2.5,4.05);
\node[capz,text=hz] at (1.25,3.82) {THE MODEL};
\node[tny,text width=2.2cm] at (1.25,3.2) {calls \textnormal{close\_task} with a status and one sentence of evidence};
\draw[flow] (2.5,3.3) -- (3.0,3.3);
\draw[rounded corners=3pt,fill=jdt,draw=jd,line width=0.5pt] (3.0,2.45) rectangle (6.2,4.15);
\node[capz,text=jd] at (4.6,3.92) {VERIFY, BEFORE STOP};
\node[tny,text width=2.9cm] at (4.6,3.15) {sees the goal, stop views, sampled earlier frames, map, and execution record};
\draw[flow] (6.2,3.3) -- (6.75,3.3);
\node[tny,anchor=south,text width=1.6cm] at (6.48,4.2) {no contradiction, or repeated request};
\draw[rounded corners=3pt,fill=black!6,draw=black!45,line width=0.5pt] (6.75,2.8) rectangle (8.2,3.8);
\node[capz,text=black!70] at (7.475,3.55) {STOP};
\node[tny,text width=1.3cm] at (7.475,3.12) {the environment scores the task};
\draw[flow] (8.2,3.3) -- (8.75,3.3);
\draw[rounded corners=3pt,fill=jdt,draw=jd,line width=0.5pt] (8.75,2.45) rectangle (11.75,4.15);
\node[capz,text=jd] at (10.25,3.92) {CERTIFY, AFTER STOP};
\node[tny,text width=2.75cm] at (10.25,3.15) {records the final assessment using the same evidence interface, reusing the pre-stop verdict when valid};
\draw[flow] (11.75,3.3) -- (12.2,3.3);
\draw[rounded corners=3pt,fill=memt,draw=mem,line width=0.5pt] (12.2,2.45) rectangle (13.9,4.15);
\node[capz,text=mem] at (13.05,3.92) {LEDGER ROW};
\node[tny,text=mem!85,text width=1.5cm] at (13.05,3.15) {\textsc{certified}\\\textsc{not done}\\\textsc{unresolved}\\\textsc{claimed}};
\draw[vflow] (4.6,2.45) -- (4.6,1.95);
\node[tny,text=jd,anchor=west] at (4.7,2.2) {definite contradiction};
\draw[rounded corners=3pt,fill=white,draw=jd,line width=0.5pt,dash pattern=on 2pt off 1.3pt] (3.0,0.9) rectangle (6.2,1.95);
\node[capz,text=jd] at (4.6,1.72) {REFUSED, ONCE};
\node[tny,text width=2.9cm] at (4.6,1.28) {counter-evidence is returned to the loop; the model continues, or insists and is then admitted};
\draw[vflow] (3.0,1.42) -| (1.25,2.55);
\node[tny,text=jd,anchor=east] at (2.9,1.5) {back to the loop};
\node[tny,anchor=north west,align=left,text width=2.75cm] at (0,1.15) {Pre-stop verification may withhold STOP once. An unknown verdict allows closure but remains \textsc{unresolved} in the ledger. Certification reuses the pre-stop verdict when the action-step count is unchanged.};
\draw[rounded corners=2pt,fill=black!3] (6.75,0.35) rectangle (13.9,1.95);
\node[capz,text=black!70,anchor=west] at (6.9,1.75) {WHAT THE JUDGE IS GIVEN};
\node[tny,anchor=north west,align=left,text width=6.9cm] at (6.9,1.6) {goal and map, four stop views and sampled earlier frames, task position, steps used and budget, frame count, marker names};
\node[capz,text=rec,anchor=west] at (6.9,0.85) {WHAT IT NEVER SEES};
\node[tny,anchor=north west,align=left,text width=6.9cm,text=rec!80] at (6.9,0.7) {the model's account of itself, its claimed status, the transcript};
\end{tikzpicture}

%% file: sections/app_benchmarks.tex
Table~\ref{tab:validation-splits} summarizes our validation-unseen evaluations.
The GOAT-Bench main comparisons across all four executors and the Opus 5
session and component study use the full Val-Unseen split, comprising
2,669 subtasks in 360 episodes across 36 scenes.
\begin{table}[ht]
\caption{Evaluation scope on validation-unseen data.}
\label{tab:validation-splits}
\centering\small
\begin{apptab}{@{}lrrrr@{}}
Benchmark & Scenes & Episodes / tours & Tasks & Tasks per sequence \\
\midrule
GOAT-Bench & 36 & 360 episodes & 2,669 & 5--10 \\
IR2R-CE & 11 & 36 tours & 1,824 & 3--100 \\
\bottomrule
\end{apptab}
\end{table}

\subsection{GOAT-Bench}
\label{app:goat-protocol}
GOAT-Bench evaluates lifelong multimodal navigation in Habitat \citep{habitat}
on HM3D scans \citep{hm3d}. An episode places the agent in one house and gives it
a sequence of goals to reach in turn, and the agent is not reset between
them. Each goal is specified in one of three ways: as an object category, as a
natural-language description that picks out one instance, or as an image of
that instance. A goal counts as reached when the agent stops within the
success radius of a navigable view point of the target instance, and
efficiency is scored by SPL against the shortest path. We evaluate all
360 validation-unseen episodes across 36 scenes, retaining all 2,669 subtasks.
The full-split comparisons use identical evaluation tasks across executors
and their independent-session baselines in Table~\ref{tab:goat-main}, and
across all arms of the Opus 5 study in Table~\ref{tab:organs-ablation}.
Each GOAT-Bench episode and each IR2R-CE tour is a separate benchmark run,
starting with empty maps, task records, and house notes. Memory persists
between tasks of that run but is not inherited by another run, even in
the same scene. Run-end notes therefore provide no input to these evaluations.
In our protocol the agent's position, map, and
markers come from the navigator's own visual SLAM rather than from the
simulator, and simulator truth is used only to score.

\subsection{IR2R-CE}
We evaluate all 36 tours in the standard IR2R-CE validation-unseen split,
retaining all 1,824 tasks across its 11 scenes.
IR2R-CE builds on VLN-CE \citep{vlnce} in Matterport3D \citep{mp3d}
and evaluates instruction following over tours. A tour joins
consecutive Room-to-Room episodes from one scene in the continuous
environment, and the agent keeps whatever it has learned from one episode
into the next, even when transferred to the next episode's start. Each
episode is scored as in R2R-CE by success and SPL, and the tour as a whole by
t-nDTW, the normalised dynamic time warping between the tour's driven path
and its reference. We use the validation-unseen split, as for GOAT-Bench.
We compute t-nDTW with the benchmark's standard trajectory handling and
scoring procedure, using the same convention as the comparison methods.

\subsection{Comparison sources and settings}
\label{app:comparison-provenance}
The Base Model column in Table~\ref{tab:goat-main} identifies the main
language or vision-language model used for the reported navigation result.
It does not list every perception or mapping module. For example,
ObsGraph additionally uses GPT-4o-mini to construct its room hierarchy
\citep{obsgraph}, while AstraNav-Memory augments Qwen2.5-VL-3B with a
DINOv3 visual encoder \citep{astranavmem}.
The published 3D-Mem GOAT result uses GPT-4o to select targets from
memory and frontier snapshots, with movement handled by Habitat's
pathfinder \citep{threedmem}. GSMem also uses GPT-4o for GOAT reasoning,
but combines it with Gaussian scene reconstruction, retrieval-rendering,
and exploration planning \citep{gsmem}. Matching the reasoning-model
name would leave these perception and control pipelines different.
Our same-backbone controls instead retain the executor and navigation
interface while changing the surrounding harness. Their gains measure
the full system change, including memory, recovery, and completion checks.

Prior rows in the main tables retain the source evaluation protocols.
The dagger in Table~\ref{tab:goat-main} identifies the 278-subtask
evaluation used by the marked methods, including our GPT-4o subset result.
Our unmarked rows and the same-backbone gains use the full split.
We report 3D-Mem's
subset result of 69.1 s-SR and 48.9 SPL, rather than its full-set result
of 62.9 and 44.7 \citep{threedmem}. MemoryExplorer reports the same
36-scene, one-episode-per-scene protocol \citep{lmee}.
GSMem describes the validation-unseen benchmark with ten episodes per
scene \citep{gsmem}.
The GT-pose columns in Tables~\ref{tab:goat-main} and~\ref{tab:ivln-main}
identify methods that use simulator-provided localization.
The CMA, TourCMA, PoolCMA, and PoolEndCMA policies instead use RGB-D,
instructions, and action history without a pose input \citep{ivln}.
Our navigator uses visual SLAM pose, while simulator truth is reserved
for evaluation. Table~\ref{tab:goat-additional} retains additional
published baselines omitted from the main table for space.

\begin{table}[ht]
\caption{Additional published GOAT-Bench Val-Unseen results.
All listed methods use simulator-provided pose.}
\label{tab:goat-additional}
\centering\small
\begin{apptab}{@{}>{\raggedright\arraybackslash}p{0.68\linewidth}lrr@{}}
Method & Type & s-SR & SPL \\
\midrule
SenseAct-NN Monolithic \citep{goatbench} & trained & 12.3 & 6.8 \\
Modular CLIP on Wheels \citep{cow,goatbench} & zero-shot & 16.1 & 10.4 \\
VLMnav \citep{vlmnav,dynavlm} & zero-shot & 20.1 & 9.6 \\
Modular GOAT \citep{goatbench} & zero-shot & 24.9 & 17.2 \\
DyNaVLM \citep{dynavlm} & zero-shot & 25.5 & 10.2 \\
SenseAct-NN Skill Chain \citep{goatbench} & trained & 29.5 & 11.3 \\
TANGO \citep{tango} & zero-shot & 32.1 & 16.5 \\
AnyGoal, single agent \citep{anygoal} & zero-shot & 41.9 & 14.4 \\
MTU3D \citep{mtu3d} & trained & 47.2 & 27.7 \\
\bottomrule
\end{apptab}
\end{table}

GOAT e-SR follows the all-subtasks criterion of \citet{ssmgnav}.
The original Monolithic baseline reports 12.3 s-SR and 6.8 SPL without
e-SR. The 0.8 e-SR reported by \citet{ssmgnav} belongs to their
reimplementation, with 12.6 s-SR and 8.8 SPL, and is not combined with
the original row.
For IR2R-CE, CMA variants and MAP-CMA come from \citet{ivln}.
OVER-NAV uses its continuous-environment result \citep{overnav}.
ETPNav and HNR use the IR2R-CE evaluations in Table~2 of
\citet{seqwalker}, not their original single-episode R2R-CE scores.
SeqWalker also uses that table, not its distinct SH IR2R-CE benchmark.
Table~\ref{tab:ivln-additional} retains three CMA variants omitted from
the main table for space.

\begin{table}[ht]
\caption{Additional published IR2R-CE Val-Unseen results \citep{ivln}.}
\label{tab:ivln-additional}
\centering\small
\begin{apptab}{@{}llcrrrr@{}}
Method & Type & GT pose & s-SR & SPL & e-SR & t-nDTW \\
\midrule
TourCMA & trained & $\times$ & 18 & 17 & -- & 36 \\
PoolCMA & trained & $\times$ & 16 & 15 & -- & 36 \\
PoolEndCMA & trained & $\times$ & 18 & 16 & -- & 38 \\
\bottomrule
\end{apptab}
\end{table}

We evaluate NavHarness with Qwen3.8-27B, GPT-4o, Opus 5, and GPT-6 Astra.
All four executors use the same mapping, memory, and recovery mechanisms.
Following MIP \citep{mip}, the reasoning core is a multi-round coding-agent
session. GPT-6 Astra uses Codex with medium reasoning effort, Opus 5 uses
Claude Code, and both Qwen3.8-27B and GPT-4o use mini-SWE-agent. NavHarness manages the
task and recovery boundaries around these sessions rather than replacing
their inner reasoning loops. Controlled component comparisons use Opus 5
throughout.

\paragraph{Models assigned to auxiliary roles.}
The standard launcher routes every role to one model.
Pre-stop verification and post-stop certification use the same judge model.
The acting session writes its recovery handover before replacement.
Consolidation uses a separate session with the same model when enabled.
Except in the judge-swap comparison in Appendix~\ref{app:supplementary-ablations},
one model fills all auxiliary roles as well as navigation, keeping model
assignment fixed within each same-backbone comparison.
None of these roles receives evaluator success labels.
The saved Opus and Astra GOAT runs confirm same-model judging.
The independent-session baseline has no judge or cross-task consolidation.

\paragraph{Default stopping procedure.}
The benchmark configuration checks the first stopping request and certifies
the completed task (\texttt{precheck=model}, \texttt{certify=model}).
``No post-stop certification'' in Table~\ref{tab:organs-ablation} removes only
the post-stop verdict and leaves the stopping procedure unchanged.
The independent-session baseline omits both checks. The additional
independent-session-plus-verification control enables pre-stop verification.
The full-system pre-stop verification ablation uses \texttt{precheck=none} while retaining
\texttt{certify=model}.

\paragraph{The two baseline sessions.}
Independent sessions use a fresh coding session for each task, with only
observation and action tools. The goal, observations, reasoning, and tool
responses remain in the current conversation. The session has no access
to external working or long-term memory, including maps, markers, ledger
entries, or handover files.
Single session continues that conversation from one task to the next,
with backend compaction when its context window fills
(Appendix~\ref{app:context}). The session baselines use the same task
budget as the full harness and compare complete system policies.
The reference configuration uses full-text ledger access (\texttt{ledger=text}).
\begin{humanprompted}
\paragraph{Method labels.}
The Type labels in the main tables are descriptive rather than mutually
exclusive method classes. \emph{Agentic} identifies tool-mediated model
reasoning, \emph{trained} and \emph{zero-shot} describe task-specific
training, and \emph{hybrid} denotes a combination of learned navigation
and higher-level reasoning. A system can satisfy more than one of these
descriptions. The labels do not establish backbone equivalence.

\paragraph{Task budgets.}
Table~\ref{tab:budget-accounting} gives the common navigation budgets
for the GOAT controlled comparisons. Recovery shares the task's remaining
allowances. The motion interface is specified in Appendix~\ref{app:interfaces},
and session management in Appendix~\ref{app:context}.

\begin{table}[ht]
\caption{Budget accounting for the controlled GOAT comparisons.}
\label{tab:budget-accounting}
\centering\small
\begin{apptab}{@{}lp{8.8cm}@{}}
Quantity & Limit or counting rule \\
\midrule
Movement budget & 500 executed forward or turning actions per task \\
Model-turn budget & 200 navigation-model turns per task \\
Action batching & Each executed action is charged individually \\
Recovery allowance & At most one recovery, sharing the remaining task budgets \\
Fallback trigger & 80 turns, without resetting either budget \\
Observation and memory & No movement-action charge \\
\bottomrule
\end{apptab}
\end{table}
\end{humanprompted}

\paragraph{Runtime and context footprint.}
\label{app:runtime}
In GOAT-Bench Val-Unseen profiling runs, Opus 5 and GPT-6 Astra have amortized
wall times of 129.5\,s and 67.8\,s per task, respectively. We sum the elapsed
times of the evaluated scene runs and divide by the number of tasks, including setup, navigation,
and boundary processing. The mean per-task peak prompt sizes
are 20.7k and 23.2k tokens, respectively. These describe context occupancy.
End-to-end wall time includes recorded auxiliary processing.
The input-cost illustration in Appendix~\ref{app:context} uses stated
volume and price assumptions, rather than a measured API bill.

\subsection{Statistics}
\label{app:stats}
The GOAT-Bench full-harness and independent-session comparisons for all
four executors and the Opus 5 session and component study use three seeds
on the same 2,669 subtasks in 360 validation-unseen episodes across 36 scenes.
Tables~\ref{tab:goat-main}
and~\ref{tab:organs-ablation} report mean scores and standard deviations
across runs. The IR2R-CE NavHarness and independent-session results likewise
report means and standard deviations over three seeds for all four executors
(Table~\ref{tab:ivln-main}).
The Opus GOAT-Bench comparisons pair shared tasks by scene
and episode identifier, counting each task once per run. Paired 95\%
confidence intervals pool the repetitions and use 5{,}000 scene-clustered
bootstrap resamples, keeping paired outcomes together. These are
pointwise intervals for component analysis, without multiple-comparison
adjustment.
For continuous deployment (Figure~\ref{fig:deployment-progress-preview}), scores and
paired changes pool the tasks across all 36 houses. The 95\% CIs use houses
as the resampling unit, keeping the Full and No consolidation results paired.

\subsection{Chain success}
Chain success is the fraction of sequences in which every task succeeded: an
episode on GOAT-Bench, a tour on IR2R-CE. It is the quantity that a
deployment cares about and that task success hides, since one failure
anywhere in the sequence fails the chain.

\subsection{Interventions}
\label{app:defensive}
Table~\ref{tab:organs-ablation} separates system comparisons (A) from
component interventions (B--D). All use Opus 5 and the same ordered
2,669-subtask full-split evaluation. The reference retains maps and task records,
uses structured recovery handovers, and enables both pre-stop verification
and post-stop certification with four stop views. Each B--D row starts
from this reference independently. No row inherits the changes of the
preceding row. Differences and CIs are always relative to full, including
the independent-session-plus-verification row.

\paragraph{A. Complete session policies.}
Independent sessions start a context-only coding session for each task,
without external memory even within a task. Single session continues
the same conversation across tasks as defined in Appendix~\ref{app:context}.
Independent sessions + pre-stop verification adds the stopping check to this policy without adding memory. Comparing
this control with full tests the remaining harness features jointly while
holding pre-stop verification enabled. None of these rows is a memory-only ablation.
Comparing 62.4 with 58.9 describes a different contrast from the paired
change in Table~\ref{tab:organs-ci}, which compares 62.4 with
the full-system 81.5.

\paragraph{B. Cross-task memory.}
Clear map after each task (keep records) removes the map and spatial markers while
retaining task records and the ability to map during each task. Unlike the
context-only independent sessions, this control still uses working memory.
Hide earlier task records (keep map) keeps the map but withholds textual
history from earlier tasks, including the ledger, previous handovers,
and historical task files. These records are neither supplied
in the opening context nor accessible through file tools, although they may
still be saved for evaluation. The current task can write its own records and
pass recovery notes between attempts. These interventions ask whether spatial
knowledge and records of earlier attempts can substitute for each other.
Recovery, structured handovers, and both verification functions remain enabled.
The additional No cross-task memory control combines these two
interventions. Every task starts without an inherited map, spatial markers,
ledger entries, or handovers, but can build and consult its own
map and records, including recovery notes passed between attempts of that
task. Task budgets and both completion checks are unchanged. Its standard
benchmark results appear in Table~\ref{tab:organs-ablation}. The extended
deployment in Section~\ref{sec:exp:day} uses a different intervention:
No consolidation retains maps and task records with the same persistence
rules as full, but disables the run-end session that writes house notes.
Both use the same executor, task sequences, budgets, recovery, and
completion checks. This comparison measures the added value of consolidated
long-term memory beyond retained spatial and task-level records.

\paragraph{C. Recovery availability and handover content.}
No recovery removes both requested and fallback restarts, while retaining
the complete 200-turn and 500-step task budget. The three handover
interventions keep recovery triggers, scope control, and the shared budget
unchanged. Empty handover supplies no failed-attempt note. Another task's
handover supplies a mismatched note. Matched-length summary replaces the
structured note with an ordinary summary. The first contrast tests whether
recovery helps. The remaining contrasts test what information must cross
the restart. They do not switch recovery off, and the summary comparison
tests information selection and organization together.

\paragraph{D. Verification use and evidence.}
No pre-stop verification retains post-stop certification but never withholds STOP.
No post-stop certification retains pre-stop verification but leaves the task account
without a post-stop certified status. One stop view instead of four keeps
both verification functions and changes the stop-position evidence from a
look-around to the forward frame. It therefore tests the evidence supplied
to verification, not the existence of either check. These are separate
rollouts, so changing a mechanism can change subsequent observations,
trajectories, and memory contents.

\subsection{Memory access and judge choice}
\label{app:supplementary-ablations}

\paragraph{Fixed-access historical memory.}
To test whether choosing when and what to retrieve matters beyond retaining
records, we replace agent-directed retrieval with a fixed-access policy.
Mapping tools, record writing, recovery handovers, and completion checks
remain unchanged. Both conditions use Opus 5,
receive the same initial memory entries, and can query the online map
freely. The control retrieves historical text at task start, after recovery,
and every ten navigation-model turns. A fixed query concatenates the goal
description and latest observation description, using the same description
interface in both conditions. BM25 retrieves up to five original excerpts
within a 2,000-token budget, including sources and recorded status. Records
are divided into 400-token chunks with 50-token overlap. Overlapping
excerpts are merged, and ties favor newer records. The navigator can still
consult current-task records and query the map, but cannot bypass the
fixed policy by reading historical text through other file tools.

We use the same full-split GOAT-Bench evaluation, three-run protocol, and
task budgets as Table~\ref{tab:organs-ablation}. Agent-directed access raises
s-SR by 7.0 percentage points over fixed access (95\% CI $[+5.9,+8.2]$),
with SPL increasing from 49.0 to 55.0
(Table~\ref{tab:fixed-access-memory}). Thus, supplying original records
on a fixed schedule does not recover the full benefit of letting the
navigator choose its queries as observations arrive. The comparison
tests query timing, selection, and follow-up together. Both conditions
start with empty memory and retain records in the same way, although
their later contents may differ as trajectories diverge. Neither
inherits house notes from earlier runs.

\begin{table}[htbp]
\centering
\small
\caption{Agent-directed and fixed-access historical memory on GOAT-Bench
with Opus 5. Scores are mean $\pm$ standard deviation over three runs.
The paired difference is NavHarness minus fixed access. The reference wall
time comes from the profiling runs in Appendix~\ref{app:runtime}.}
\label{tab:fixed-access-memory}
\begin{tabular*}{\linewidth}{@{\extracolsep{\fill}}lcc@{}}
\toprule
Metric & NavHarness & Fixed access \\
\midrule
s-SR (\%) & 81.5\,$\pm$\,0.7 & 74.5\,$\pm$\,0.8 \\
SPL & 55.0\,$\pm$\,0.6 & 49.0\,$\pm$\,0.7 \\
Mean time / task (s) & 129.5 & 137.2 \\
\midrule
Paired $\Delta$s-SR (pp) & \multicolumn{2}{c}{$+7.0$} \\
95\% CI (pp) & \multicolumn{2}{c}{$[+5.9,+8.2]$} \\
\bottomrule
\end{tabular*}
\end{table}

\begin{table}[!ht]
\centering
\small
\caption{Same-model and cross-model judging. Both completion checks use
the listed judge, with all other roles unchanged. Scores are mean $\pm$
standard deviation over three runs.}
\label{tab:cross-model-judging}
\begin{tabular*}{\linewidth}{@{\extracolsep{\fill}}llcc@{}}
\toprule
Navigator & Judge & s-SR (\%) & SPL \\
\midrule
Qwen3.8-27B & Qwen3.8-27B & 71.7\,$\pm$\,0.5 & 48.2\,$\pm$\,0.4 \\
Qwen3.8-27B & Opus 5 & 71.3\,$\pm$\,0.4 & 48.5\,$\pm$\,0.5 \\
\addlinespace
Opus 5 & Opus 5 & 81.5\,$\pm$\,0.7 & 55.0\,$\pm$\,0.6 \\
Opus 5 & Qwen3.8-27B & 81.7\,$\pm$\,0.6 & 55.2\,$\pm$\,0.5 \\
\bottomrule
\end{tabular*}
\end{table}

\paragraph{Cross-model judging.}
We swap the judges between Qwen3.8-27B and Opus 5 to test whether
navigation performance depends on using the navigator's own model for
completion checks. Each navigator is evaluated with its default judge
and with the other model judging both pre-stop verification and post-stop
certification. The navigator, recovery-note writer, and consolidation model
remain unchanged, as do the evidence, instructions, stopping rules, and
task budgets supplied to each configuration.

The swaps change s-SR by $-0.4$ points for Qwen and $+0.2$ for Opus,
with similarly small changes in SPL (Table~\ref{tab:cross-model-judging}).
Using a different judge therefore leaves navigation performance close to
the default in both configurations. This does not mean completion checks
are unimportant: group D of Table~\ref{tab:organs-ablation} tests their
contribution separately. Nor do similar navigation scores establish equal
judgment accuracy or exclude errors shared by the two models.

%% file: sections/app_results.tex
\section{Complete Results}
\label{app:results}

This appendix gives the paired statistics behind the main tables and the
diagnostics behind each component.
The GOAT-Bench score comparisons use all 2,669 subtasks in 360
validation-unseen episodes, matching Tables~\ref{tab:goat-main}
and~\ref{tab:organs-ablation}.
Three-seed score estimates and confidence intervals are identified
explicitly. The event percentages below are descriptive log summaries
and should not be read as additional three-seed effect estimates.

\subsection{Component effect profile}
\label{app:component-effects}
\begin{figure}[ht]
\centering
\input{figures/fig-component-effects}
\caption{GOAT-Bench full-split component effects with Opus 5, grouped as in Table~\ref{tab:organs-ablation}. Bars show the loss
relative to the full harness, with 95\% confidence intervals of the
paired change over three seeds. Positive values favor the full harness. Every
interval excludes zero.}
\label{fig:component-effects}
\end{figure}
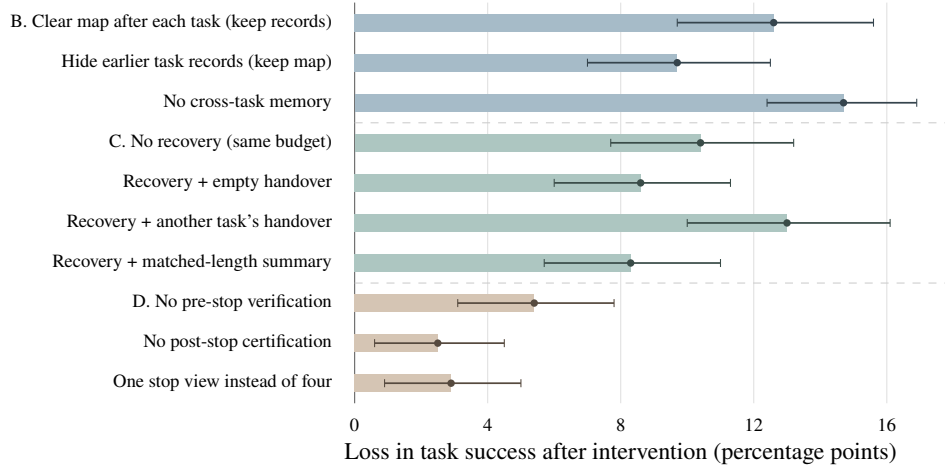

Figure~\ref{fig:component-effects} summarizes the controlled Opus
interventions in Groups B--D of Table~\ref{tab:organs-ablation}, including
the matched-length ordinary-summary control.
The pre-stop verification intervention is included separately from certification.
It displays the aggregate effect of each intervention with paired uncertainty.
The diagnostics below connect these effects to repeated exploration,
failed attempts, and inaccurate completion records.
Table~\ref{tab:organs-ablation} reports the scores and paired confidence
intervals together. The intervals pool three seeds per configuration using
the scene-clustered procedure in Appendix~\ref{app:stats}.

\subsection{Session and system comparisons}
\label{app:session-comparisons}
Independent sessions use only the current task's conversation, with no
external map, task records, or long-term memory. They differ from the map-reset
ablation, which retains mapping within each task.
Tables~\ref{tab:goat-main}--\ref{tab:organs-ablation} report the system
and session-policy comparisons, including the IR2R-CE independent-session
results for all four reasoning models.

With Qwen3.8-27B held fixed on the full 2,669-subtask evaluation, NavHarness
improves s-SR by 30.3 points, SPL by 20.7 points, and e-SR by 11.3 points.
The corresponding gains with GPT-4o are 34.8, 26.9, and 20.5 points.
With GPT-6 Astra held fixed on the same evaluation, NavHarness
improves s-SR by 18.6 points, SPL by 22.2 points, and e-SR by 23.3 points
over independent sessions. These are aggregate differences. Paired
uncertainty requires the corresponding task-level records, and the paired
intervals below concern Opus only.

In the full-split Opus study, independent sessions lose 22.6 success
points relative to full (95\% CI $[-26.3,-19.0]$), and a single session loses
26.6 points ($[-30.5,-22.8]$). Their SPL scores are lower by 20.1 and 22.2
points, respectively. These comparisons evaluate complete system policies.
SPL combines task success with path
efficiency, so these aggregate differences do not isolate path savings on
tasks completed by both systems.
Table~\ref{tab:organs-ablation} gives all component scores and paired
success intervals, including the empty-note and ordinary-summary controls.
The persistent-map and ledger removals also reduce SPL by 15.2 and 13.2
points.

\begin{table}[h]
\caption{The full harness by goal modality on GOAT-Bench. Percentages.}
\label{tab:modality}
\centering\small
\begin{apptab}{@{}lrr@{}}
Goal & s-SR & SPL \\
\midrule
Object category & 86.5 & 62.4 \\
Language description & 75.2 & 48.8 \\
Image & 82.2 & 53.0 \\
\bottomrule
\end{apptab}
\end{table}

\subsection{Within-task recovery and handovers}
\label{app:recovery-diag}
The matched-length ordinary-summary control reaches 73.2\% s-SR and 47.0\%
SPL, compared with 81.5\% and 55.0\% for the full harness. Its paired change
in s-SR is $-8.3$ percentage points, with a 95\% confidence interval of
$[-11.0,-5.7]$ points. This comparison tests the structured note against a
general summary, complementing the empty-note and task-mismatched-note
interventions in Table~\ref{tab:organs-ablation}.

\begin{figure}[h]
\centering
\resizebox{\linewidth}{!}{\input{figures/fig-c-recovery}}
\caption{Recovery on GOAT-Bench. Left: the share of recoveries that lead to
success, by what is handed to the fresh attempt. Middle and right: the model
requests recovery far earlier than the fallback and its requests succeed more
often.}
\label{fig:c-recovery}
\end{figure}
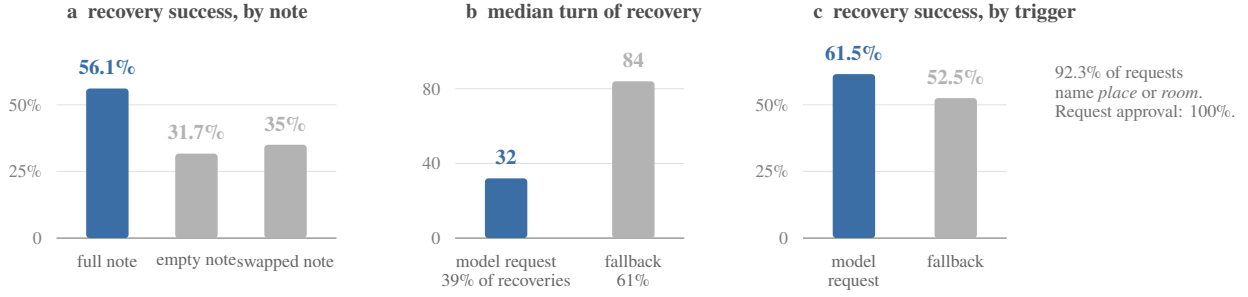

Figure~\ref{fig:c-recovery} breaks the recovery events on GOAT-Bench down
three ways. The full harness recovers on about a quarter of its tasks, with 39\%
of recoveries requested by the model and 61\% fired by the
80-turn fallback. Among the recovery events recorded in each arm,
56.1\% lead to success with the attempt's own note,
against 31.7\% with an empty note and 35.0\% with the note of a
different task. The model asks early, at a median of 32 turns against 84 for
the fallback, and its requests succeed more often, 61.5\% against 52.5\%,
although the two triggers select different failures. The recovered-task sets can
also differ across note interventions. These conditional rates describe
the events observed in each run, while Table~\ref{tab:organs-ablation}
tests the effect of the note intervention on overall task success.

\subsection{Cross-task memory}
\label{app:state-diag}
Disabling all cross-task memory while retaining within-task mapping,
recovery, and verification yields 66.8\% s-SR and 37.4\% SPL.
The paired s-SR change is $-14.7$ points (95\% CI $[-16.9,-12.4]$),
isolating the contribution of memory carried between tasks within the
otherwise intact harness.
Removing either task records or spatial persistence increases physical
exploration. Mean steps per task go from 127 with the full harness to 159
without the ledger and 168 without the persistent map, and without the map
the model drops 3.4 markers per task instead of 1.3 (about $2.6\times$)
and requests recovery 1.8
times as often, rebuilding each task what the previous one had already
recorded.

\subsection{Verification uses and evidence}
\label{app:closure-diag}
\paragraph{Pre-stop verification interventions.}
\label{app:gate-diag}
Disabling only pre-stop verification reduces full-system success from 81.5\% to
76.1\%, a change of $-5.4$ points with paired 95\% CI $[-7.8,-3.1]$.
SPL changes from 55.0\% to 47.0\%.
The interval excludes zero over the three seeds. With pre-stop verification enabled in both
configurations, the full system exceeds independent sessions by
19.1 success points (95\% CI 16.4--22.4). This comparison
holds the stopping check fixed, but evaluates the remaining harness
features jointly, including persistent state and recovery.

\paragraph{Local stopping effects from logs.}
The first refused stopping request provides a second view of pre-stop verification. For task $i$, let $y_i^{\mathrm{stop}}$ be evaluator success if
STOP were executed at that recorded state, and $y_i^{\mathrm{final}}$
the task's eventual success after the refusal. A rescue has values
$(0,1)$ and a harm has values $(1,0)$. The local net change is
\begin{equation}
\widehat{\delta}_{\mathrm{stop}}=
\frac{1}{N}\sum_{i\in\mathcal R}
\big(y_i^{\mathrm{final}}-y_i^{\mathrm{stop}}\big),
\end{equation}
where $\mathcal R$ contains tasks with a refused first request and $N$ is
the number of tasks in the GOAT-Bench Val-Unseen diagnostic logs.
This paired comparison conditions on the history produced by the policy with pre-stop verification enabled. It estimates the current-task consequence of continuing at those
states, not the effect of rerunning an entire sequence without pre-stop verification.
Recovering $y_i^{\mathrm{stop}}$ requires the benchmark's success criterion
at that state, not the agent's claim or the judge's verdict.

\begin{table}[ht]
\caption{Pre-stop diagnostics. Values are percentages
except the net change, which is in percentage points.}
\label{tab:gate-diagnostics}
\centering\small
\begin{apptab}{@{}lrr@{}}
Diagnostic & Full & Independent + pre-stop verification \\
\midrule
Tasks with a refused first stop & 11 & 20 \\
Rejection precision & 73 & 75 \\
Rescue among correctly refused stops & 45 & 35 \\
Rescues / all tasks & 3.6 & 5.0 \\
Harms / all tasks & 0.7 & 1.1 \\
Net local success change & $+2.9$ & $+4.0$ \\
\bottomrule
\end{apptab}
\end{table}
The reported summaries suggest that independent sessions reach questionable
stopping states more often, while the rejection precision is similar
in the two configurations. The local estimates measure continuation after
a refused STOP, whereas rollout differences include changes to the entire
trajectory and later state.

\begin{figure}[!htb]
\centering
\input{figures/fig-c-closure}
\caption{Verification on GOAT-Bench. Left: four-view rejections as a share of claimed completions. Right: the
success of the task that follows a false or a true ledger record, with and
without certification.}
\label{fig:c-closure}
\end{figure}
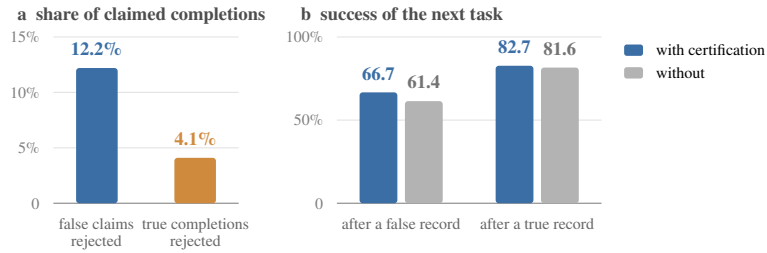

\paragraph{Post-stop certification.}
Among claimed completions, the four-view certification judge correctly rejects false
claims on 12.2\% and incorrectly rejects true claims on 4.1\%
(Table~\ref{tab:judge}). Both percentages use all claimed completions as
their denominator. The next task receives the verdict alongside the claim and account
(Figure~\ref{fig:c-closure}). Success of the task that follows
a false record is 66.7\% with certification and 61.4\% without. After a true
record it is 82.7\% and 81.6\%. These successor rates condition on the
histories produced by each configuration, rather than matched predecessor
states, and do not isolate the effect of certification.
Table~\ref{tab:judge} compares the certification judge's verdicts with environment
truth under the four stop views and under a single forward frame. The two
runs have different terminal observations, so the comparison describes the
runs rather than isolating the view count.

\begin{table}[h]
\caption{Post-stop certification outcomes and accuracy (\% of claimed completions).
Parentheses give accept-all accuracy.}
\label{tab:judge}
\centering\small\setlength{\tabcolsep}{4pt}
\begin{apptab}{@{}lrrrrrr@{}}
Evidence & \shortstack{Certified,\\true} & \shortstack{Certified,\\false}
& \shortstack{Rejected,\\true} & \shortstack{Rejected,\\false}
& \shortstack{Acc.\\(accept-all)} & $\kappa$ \\
\midrule
Four stop views & 78.5 & 5.2 & 4.1 & 12.2 & 90.7 (82.6) & 0.67 \\
One forward frame & 74.2 & 10.7 & 6.4 & 8.7 & 82.9 (80.6) & 0.40 \\
\bottomrule
\end{apptab}
\end{table}

On the same claimed-completion population, four-view accuracy is 90.7\%,
compared with 82.6\% for accepting every claim. The corresponding values
for the single-view run are 82.9\% and 80.6\%. Cohen's $\kappa$ is
approximately 0.67 and 0.40, respectively, computed from the rounded
percentages above. The four-view checker rejects 70.1\% of false claims
and accepts 95.0\% of true claims. Of its rejections, 25.2\% concern true
completions. Both checkers improve on accept-all accuracy, with a larger
margin in the four-view run. These results support using certification
to distinguish supported outcomes from unsupported completion claims,
while retaining the possibility of false acceptance and false rejection.
The post-stop
certification ablation leaves the pre-stop verification unchanged
(Appendix~\ref{app:benchmarks}), so it does not isolate that check's effect
on navigation success.

%% file: figures/fig-component-effects.tex
\begin{tikzpicture}[x=0.44cm,y=0.53cm]
\definecolor{memorybar}{HTML}{7696AB}
\definecolor{recoverybar}{HTML}{81A89F}
\definecolor{verificationbar}{HTML}{BEA58C}
\foreach \x in {0,4,8,12,16} {
  \draw[black!12] (\x,-0.5) -- (\x,9.5);
  \node[font=\scriptsize,anchor=north] at (\x,-0.65) {\x};
}
\draw[black!60] (0,-0.5) -- (0,9.5);
\foreach \y in {6.5,2.5} {
  \draw[black!20,dashed] (0,\y) -- (18,\y);
}
\foreach \y/\v/\lo/\hi/\lab in {
9/12.6/9.7/15.6/B. Clear map after each task (keep records),
8/9.7/7.0/12.5/Hide earlier task records (keep map),
7/14.7/12.4/16.9/No cross-task memory,
6/10.4/7.7/13.2/C. No recovery (same budget),
5/8.6/6.0/11.3/Recovery + empty handover,
4/13.0/10.0/16.1/Recovery + another task's handover,
3/8.3/5.7/11.0/Recovery + matched-length summary,
2/5.4/3.1/7.8/D. No pre-stop verification,
1/2.5/0.6/4.5/No post-stop certification,
0/2.9/0.9/5.0/One stop view instead of four} {
  \ifnum\y>6 \colorlet{rowcolor}{memorybar}\else
    \ifnum\y>2 \colorlet{rowcolor}{recoverybar}\else
      \colorlet{rowcolor}{verificationbar}\fi\fi
  \fill[rowcolor!65] (0,\y-0.22) rectangle (\v,\y+0.22);
  \draw[rowcolor!45!black,line width=0.65pt] (\lo,\y)--(\hi,\y);
  \draw[rowcolor!45!black] (\lo,\y-0.10)--(\lo,\y+0.10);
  \draw[rowcolor!45!black] (\hi,\y-0.10)--(\hi,\y+0.10);
  \fill[rowcolor!45!black] (\v,\y) circle[radius=1.4pt];
  \node[font=\scriptsize,anchor=east] at (-0.4,\y) {\lab};
}
\node[font=\small,anchor=north] at (8,-1.25)
  {Loss in task success after intervention (percentage points)};
\end{tikzpicture}

%% file: figures/fig-c-recovery.tex
\begin{tikzpicture}[x=1cm,y=1cm]
\definecolor{v4c}{HTML}{3A6EA5}\definecolor{selfc}{HTML}{D08A3E}\definecolor{neverc}{HTML}{8C8C8C}
\draw[black!12,line width=0.3pt] (0.25,0.000) -- (3.55,0.000); \node[font=\fontsize{6}{6}\selectfont,text=black!50,anchor=east] at (0.20,0.000) {0};
\draw[black!12,line width=0.3pt] (0.25,0.786) -- (3.55,0.786); \node[font=\fontsize{6}{6}\selectfont,text=black!50,anchor=east] at (0.20,0.786) {25\%};
\draw[black!12,line width=0.3pt] (0.25,1.571) -- (3.55,1.571); \node[font=\fontsize{6}{6}\selectfont,text=black!50,anchor=east] at (0.20,1.571) {50\%};
\draw[black!45,line width=0.4pt] (0.25,0) -- (3.55,0);
\fill[v4c,rounded corners=1.2pt] (0.60,0) rectangle (1.10,1.763);
\node[font=\scriptsize\bfseries,text=v4c,anchor=south] at (0.85,1.803) {56.1\%};
\node[font=\fontsize{6}{6.5}\selectfont,text=black!60,anchor=north,align=center] at (0.85,-0.07) {full note};
\fill[black!30,rounded corners=1.2pt] (1.65,0) rectangle (2.15,0.996);
\node[font=\scriptsize\bfseries,text=black!30,anchor=south] at (1.90,1.036) {31.7\%};
\node[font=\fontsize{6}{6.5}\selectfont,text=black!60,anchor=north,align=center] at (1.90,-0.07) {empty note};
\fill[black!30,rounded corners=1.2pt] (2.70,0) rectangle (3.20,1.100);
\node[font=\scriptsize\bfseries,text=black!30,anchor=south] at (2.95,1.140) {35\%};
\node[font=\fontsize{6}{6.5}\selectfont,text=black!60,anchor=north,align=center] at (2.95,-0.07) {swapped note};
\node[font=\scriptsize\bfseries,text=black!70,anchor=south west] at (0.25,2.42) {a \ recovery success, by note};
\draw[black!12,line width=0.3pt] (4.95,0.000) -- (7.65,0.000); \node[font=\fontsize{6}{6}\selectfont,text=black!50,anchor=east] at (4.90,0.000) {0};
\draw[black!12,line width=0.3pt] (4.95,0.880) -- (7.65,0.880); \node[font=\fontsize{6}{6}\selectfont,text=black!50,anchor=east] at (4.90,0.880) {40};
\draw[black!12,line width=0.3pt] (4.95,1.760) -- (7.65,1.760); \node[font=\fontsize{6}{6}\selectfont,text=black!50,anchor=east] at (4.90,1.760) {80};
\draw[black!45,line width=0.4pt] (4.95,0) -- (7.65,0);
\fill[v4c,rounded corners=1.2pt] (5.30,0) rectangle (5.80,0.704);
\node[font=\scriptsize\bfseries,text=v4c,anchor=south] at (5.55,0.744) {32};
\node[font=\fontsize{6}{6.5}\selectfont,text=black!60,anchor=north,align=center] at (5.55,-0.07) {model request\\39\% of recoveries};
\fill[black!30,rounded corners=1.2pt] (6.80,0) rectangle (7.30,1.848);
\node[font=\scriptsize\bfseries,text=black!30,anchor=south] at (7.05,1.888) {84};
\node[font=\fontsize{6}{6.5}\selectfont,text=black!60,anchor=north,align=center] at (7.05,-0.07) {fallback\\61\%};
\node[font=\scriptsize\bfseries,text=black!70,anchor=south west] at (4.95,2.42) {b \ median turn of recovery};
\draw[black!12,line width=0.3pt] (9.05,0.000) -- (11.45,0.000); \node[font=\fontsize{6}{6}\selectfont,text=black!50,anchor=east] at (9.00,0.000) {0};
\draw[black!12,line width=0.3pt] (9.05,0.786) -- (11.45,0.786); \node[font=\fontsize{6}{6}\selectfont,text=black!50,anchor=east] at (9.00,0.786) {25\%};
\draw[black!12,line width=0.3pt] (9.05,1.571) -- (11.45,1.571); \node[font=\fontsize{6}{6}\selectfont,text=black!50,anchor=east] at (9.00,1.571) {50\%};
\draw[black!45,line width=0.4pt] (9.05,0) -- (11.45,0);
\fill[v4c,rounded corners=1.2pt] (9.40,0) rectangle (9.90,1.933);
\node[font=\scriptsize\bfseries,text=v4c,anchor=south] at (9.65,1.973) {61.5\%};
\node[font=\fontsize{6}{6.5}\selectfont,text=black!60,anchor=north,align=center] at (9.65,-0.07) {model\\request};
\fill[black!30,rounded corners=1.2pt] (10.60,0) rectangle (11.10,1.650);
\node[font=\scriptsize\bfseries,text=black!30,anchor=south] at (10.85,1.690) {52.5\%};
\node[font=\fontsize{6}{6.5}\selectfont,text=black!60,anchor=north,align=center] at (10.85,-0.07) {fallback};
\node[font=\scriptsize\bfseries,text=black!70,anchor=south west] at (9.05,2.42) {c \ recovery success, by trigger};
\node[font=\fontsize{6}{6.5}\selectfont,text=black!55,anchor=north west,align=left] at (11.9,2.15) {92.3\% of requests\\name \emph{place} or \emph{room}.\\Request approval: 100\%.};
\end{tikzpicture}

%% file: figures/fig-c-closure.tex
\begin{tikzpicture}[x=1cm,y=1cm]
\definecolor{v4c}{HTML}{3A6EA5}\definecolor{selfc}{HTML}{D08A3E}\definecolor{neverc}{HTML}{8C8C8C}
\draw[black!12,line width=0.3pt] (0.60,0.000) -- (3.05,0.000); \node[font=\fontsize{6}{6}\selectfont,text=black!50,anchor=east] at (0.54,0.000) {0};
\draw[black!12,line width=0.3pt] (0.60,0.733) -- (3.05,0.733); \node[font=\fontsize{6}{6}\selectfont,text=black!50,anchor=east] at (0.54,0.733) {5\%};
\draw[black!12,line width=0.3pt] (0.60,1.467) -- (3.05,1.467); \node[font=\fontsize{6}{6}\selectfont,text=black!50,anchor=east] at (0.54,1.467) {10\%};
\draw[black!12,line width=0.3pt] (0.60,2.200) -- (3.05,2.200); \node[font=\fontsize{6}{6}\selectfont,text=black!50,anchor=east] at (0.54,2.200) {15\%};
\draw[black!45,line width=0.4pt] (0.60,0) -- (3.05,0);
\fill[v4c,rounded corners=1.2pt] (0.90,0) rectangle (1.45,1.789);
\node[font=\scriptsize\bfseries,text=v4c,anchor=south] at (1.17,1.829) {12.2\%};
\node[font=\fontsize{6}{6.5}\selectfont,text=black!60,anchor=north,align=center] at (1.17,-0.07) {false claims\\rejected};
\fill[selfc,rounded corners=1.2pt] (2.20,0) rectangle (2.75,0.601);
\node[font=\scriptsize\bfseries,text=selfc,anchor=south] at (2.48,0.641) {4.1\%};
\node[font=\fontsize{6}{6.5}\selectfont,text=black!60,anchor=north,align=center] at (2.48,-0.07) {true completions\\rejected};
\node[font=\scriptsize\bfseries,text=black!70,anchor=base west] at (0.0,2.42) {a \ share of claimed completions};
\draw[black!12,line width=0.3pt] (4.35,0.000) -- (7.85,0.000); \node[font=\fontsize{6}{6}\selectfont,text=black!50,anchor=east] at (4.29,0.000) {0};
\draw[black!12,line width=0.3pt] (4.35,1.100) -- (7.85,1.100); \node[font=\fontsize{6}{6}\selectfont,text=black!50,anchor=east] at (4.29,1.100) {50\%};
\draw[black!12,line width=0.3pt] (4.35,2.200) -- (7.85,2.200); \node[font=\fontsize{6}{6}\selectfont,text=black!50,anchor=east] at (4.29,2.200) {100\%};
\draw[black!45,line width=0.4pt] (4.35,0) -- (7.85,0);
\fill[v4c,rounded corners=1.2pt] (4.65,0) rectangle (5.15,1.467); \node[font=\scriptsize\bfseries,text=v4c,anchor=south] at (4.90,1.507) {66.7};
\fill[black!30,rounded corners=1.2pt] (5.25,0) rectangle (5.75,1.351); \node[font=\scriptsize\bfseries,text=black!55,anchor=south] at (5.50,1.391) {61.4};
\node[font=\fontsize{6}{6.5}\selectfont,text=black!60,anchor=north] at (5.20,-0.07) {after a false record};
\fill[v4c,rounded corners=1.2pt] (6.45,0) rectangle (6.95,1.819); \node[font=\scriptsize\bfseries,text=v4c,anchor=south] at (6.70,1.859) {82.7};
\fill[black!30,rounded corners=1.2pt] (7.05,0) rectangle (7.55,1.795); \node[font=\scriptsize\bfseries,text=black!55,anchor=south] at (7.30,1.835) {81.6};
\node[font=\fontsize{6}{6.5}\selectfont,text=black!60,anchor=north] at (7.00,-0.07) {after a true record};
\node[font=\scriptsize\bfseries,text=black!70,anchor=base west] at (3.75,2.42) {b \ success of the next task};
\fill[v4c,rounded corners=1pt] (8.15,1.95) rectangle (8.40,2.1); \node[font=\fontsize{6}{6.5}\selectfont,anchor=west] at (8.45,2.025) {with certification};
\fill[black!30,rounded corners=1pt] (8.15,1.65) rectangle (8.40,1.8); \node[font=\fontsize{6}{6.5}\selectfont,anchor=west] at (8.45,1.725) {without};
\end{tikzpicture}

%% file: sections/app_analysis.tex
\section{Continuous Deployment and Mechanism Analysis}
\label{app:analysis}

\subsection{Component effects and failure diagnostics}
Table~\ref{tab:stress-map} links aggregate intervention losses to the
failure patterns observed in the corresponding runs. The diagnostics
help explain what each component supports, but are not separate effect
estimates for tasks stratified by difficulty or failure type.

\begin{table}[h]
\caption{Component effects and associated diagnostics. Changes in s-SR
are aggregate interventions, not effects estimated within each stress
condition. The note intervention substitutes another task's note.}
\label{tab:stress-map}
\centering\small
\setlength{\tabcolsep}{5pt}
\begin{apptab}{lrl>{\raggedright\arraybackslash}p{5.6cm}}
\thd{Component} & \thd{$\Delta$s-SR} & \thd{Absorbs} & \thd{What shows when it is removed} \\
\midrule
Map and markers & $-12.6$ & revisits and displacement & steps $+32\%$, markers written about $2.6\times$ \\
Ledger & $-9.7$ & repeated goals & steps $+25\%$ despite retaining the map \\
Recovery note & $-13.0$ & a stalled search & recovery success 35\% instead of 56\% \\
Recovery & $-10.4$ & an unproductive attempt & requests come long before the fallback \\
Four-view evidence & $-2.9$ & an ambiguous closure & claimed-completion $\kappa$: 0.40 vs.\ 0.67 in separate runs \\
Certification & $-2.5$ & an unsupported claim & 12.2\% false and 4.1\% true completions rejected (share of claims) \\
\bottomrule
\end{apptab}
\end{table}

\input{sections/app_memcmp}

\subsection{Archive construction and access}
\label{app:archive-efficiency}
\begin{humanprompted}
During deployment, long-term memory carries task experience
across simulated days through house overviews, room notes, and skill
files. The overview provides an entry point to this collection rather
than replacing its detailed records. The executor reads these records
through file tools. Task accounts preserve new experience as it is
recorded, while a separate consolidation session summarizes that
experience for later use.
The 36-house comparison in Section~\ref{sec:exp:day} holds map and
task-record retention fixed and switches only this consolidation session
off. Pooled s-SR rises from 72.8\% to 80.5\% with consolidation, a paired
gain of 7.7 points (95\% CI $[+6.3,+9.1]$), while SPL rises from 36.5 to 44.3,
a gain of 7.8 points (95\% CI $[+6.5,+9.2]$). Thus house notes improve
later navigation even when maps and detailed task records remain available.
The cases in Appendix~\ref{app:selfcorrect} trace this process from
observations to later decisions, including a corrected dresser
description, a refrigerator identity resolved across views, and a
revised instruction about a plant's location. The archive preserves
successful routes, failed searches, and open questions alongside their
sources so that later sessions can revisit the evidence behind a note.
\end{humanprompted}
\FloatBarrier

%% file: sections/app_memcmp.tex
\begin{humanconfirmed}
\subsection{With and without consolidation across multiple houses}
\label{app:selfcorrect}

\begin{humanprompted}
The deployment schedule extends GOAT-Bench to ten tours in the same
house, organized across simulated days. Tasks within a tour retain
GOAT-Bench's continuous motion, but tour boundaries are discontinuous:
the robot is placed at the next tour's prescribed starting location.
This models a household robot resuming work after a shutdown. It must
recover its bearings from current observations, SLAM, and remembered
landmarks before reusing stored routes and target locations. The
spatial-record cases below examine how sessions handle inherited
coordinates when the SLAM frame changes and how descriptive knowledge
can still guide navigation.
\end{humanprompted}

The extended evaluation covers all 36 houses
(Figure~\ref{fig:deployment-progress-preview}). This appendix provides detailed profiles
and case studies for three of these houses, labeled A, B, and C, comparing two configurations
of the same executor on matched task sequences. Both retain maps, markers,
the place graph, and task records, including the ledger and handovers,
with the same persistence and access rules. The \emph{No consolidation}
control disables only run-end consolidation. The \emph{full} configuration
also distills the journal into house notes for later runs, which navigation
sessions read through file tools. All other settings are identical:
Claude Opus 5, one session per task, 200 turns and 500 steps per task, at
most one recovery, the four-view certification judge and the pre-stop verification.
On the first day of each house no notes exist yet; the
executor can only read the journal being appended that day. The executor
may read its notes but not write them: Read, Glob and Grep are fenced to the
notes directory, and every Write or Edit is refused.
In the case-study logs, \emph{ledger} or \emph{journal} also labels the
full task records, including the actor's account, rather than just the
compact index in \textnormal{run/ledger.md}. We label these detailed
excerpts as task records below and retain the original wording in quotations.
Boxed passages are the model's own writing, quoted verbatim, with our highlights: \hlm{a judgement about its own memory}, \hlw{an admission of error or cost}, \hlo{advice to its future self, or a diagnosis}.

\paragraph{What was asked for, and what was not.}
\begin{humanprompted}
Both configurations can build a working map, add markers, and produce
task records during navigation, and both can retrieve these records later.
Only the full configuration consolidates this experience into an index,
house overview, room notes, and navigation skills. The comparison tests
what this long-term organization adds beyond persistent maps and task records.
\end{humanprompted}

Both configurations receive the same deployment task prompt, tool
descriptions, and handover instruction. The logs use CLI-native names
such as Read, Glob, and Grep. Table~\ref{tab:memcmp-boundary} describes
the instructions associated with the case-study records. Both configurations
have access to the ledger and detailed task records; only full receives
the house notes produced by consolidation.

Table~\ref{tab:memcmp-boundary} separates explicitly requested operations
from more specific behaviors observed during execution. A behavior not
explicitly scripted can still follow from the general memory instructions.

\input{sections/memcmp_tab_boundary}

Both arms use the same task order in each case-study house.
Figure~\ref{fig:memcmp-houses} shows the separate house profiles, and
Figure~\ref{fig:memcmp-glance} summarizes the complete runs.
Figure~\ref{fig:deployment-progress-preview} reports the pooled 36-house evaluation.

\begin{figure}[!htb]
\centering
\includegraphics[width=\linewidth]{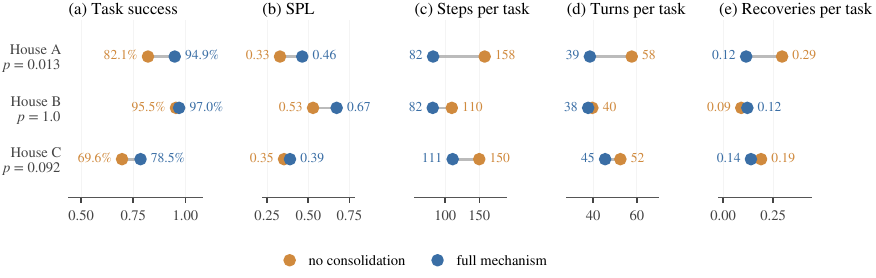}
\caption{Aggregate comparison for deployments A--C. Success and SPL use
environment truth. Steps, turns, and recoveries are means per task;
$p$ is an exact McNemar test on task success.}
\label{fig:memcmp-glance}
\end{figure}

\begin{figure}[!htb]
\centering
\includegraphics[width=0.9\linewidth]{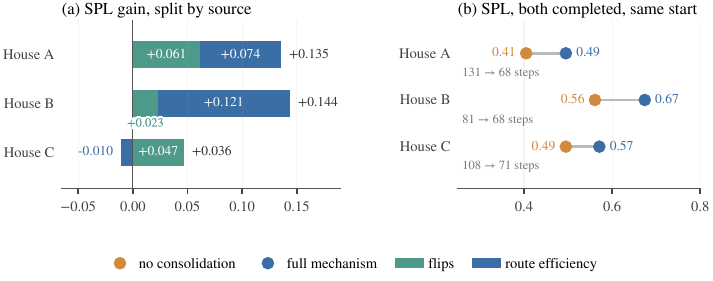}
\caption{Where the SPL gain comes from. \textbf{(a)} The gain split into
tasks that only one run completes (\emph{flips}) and tasks both runs
complete (\emph{route efficiency}); in House C the efficiency term is
negative. \textbf{(b)} Because tasks chain, the two runs do not always start
a task from the same place, and SPL is normalised by each run's own shortest
path; on the both-completed tasks whose starts differ by less than 0.5\,m,
efficiency rises in each environment.}
\label{fig:memcmp-spl}
\end{figure}

\begin{figure}[!htb]
\centering
\includegraphics[width=0.9\linewidth]{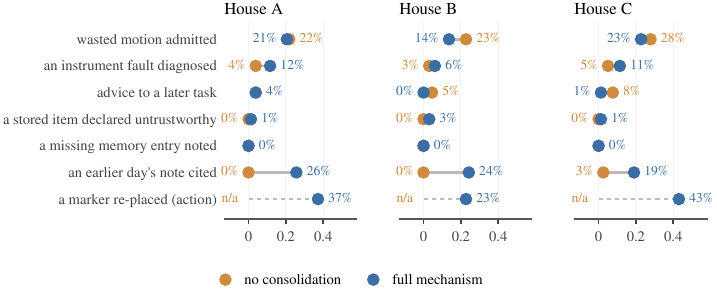}
\caption{Keyword-match rates in task accounts. Textual mentions need not
correspond to a verified memory read or revision. The last row is an action rather than a
phrase: a marker re-placed under a label that already existed.
Both configurations retain markers; untabulated rates are marked n/a.}
\label{fig:memcmp-kinds}
\end{figure}

\begin{figure}[p]
\centering
\includegraphics[width=0.95\linewidth,height=0.45\textheight,keepaspectratio]{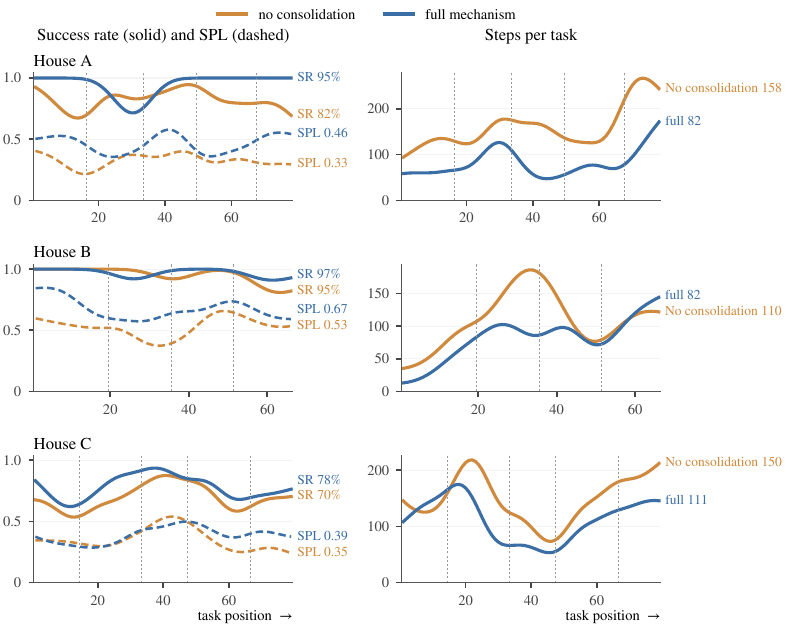}
\caption{Performance trends in deployments A--C. Left, success rate (solid) and SPL
(dashed); right, steps per task. Ochre denotes no consolidation and blue the full mechanism. The
label at each line end is the run's mean over all tasks; dashed verticals
are the day boundaries of the full run. The raw outcome of every task is in
Figure~\ref{fig:memcmp-cases-strip}.}
\label{fig:memcmp-houses}
\par\medskip
\resizebox{\linewidth}{!}{\input{figures/fig-memcmp-cases-strip}}
\caption{Where the cases sit in deployments A--C. One cell per task in
order, teal for success and red-brown for failure by environment truth; the
upper strip of each house is the full run, the lower the run without
consolidation, and dashed lines are the day boundaries of the full run.
Numbered dots mark the tasks each case quotes:
\textcolor{figblue}{revising spatial records} (Cases 1--4),
\textcolor{figochre}{recording failure and checking completion} (Cases 5--7),
\textcolor{figteal}{instruments and procedure} (Cases 8--9),
\textcolor{figred}{auditing and consolidating memory} (Cases 10--11),
\textcolor{black!70}{declining stored experience} (Cases 12--14). The cases
are spread over the days rather than clustered at their starts, and most of
them fall on tasks the two runs decided differently.}
\label{fig:memcmp-cases-strip}
\end{figure}

\paragraph{Different environments, different gains.}
In House A both success and route efficiency improve, and the SPL gain
splits evenly between tasks that flip to success and tasks both runs
complete. In House B success rates are already high and differ only
slightly, while route efficiency improves substantially:
the both-completed tasks are shorter, and the share of tasks closed within
ten steps rises from 8\% to 18\%. In House C the gain is almost entirely in
success, and the both-completed tasks appear no more efficient; that is an
artefact of chained starts, since on the subset whose starts coincide the
efficiency rises in each environment (Figure~\ref{fig:memcmp-spl}). By goal
modality the gain is broadest on described goals: success on them rises
from 77\% to 95\% in House A and from 60\% to 72\% in House C, and their
steps per task fall from 217 to 93 and from 204 to 157; the object goals of
House B are all completed by both runs and differ only in steps, 66 against
41 per task. The closing judge accepts a false completion on 10\%, 1.5\% and 19\% of
tasks without consolidation and on 3\%, 0\% and 18\% with it (Houses A,
B, C), and refuses a true one on 8--11\% without and 6--15\% with. The
fourteen cases below are selected from forty-eight analyzed cases to
illustrate spatial revision, failure recording, instrument diagnosis,
consolidation, and selective reuse. Each includes the corresponding
no-consolidation execution where available. This behavior-based selection includes
both benefits and failures, rather than estimating their frequency
across all navigation tasks.
The cases trace how memory enters individual decisions. Component effects
are tested by the controlled interventions, while differences between
these full-system and no-consolidation trajectories are not decomposed into
independent gains from each record or tool.

Task by task (Figure~\ref{fig:memcmp-houses}), the full run's success in
House A drops below the no-consolidation run's only on the second day, where two
failures on the same lamp table (Case~7) coincide, and its steps stay below
throughout. In House B the success curves are close, while the larger
difference is in the steps of the first two days, when the closet tasks are
answered from the marker under the body (Case~6); in House C the SPL
curves cross several times, the gain being in tasks that flip to success.
Figure~\ref{fig:memcmp-cases-strip} places the cases on the same strips.
\FloatBarrier
\subsubsection{Revising spatial records (Cases 1 to 4)}

\paragraph{Case 1. Correction that tracks the map frame, not the object.}
This case distinguishes a tracker restart between runs from tracking loss
within a run. Saved markers survive either event, but their coordinates
are usable only when aligned with the current tracker frame.
The following measurements track re-placements in the full runs.
In House A the executor re-placed a marker under an
existing label 29 times during the deployment. Each re-placement can be measured
twice: in the map, as the distance between the old and the new marker, and
against the simulator, as the distance between the two standing positions.
Sixteen are refinements, with both distances small. Thirteen are frame
events, nine at the start of a day, when SLAM restarts from the origin and
yesterday's markers keep yesterday's coordinates, and four after a tracking
loss within a day; in
seven of these the ``correction'' was itself placed in a frame that was
already broken. Houses B and C show the same shape
(Figure~\ref{fig:memcmp-calibration}): of the 15 re-placements in House B,
11 are frame events and 2 are moves the simulator confirms; of the 34 in
House C, 21 are frame events.

\begin{figure}[!htb]
\centering
\includegraphics[width=1.0\linewidth]{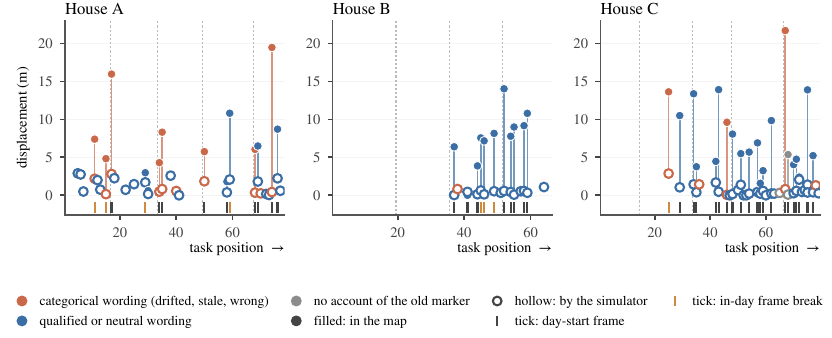}
\caption{Every re-placement of an existing label in the three full runs, placed at its task position. For each, the filled dot is the displacement in the map and the hollow dot the displacement of the standing position by the simulator, joined by a line; the colour is the wording of the account about the old marker. Ticks below the axis mark the cause: black, a day-start frame reset; ochre, a frame break within a day. Categorical wording follows the map, not the object: the large map displacements with small true ones are frame events, and the accounts call them drift.}
\label{fig:memcmp-calibration}
\end{figure}

\begin{ledger}[barwarn]{House A · task 17 · day 2 · full · task record}{Find any carpet.}{first task of the day; in the new frame the inherited marker was 16\,m from the rug}
I re-dropped the ``freezer room'' marker on the rug --- \hlw{the old one was badly drifted, and previewing routes to it failed}. [\ldots] once red fills the bottom of the frame you are already on it.
\end{ledger}
\begin{ledger}[barmem]{House A · task 17 · no-consolidation · task record}{Find any carpet.}{24 steps; closed on the hallway carpet, accepted by the verifier, a failure by the environment (4.6\,m)}
Goal was to stand on the hallway carpet. I reached it --- I'm on the beige carpet now, in the corridor lined with wooden doors [\ldots] \hlm{I dropped one marker, ``carpet hallway'', where I stopped. Nothing else explored yet.}
\end{ledger}

\begin{insight}
\textbf{What it shows.} The accounts track the size of the observed offset more reliably than
its cause. Categorical words such as \textit{drifted} and \textit{stale}
appear in nine of twelve large offsets, while fifteen of sixteen small
offsets receive qualified or no comment. The position records associate
the large offsets with coordinate-frame changes, but the notes generalize
them into \textit{Markers here drift}. Retaining a marker therefore also
requires checking whether its frame remains usable. On the thirteen
frame-event tasks the full run still completes more tasks with fewer
steps while using its other retained knowledge.
\end{insight}

\paragraph{Case 2. A misplaced marker, caught nineteen tasks later.}
In House A the label \textit{laundry bath} was first placed at task 4 and
placed again at task 15, in a frame that a tracking loss had already
displaced: the new marker sat 4.8\,m from the old one in the map while the
body stood 0.15\,m from where it had stood before (the account: ``the old
`laundry bath' marker had drifted into the entry hall --- I re-dropped it in
front of the boiler''). Nothing used the label
until task 34, the first task of day 3, when the day-start reset put the
inherited marker 4.1\,m off in the new frame. The executor followed its
stored route toward it, ended on a spurious ``floor 2'', asked for a
floor-scope recovery that was refused (``the robot created and placed me
on `floor 2', while the whole house map with all my markers is floor 1''),
walked in by landmarks and re-placed the label, calling it drifted. From then on the marker was
accurate in that day's frame, the preview reported it reachable, and the
next two boiler tasks walked straight to it; the account of task 44 reads
``the laundry bath one was accurate'' (Figure~\ref{fig:memcmp-laundry}).
\begin{figure}[!htb]
\centering
\includegraphics[width=0.8\linewidth]{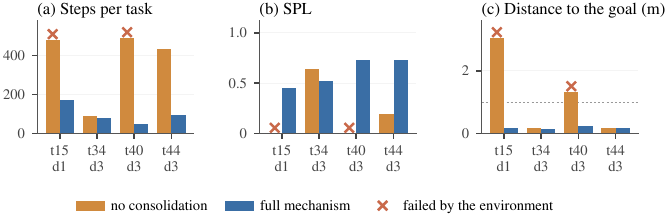}
\caption{House A, the tasks that placed or used the \textit{laundry bath} marker, in both runs: steps, SPL and distance to the goal by the simulator (the dashed line is the 1\,m success radius; a cross marks a failure). The starts coincide at tasks 34 and 40 and lie 1.4\,m and 0.1\,m apart at tasks 15 and 44; the no-consolidation run recovered once at each of tasks 15, 40 and 44, the full run never.}
\label{fig:memcmp-laundry}
\end{figure}

\begin{ledger}[barok]{House A · task 40 · day 3 · full · task record}{Find the specific boiler shown in the goal image.}{51 steps, SPL 0.73; the preview reported the marker reachable at 2.5\,m}
\hlo{From the start I followed the saved route} --- tiled entry hall, out to the carpeted hallway past the kitchen opening and staircase, then the door on the right into the laundry/bath room. Re-dropped the ``laundry bath'' marker inside the room, \hlw{since the old one had drifted}.
\end{ledger}
\begin{ledger}[barwarn]{House A · task 40 · no-consolidation · task record}{The same goal.}{491 steps, one recovery, failure 1.33\,m from the goal}
Searched for the boiler from the photo and \hlw{never found it}. I started in the carpeted bedroom next to the laundry door, \hlw{went through the laundry with the washer and dryer}, followed the corridor [\ldots] checked the couch room, the bathroom hall, the red-rug rec room, and the alcove with the chest freezers and fridge.
\end{ledger}

\begin{insight}
\textbf{What it shows.} The misplaced marker is not diagnosed when first written. It remains
unused for nineteen tasks and is re-placed after a subsequent frame
reset (Figure~\ref{fig:memcmp-lineage}a). On tasks 40 and 44 the session
then combines the reachable marker with its written landmark route.
It takes 51 versus 491 steps on task 40, and 93 versus 436 on task 44,
without recovery. On task 34, however, following the stored route west
adds 11 steps relative to no-consolidation. The records show why checking current
location matters before reusing a previously successful route.
\end{insight}

\paragraph{Case 3. Moves the simulator never saw.}
In House C a calendar hangs on the refrigerator surround and is asked for
five times. The same label was placed four times, at positions within
0.5\,m of one another by the simulator, while its map coordinates changed
with each day's frame; the accounts call the earlier marker \textit{moved},
\textit{misled} and \textit{drifted}, and only one of the three matches the
truth; at task 40, 58 steps later, ``the existing `back hall by fridge'
one sufficed''. Figure~\ref{fig:memcmp-backhall} shows every task in both runs; the
starts coincide within 0.9\,m in all three pairs discussed here. The full
run re-placed \textit{back hall by fridge} at tasks 17, 35, 54 and 70, at
map positions (4.54, $-$3.53), (1.34, $-$5.51), (0.01, 0.02) and
($-$2.12, $-$3.41), and placed \textit{galley kitchen} at task 42; the
no-consolidation run named the place afresh each time (\textit{kitchen},
\textit{kitchen entry}, \textit{kitchen fridge corner}, \textit{fridge
passage}, \textit{fridge end wall}, and four labels across the house on
its failed task 35). The no-consolidation run found the calendar at task 40 in 283 steps after a
first close from the doorway was refused, and at task 54 it overshot in
long turn-and-walk batches. At task 42 the repaired marker's own label
pulled the full run toward the wrong goal (``trusting the `back hall by
fridge' marker as a shortcut misled me'').

\begin{figure}[!htb]
\centering
\includegraphics[width=0.8\linewidth]{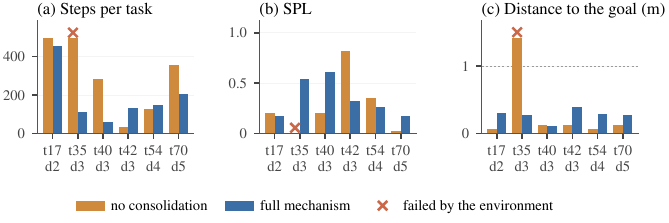}
\caption{House C, the calendar and the refrigerator beside it, every task in both runs: steps, SPL and distance to the goal by the simulator (dashed line, the 1\,m success radius; a cross marks a failure).}
\label{fig:memcmp-backhall}
\end{figure}
\begin{figure}[!htb]
\centering
\includegraphics[width=1.0\linewidth]{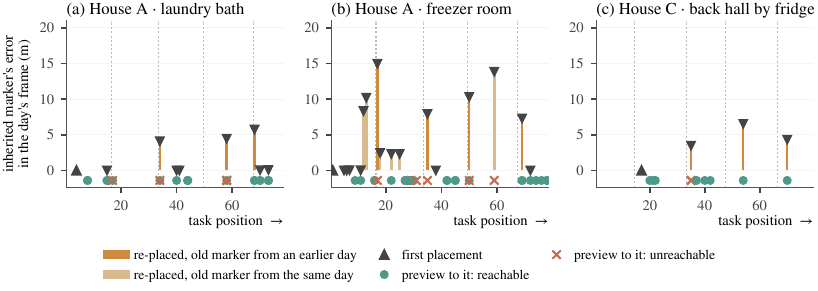}
\caption{The life of one label in the frame of each day. Bars are the inherited marker's error in the frame of the day at the moment it was re-placed (ochre when the old marker came from an earlier day, pale when from the same day); triangles mark placements; the dots and crosses under the axis are previews aimed at that label, reachable (teal) or refused (red-brown). A marker is accurate within its day and wrong at the start of the next; previews fail until it is re-placed. (a) and (b) House A; (c) House C.}
\label{fig:memcmp-lineage}
\end{figure}

\begin{ledger}[barwarn]{House C · task 35 · day 3 · full · task record}{Find any calendar.}{113 steps; the marker moved 0.02\,m in truth}
I found it and stood in front of it, the picture panel above the dated grid on the panelled fridge surround where the back hall opens off the kitchen [\ldots] \hlw{Path previews called every kitchen-side marker unreachable --- ignore that.} [\ldots] \hlm{Moved the ``back hall by fridge'' marker to where the calendar is visible.}
\end{ledger}
\begin{ledger}[barwarn]{House C · task 35 · no-consolidation · task record}{Find any calendar.}{496 steps, one recovery, failure 1.42\,m from the goal}
Looked for any calendar and did not find one. Starting at the piano corner, I worked through the open living and dining area, into the kitchen, then down the hallway [\ldots] Several hallway doors stayed shut; I ran out of steps before opening or reaching them, and \hlw{white paper on the hall wall turned out not to be a calendar}.
\end{ledger}

\begin{insight}
\textbf{What it shows.} Re-placement brings a marker into the current coordinate frame, after
which it can guide the next visit (58 steps at task 40). At task 35 the
session instead reads the calendar skill and four stored photographs
while the marker remains unreachable. Task 42 shows the opposite
behavior: following the label adds about one hundred steps. These records
distinguish useful room recognition from unreliable metric guidance.
\end{insight}

\paragraph{Case 4. Narrowing a label over repeated visits.}
In House C the label \textit{piano corner} was placed on day 1 and re-placed
six times. Its map coordinates jumped by 10 to 13\,m at each day start;
by the simulator the placements lie within 2.6\,m of the goal, and the last
three narrow from 1.64 to 1.30 to 0.00\,m (Figure~\ref{fig:memcmp-narrowing}).
The one placed at task 72 recorded the executor's own habit of standing
back to frame the piano, and that task failed 1.27\,m from the goal while
the no-consolidation run succeeded in 90 steps; the account blamed a drift that the
day's sheet did not contain. Two tasks later the executor re-placed the
label at the piano itself and said why; at tasks 74 and 78 both runs
succeed, the full run in 136 and 71 steps against 171 and 87.
\begin{figure}[!htb]
\centering
\includegraphics[width=0.7\linewidth]{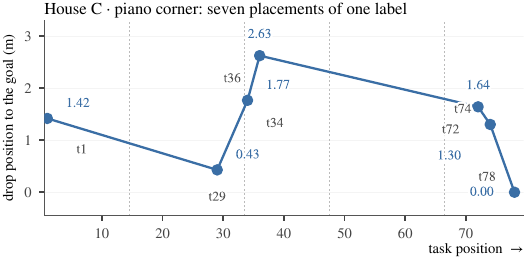}
\caption{House C, the \textit{piano corner} label: distance from each placement to the goal by the simulator. The map coordinates of the same placements differ by up to 13\,m between days; the true positions lie within 2.6\,m, and the last three narrow to the goal.}
\label{fig:memcmp-narrowing}
\end{figure}

\begin{ledger}[barwarn]{House C · task 72 · day 5 · full}{Find the piano.}{message on arrival, then the task record; the run ends 1.27\,m from the goal}
I'm at the piano corner, \hlw{parked about three paces back} and facing it square [\ldots] The 'piano corner' marker has been re-dropped here --- \hlw{the old stored marker had indeed drifted, as the notes warned.}\par Ledger: [\ldots] \hlm{Re-dropped the ``piano corner'' marker.}
\end{ledger}
\begin{ledger}[barmem]{House C · task 78 · day 5 · full}{Find the specific piano shown in the goal image.}{message, the task record, then the piano skill as consolidated that evening; the re-placement ends 0.08\,m from the goal}
I'm standing right at the piano from the goal photo. Let me \hlm{correct the drifted marker to the true spot}.\par Ledger: [\ldots] I re-dropped the ``piano corner'' marker at the true spot, \hlm{since the old one had drifted and pointed me at empty floor}.\par skills/find-a-piano.md, day 5: The stored 'piano corner' marker had drifted onto empty floor.
\end{ledger}
\begin{ledger}[barok]{House C · task 72 · no-consolidation · task record}{Find the piano.}{90 steps, success 0.08\,m}
I was sent to the piano and reached it: I ended up \hlo{standing right beside the upright piano and its bench} by the window, in the log-walled living room. [\ldots] I dropped markers at the kitchen sink island, the log dining room, and the piano corner.
\end{ledger}

\begin{insight}
\textbf{What it shows.} Re-placing a named marker updates its stored location. The final
placements approach the goal from 1.64 to 1.30 to 0.00\,m over tasks 72,
74, and 78. The position records suggest that an earlier marker
represented a viewing position rather than the object; the quoted
account itself describes drift and empty floor. The full run fails
task 72, which no-consolidation completes, but takes 71 rather than 87 steps on
task 78. Across the seven piano tasks it averages 74 steps against 122.
Marker updates and note-based recognition accompany this improvement,
without guaranteeing a shorter search on every visit.
\end{insight}

\FloatBarrier
\subsubsection{Recording failure and checking completion (Cases 5 to 7)}

\paragraph{Case 5. One sentence, read through a note.}
In House A the description ``wooden dresser with a mirror on it, to the
right of the hunting trophy, above the staircase handrail, next to the
shelf'' is asked three times. The no-consolidation run writes an honest exclusion
list each time and only on the third attempt, after 493 steps and a
recovery, records that ``the dresser was on the basement level all along,
near the stairs I started beside'', although its own successful accounts of two
earlier dresser tasks already contained the answer. At task 56 both runs start inside the freezer room, about one to two
metres from the goal, and at task 69 both start beside the boiler
(Figure~\ref{fig:memcmp-dresser}). The full run placed no marker on
these tasks except \textit{freezer room} at task 69, and its successful end
poses at tasks 56, 67 and 69 lie within 0.22\,m of one another; the no-consolidation
run left three labels on each failed search (\textit{front entry foyer},
\textit{office bedroom}, \textit{basement sitting room} at task 10;
\textit{red rug freezer room}, \textit{main floor dining}, \textit{upstairs
kitchen} at task 56). The full run reaches it
through the journal, a room note and a skill: the day-1 room note carried
``a tall wooden dresser with a mirror'', the day-3 skill added ``Correction
from t10: no mirror was visible from this side'' after the message of task
43 (``the mirror detail may be stale''), and the day-4 skill made the
sentence itself the key.

\begin{figure}[!htb]
\centering
\includegraphics[width=0.8\linewidth]{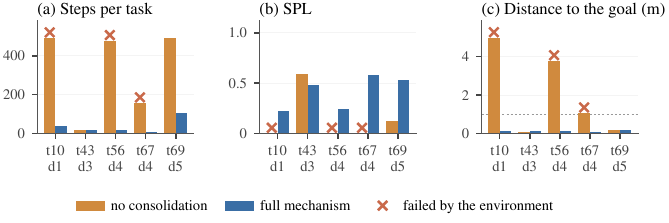}
\caption{House A, the mirrored dresser, every task that asked for it, in both runs: steps, SPL and distance to the goal by the simulator (dashed line, the 1\,m success radius; a cross marks a failure).}
\label{fig:memcmp-dresser}
\end{figure}

\begin{ledger}[barok]{House A · skills/find-dresser.md · day-4 version, then the opening messages of tasks 56 and 69 · full}{Find this specific dresser: wooden dresser with a mirror on it \ldots}{}
Goal texts for it can mislead: [\ldots] certified to this dresser (09-22 t7), even though no mirror is visible from this side. \hlo{Identify by the antler plaque above, the yellow tool shelf beside, the corner freezer past it --- never by a mirror.}\par Task 56: [\ldots] (\hlm{the hunting trophy = antlers is a freezer-room landmark}).\par Task 69: \hlm{My notes say this exact goal text certified to the freezer-room dresser.}
\end{ledger}
\begin{ledger}[barwarn]{House A · task 56 · no-consolidation}{The same goal.}{mid-task messages in order, then the verifier; 478 steps, one recovery, ends upstairs 3.78\,m from the goal}
\hlw{Found the up-staircase with handrail. Going up.} I see what looks like antlers (hunting trophy) on the upper-left wall. I've searched the whole main floor.\par Verifier: the only dresser-like furniture seen (frame 7, near the mounted trophy) was passed mid-route and left behind.\par Ledger: [\ldots] \hlw{The only antlers hung high on the entry wall, with nothing beneath.}
\end{ledger}

\begin{insight}
\textbf{What it shows.} The no-consolidation run records useful information but does not
reuse it in these later dresser searches.
The phrase \emph{above the staircase handrail} sends it upstairs three times.
The full run instead reads a note connecting that wording to the
freezer-room dresser. An observation at task 43 passes through the task
account and consolidation into a skill that later sessions cite.
At task 56 both runs start in the same room, and the full session reads
the note before moving without first querying the map. The retrieved
description resolves the floor and target before route planning begins.
\end{insight}

\paragraph{Case 6. Arriving already there, and accepted elsewhere.}
In House B, the small gap in success rates contrasts with a much larger
gain in path efficiency. Retained memory reduces the movement needed
to recognize and reach familiar targets (Figure~\ref{fig:memcmp-closet}):
18\% of tasks close within ten steps with the full mechanism, 8\% without,
and on the first day, before any notes exist, the whole difference is one
line of the map, the marker's coordinates equalling the current pose
(task 2: ``I woke standing on my `walk-in closet' marker from yesterday,
facing the rails'', 0 steps, SPL 1.00; the marker was in fact dropped at
task 1 the same day). Task
62 is the extreme: the house has two closets with hanging clothes, the
no-consolidation run stands in the wrong one, closes after four steps and is accepted
from one frame; the full run wakes on the within-day marker of the right
one and closes with no step, although the skill note it had read pointed to
the wrong closet (``go to the `walk-in closet' marker \ldots{} expect it may
be stale''). The day-4 consolidation later wrote ``Do not head for the
09-19 `walk-in closet' marker first.''

\begin{figure}[!htb]
\centering
\includegraphics[width=0.85\linewidth]{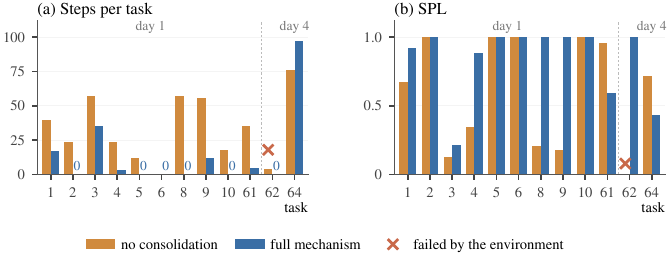}
\caption{House B, the twelve hanging-clothes tasks given in words, in both runs: the first day's block (tasks 1 to 10; task 7, whose goal is an image, is left out), then tasks 61, 62 and 64 on day 4. \textbf{(a)} Steps per task, zero-step closures labelled; the cross marks the no-consolidation run's task 62, closed after four steps in the wrong closet, accepted by the verifier and 9.8\,m from the goal. \textbf{(b)} SPL. At task 62 the runs start in different closets, about 8.8\,m apart, because task 61 diverged.}
\label{fig:memcmp-closet}
\end{figure}

\begin{ledger}[barok]{House B · task 62 · day 4 · full}{Find the hanging clothes in the closet.}{messages in order, then the task record; 0 steps, success 0.19\,m}
\hlo{I'm standing exactly on my 'upstairs closet rail' marker from a previous task}, and the camera confirms it: coats, shirts and jeans hanging on rails [\ldots] \hlo{This is the same spot that was certified for a hanging-clothes goal before} --- no movement needed.\par Ledger: \hlo{Worth noting my older skill note pointed at the grey bedroom's walk-in closet instead, which was unnecessary here.}
\end{ledger}
\begin{ledger}[barwarn]{House B · task 62 · no-consolidation}{The same goal.}{messages, the completion claim, the verifier, then the task record; 4 steps, accepted, 9.79\,m from the goal}
The map is nearly blank and no markers survive. Starting fresh.\par Verifier: The final frame is a close-up of densely packed hanging garments on a rack inside the closet, \hlw{matching the target}\par Ledger: [\ldots] \hlw{The map came up nearly blank with no surviving markers, so I had no prior route to reuse; nothing else was attempted.}
\end{ledger}

\begin{insight}
\textbf{What it shows.} In the first day's block, the two runs start within a metre of one
another. The full session recognizes a familiar marker and checks its
view, avoiding additional translational search. About a third of House
B's SPL gain occurs in this block. Task 62 has different starting
locations and illustrates a separate issue: the verifier accepts the
wrong closet in the no-consolidation run. The full session recognizes its current
within-day marker, sets aside a conflicting skill note, and records
the conflict for consolidation. Similar overall success rates can
therefore conceal substantial differences in search efficiency.
\end{insight}

\paragraph{Case 7. A false completion certified into memory.}
This case shows a mistaken object identification surviving the closing
checks and entering memory. In House A, the checker accepted false
completions on 10\% of tasks without consolidation and 3\% with it.
The no-consolidation trace below contains a first-request refusal followed by a
repeated claim, which the stopping rule permits
(Appendix~\ref{app:closure}). With memory, one false completion enters
the journal as \CERT{}, is written into the living-room note (``CERTIFIED, 09-20 t16: a wooden
chest of drawers with a table lamp on it'') and the dresser skill (``the
answer to `dresser next to the armchair'\,''), and is repeated at zero cost
by the next task, which reads no file; on day 5 it pulls task 78 26 steps
off course (``that matches my living-room `dresser next to the armchair'
note. Let me approach'').

\begin{figure}[!htb]
\centering
\includegraphics[width=0.85\linewidth]{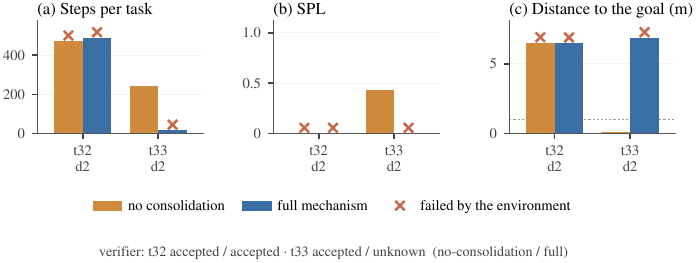}
\caption{House A, the lamp table beside the armchair. Both runs close task 32 on the same object and task 33 starts where task 32 ended; the verifier's decision on each closure, for the run without
consolidation and the full run, is given beneath the panels.}
\label{fig:memcmp-lamptable}
\end{figure}

\begin{ledger}[barwarn]{House A · tasks 32 and 33 · day 2 · full}{Find this specific dresser: dresser next to the armchair. / Find any dresser.}{completion claim, ledger index entry, the next task's first message, then the verifier}
I am standing beside a wooden chest of drawers with a table lamp on it, directly next to the black leather recliner armchair.\par Journal: \hlw{task 16 · CERTIFIED} · [\ldots] markers: living room\par Task 33: [\ldots] \hlm{this is the living room corner from task 16.}\par Verifier: reads more like a nightstand/end table than a dresser.
\end{ledger}
\begin{ledger}[barwarn]{House A · task 32 · no-consolidation}{Find this specific dresser: dresser next to the armchair.}{the pre-stop refusal, a second claim accepted, then the task record; failure 6.51\,m}
Gate: no dresser and no armchair adjacent to it [\ldots]\par Ledger: I closed it as complete at the black leather recliner in the living room, with the lamp chest beside it under the framed picture --- but \hlw{the checker disagreed once and I overrode it, so treat that identification as shaky}.
\end{ledger}

\begin{insight}
\textbf{What it shows.} Here, an incorrect certification lets an object
identification be retained as fact in the house notes. The next task's
``unknown'' verdict records uncertainty about that task but does not
automatically retract the earlier account. \end{insight}

\FloatBarrier
\subsubsection{Instruments and procedure (Cases 8 and 9)}

\paragraph{Case 8. Diagnosing the same fault once, or ten times.}
In House A the tracker snaps the pose back to the origin after a loss. Both
runs recognise that the map is right and the position wrong. The no-consolidation
run diagnoses it afresh in ten tasks and prescribes contradictory remedies;
the full run carries forward the advice to follow landmarks when the
preview is unreliable. The more precise explanations occur in a refused
recovery request, Edit, and Write, while the retained skill still says
\textit{Markers here drift}. At task 35 a refused edit
named ``a frame offset from the old floor-1 map'', and at task 78 the run
rode out a frozen pose on sight: ``my pose readout is frozen but the views
genuinely change --- I'll trust the views''.

\begin{ledger}[barmem]{House A · tasks 29, 34 and 70 · full}{}{three correct diagnoses, three refusals; the surviving note is quoted last}
Task 29: [\ldots] \hlo{the map merges both half-levels.}\par Task 34, recovery request, refused: \hlo{the robot created and placed me on `floor 2'} [\ldots]\par Task 70, refused Write: \hlo{the marker has NOT drifted.} [\ldots] Rule: use the marker to know you have ARRIVED, never to choose a heading.\par skills/verify-markers-before-trusting.md, day 5: Even a marker that was accurate all yesterday can go stale.
\end{ledger}
\begin{ledger}[barmem]{House A · task 28 · day 2 · full}{Find any boiler.}{message, the task record, then the rule it became: the first rung of the ladder}
\hlm{my pose jumped west} while I walked into a previously-unmapped area.\par Ledger: SLAM lost itself on the way in and the marker's position looked wrong; \hlm{recovery changed nothing}, so trust the route, not the marker.\par skills/ignore-route-preview.md, day 2: If SLAM appears lost, keep walking the landmark chain rather than trying recovery --- recovery changed nothing in t12.
\end{ledger}
\begin{ledger}[barwarn]{House A · task 71 · no-consolidation · task record}{Find the specific boiler shown in the goal image.}{429 steps, 204 turns, SPL 0.01; the remedy contradicts the one of task 49}
I never confirmed I reached it --- movement kept stalling, with forward steps colliding and \hlw{my position repeatedly snapping back to where I started}, so the same laundry-nook view came round again and again. [\ldots] \hlw{Turning in place to find a clear opening mostly did not help.}
\end{ledger}

\begin{insight}
\textbf{What it shows.} Both runs diagnose recurring tracking faults. The full run
reuses an earlier diagnosis through the preview skill derived from
task 28. It records seven recovery requests over the deployment against
thirteen without consolidation. At task 78 it proceeds using changing camera
views despite a frozen pose, whereas the no-consolidation run requests recovery
on a similar fault at task 71 and spends 429 steps at SPL 0.01.
These are examples from different tasks, not a matched-task effect
estimate. They show how a stored procedure changes the response to a
recurring instrument problem.
\end{insight}

\paragraph{Case 9. A spurious floor, diagnosed but lost from the summary.}
Only a run with memory meets this fault: markers persist across days, but
each day SLAM opens a new frame and the floor counter increments, so every
old marker falls in a dead frame and the preview reports it unreachable. In
House B the executor wrote the fault up as a bug report the moment it
occurred on day 2, and by day 4 had the mechanism (task 52: ``the coordinate frame
shifted between sessions \ldots{} it points at a location in a dead
frame''), the cure and ``do not request recovery for this''.
The detailed diagnosis appears in attempted edits that were refused,
whereas the saved task account below records only an unreachable marker
and the use of landmarks. The later skill note retains a remedy without
the explanation of the frame change. Consolidation can read the ledger,
task handovers, and existing notes (Appendix~\ref{app:memory});
the failure here is loss of that explanation from the retained summary,
not a restriction to reading the ledger alone.

\begin{ledger}{House B · task 20 · day 2 · full}{Find any microwave.}{recovery request at the moment of the fault, the refused journal draft, then the task record that reached the journal}
After only turning in place, get\_map switched from `floor 1/1' (127 m2, all my markers) to a brand-new `floor 2/2' with 17.85 m2 and no markers, and preview\_path to the kitchen marker went from reachable (12.5 m) to unreachable.\par Refused journal draft: The map now shows \hlw{TWO floor layers} [\ldots]\par Ledger: \hlw{My old kitchen marker was unreachable, so I navigated by landmark instead.}
\end{ledger}
\begin{ledger}[barwarn]{House B · task 54 · day 4 · full}{}{refused Edit, then the first line of the skill note as consolidated on day 5}
WHOLE-FRAME SHIFTS HAPPEN, not just single stale markers. [\ldots] \hlo{So the map origin, not the furniture, had moved.} [\ldots] \hlo{Do NOT call request\_recovery for this --- nothing about the robot's position is wrong.}\par Day 5 note: Saved marker positions go stale. [\ldots] Ignore the preview, walk in by landmark, re-drop under the same name where you finish.
\end{ledger}

\begin{insight}
\textbf{What it shows.} The executor diagnoses the frame change three times, with increasingly
specific explanations, but the retained note keeps the remedy rather
than the cause. The recorded layer changes do not correspond to actual
storeys. A warning intended for a frame reset consequently becomes a
general distrust of route previews, even when the current frame is
usable. The traces include extra steps and one recovery at task 20,
a failure at task 26, and four to eight preview probes per task on
day 4. Summaries need to preserve the conditions under which a remedy
applies, not just the remedy itself.
\end{insight}

\FloatBarrier
\subsubsection{Auditing and consolidating memory (Cases 10 and 11)}

\paragraph{Case 10. Closing an open question.}
The question was not the executor's; the day-1 consolidation of House B set
two descriptions of the kitchen refrigerator side by side and asked whether
they were one appliance seen from two sides or two (``still open:
whether the kitchen holds one fridge or two''; house.md: ``possibly one
appliance seen from two sides, possibly two''). On day 2 the executor
read the question, finished its task, read the overview three times more and
went to look at both faces of the appliance. The answer travelled through
the ledger, the day summary and a skill without loss. Two days later a very
dark goal photograph arrived; the no-consolidation run reasoned from darkness to a
basement and searched for a second refrigerator, ending beside the target. The full run placed \textit{kitchen dining
doorway} at task 21 and \textit{fridge nook} at task 36; the no-consolidation run
placed \textit{kitchen} or \textit{kitchen fridge} on each success and, on
its failed task 36, four labels ending at \textit{basement stairhead}. At
task 21 the full run spends 36 of its 56 steps answering the open question; at task 36 both runs start from the
same place (Figure~\ref{fig:memcmp-fridge}), and the full run's ledger
line reads ``the old notes were right that you should come at it from the
dining-doorway side rather than head-on from the range'' (36 steps, SPL
0.70).

\begin{figure}[!htb]
\centering
\includegraphics[width=0.8\linewidth]{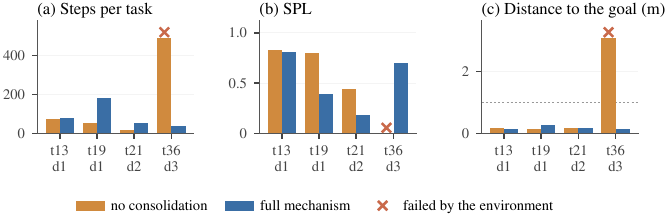}
\caption{House B, the kitchen refrigerator, every task that asked for it, in both runs: steps, SPL and distance to the goal by the simulator (dashed line, the 1\,m success radius; a cross marks a failure). At task 36 the goal is a dark photograph.}
\label{fig:memcmp-fridge}
\end{figure}

\begin{ledger}[barok]{House B · task 21 · day 2 · full}{Find this specific refrigerator: \ldots}{messages after the task was complete, the refused write, then the task record; house.md was read three more times}
\hlo{I'll resolve the long-standing open question in my notes}, then close out. [\ldots]\par Refused write: the ``tall dark fridge by the doorway'' (task 13) and the ``stainless side-by-side with ice and water dispenser'' (task 19) are \hlo{one appliance, not two}. [\ldots]\par Ledger: That dark panel and the stainless front are one fridge, not two.
\end{ledger}
\begin{ledger}[barwarn]{House B · task 36 · no-consolidation}{The same goal.}{messages, then the task record; 490 steps, one recovery, failure 3.08\,m}
The goal image is very dark [\ldots] \hlw{a very dark refrigerator in an unlit space --- almost certainly a basement, not this bright kitchen}.\par Ledger: I did not find it. [\ldots] the kitchen off it whose fridge is bright white and clearly not the one pictured [\ldots] \hlw{The dark fridge is most likely down those basement stairs.}
\end{ledger}

\begin{insight}
\textbf{What it shows.} Consolidation turns conflicting descriptions into an open question.
The executor gathers additional views to resolve it, and the answer
survives subsequent summaries. At task 36, it recognizes a dark goal
photograph as another view of the known refrigerator, while the no-consolidation
run searches for a second appliance. The runs take 36 and 490 steps,
respectively. The trace connects the stored answer to goal
interpretation, but does not partition this 454-step difference between
the note, the map, and the rest of the retained state.
\end{insight}

\paragraph{Case 11. Overturned rules, and a route nobody read.}
Each of the seven skill notes of House A has a version history, and each
overturning can be asked where its evidence came from. The three driven by
the executor's own observations (the carpet stopping rule, the mirror's
visibility, the bare-lid criterion) were right; the one driven by a verifier
verdict was wrong, since after one ``unknown'' on a goal the simulator had
certified three times the day-5 consolidation reversed the rule and now
sends that wording to the other floor; the two driven by the harness's own
artefacts (a ``dead end'', a ``drift'') kept a right prescription with a
wrong reason. The no-consolidation records contain useful experience that later
sessions do not reuse in these searches: the route to the boiler, written
in full at task 8, was never read, and
the three later boiler tasks each spent their whole budget on a fresh
hypothesis (task 15: ``the storage space past the workbench corner is the
one utility nook I never entered''; task 58: ``the ceiling ductwork runs
toward it, so the boiler is likely behind it''). Over the nine boiler tasks
the no-consolidation run succeeds on 89\% at 274 steps per task, the full run on all
nine at 107.

\begin{ledger}[barwarn]{House A · skills/find-dresser.md · day-5 version, and the verifier verdict at task 69 that drove it · full}{}{the simulator had confirmed the freezer-room dresser three times}
Certified once for exactly that wording (09-22 t7); no mirror is visible on it, and a repeat attempt on 09-23 t2 went unresolved. \hlw{If the text names both a mirror and the trophy, prefer the window-bedroom chest and note the conflict.}\par Verifier, task 69, unknown: [\ldots] \hlw{but no mirror or recognizable full dresser is visible}.
\end{ledger}
\begin{ledger}[barok]{House A · task 65 · day 4 · full}{Find the specific freezer shown in the goal image.}{the bare-lid rule as read, the messages, the task record, then the day-4 consolidation}
Skill, day 3: [\ldots] the certified answer was the right-hand one, not the corner one.\par Close-up lid [\ldots] my notes say the right-hand freezer. [\ldots] \hlo{that matches the corner freezer under the yellow shelf}.\par Ledger: My ``bare lid means right-hand freezer'' rule nearly misled me; \hlo{the corner and rust spots decided it}.\par Day-4 consolidation: [\ldots] treat that rule as doubtful and let rust spots and the corner decide.
\end{ledger}
\begin{ledger}[barok]{House A · task 8 · no-consolidation · task record}{Find this specific boiler: gas boiler \ldots below the heater and slightly to the right \ldots}{a complete route not reused in the later boiler searches}
Found it and stood beside it --- the upright cylinder with its gas valve, under the furnace ducting, next to the washer and dryer. \hlo{Route went from the freezer alcove where I started, down the hall past a doorway, then right into the laundry utility room at the end.}
\end{ledger}

\begin{insight}
\textbf{What it shows.} Source references make the revisions auditable. In these examples,
direct observations correct misleading rules, while an uncertain
verifier output can cause consolidation to replace a useful rule with
an incorrect one. The no-consolidation run also writes a complete boiler route,
but subsequent boiler searches do not reuse it. Across the boiler tasks, the full
run combines a remembered landmark route with current markers and
averages 107 steps rather than 274. This is a system-level contrast,
not a measured step saving attributable to one unread sentence.
\end{insight}

\FloatBarrier
\subsection{When stored experience is declined (Cases 12 to 14)}
\label{app:declined}

The following cases illustrate three forms of selective reuse, each with
a no-consolidation counterpart: a route declined on a prior from the
notes (Case 12), a rule declined because the goal contradicts it (Case 13),
and a bad entry caught by its consequences and repaired from the journal
(Case 14). The control retains maps and task records but has no consolidated
house notes. Its quoted doubts concern pose and preview within the task.

\paragraph{Case 12. Declining a correct marker and following a remembered route.}
At task 70 of House A the marker is accurate to 0.06\,m and the preview says
reachable; the executor declines the straight line on the prior in its
notes that this marker ``has drifted before'', walks the landmark chain, and
corrects its reason on arrival (the refused write: ``my old note \#8
claimed the `laundry bath' marker had drifted. It had *not*''); the ledger
line keeps the recantation (``the marker is true, but only the hallway route
gets you there''). Both
runs start beside the freezer-room dresser and both reach the boiler within
0.2\,m; the full run takes 94 steps (SPL 0.80) and no recovery, the no-consolidation
run 317 steps, one recovery and SPL 0.20.

\begin{ledger}[barmem]{House A · task 70 · day 5 · full}{Find any boiler.}{mid-task messages in order, after the notes were read and the preview reported the marker reachable at 4.7\,m}
Reachable at 4.7 m --- \hlm{but my notes warn this marker has drifted before}. [\ldots] The marker points straight into the dresser wall --- it has drifted, exactly as my notes warned. Falling back on the landmark route. [\ldots] The yellow slatted pine door is on my left --- my key landmark. [\ldots]\par On arrival: \hlo{The marker was accurate all along (0.36 m away)} --- my straight-line bearings were cutting through walls.
\end{ledger}
\begin{ledger}[barwarn]{House A · task 70 · no-consolidation}{Find any boiler.}{mid-task message after a tracking loss, then the task record; 317 steps, one recovery, SPL 0.20}
\hlw{My reported position is clearly wrong} --- the map says I'm at the rec room origin, but I'm in the west plaid-couch room.\par Ledger: [\ldots] \hlw{Early on I lost my bearings against a wall near the plaid-couch room and wasted turns re-orienting before the hallway route worked.}
\end{ledger}

\begin{insight}
\textbf{What it shows.} The session initially distrusts a correct marker because an older note
warns about drift and the direct bearing crosses a wall. It follows the
remembered landmark route and revises its diagnosis on arrival. Both
runs succeed, in 94 and 317 steps, respectively. The trace shows that
a useful route can coexist with an incorrect explanation for choosing
it. It does not establish that rejecting the marker has zero cost
relative to using it with a valid route.
\end{insight}

\paragraph{Case 13. Declining a rule the goal photograph contradicts.}
Consolidation can turn ``I did not need to'' into ``do not''. In House C
the account of task 56 said two floor plants in a corner were too tight to
frame and that it had chosen another; the day-4 skill made this an
\textit{Avoid}. The goal photograph of task 67 is one of those plants. Both
runs start at the episode's reset point; the full run succeeds in 123 steps
(SPL 0.75, 0.17\,m from the plant); the no-consolidation run fails.

\begin{ledger}[barmem]{House C · task 56 · day 4 · full · task record}{Find any decorative plant.}{the account that the consolidation turned into a prohibition}
I started in the treadle desk nook in the northeast corner; its two floor-level plants --- a dried arrangement in a wooden bucket and a low leafy pot by the picture window --- \hlm{sat too tight in the corner to frame, so I left them}. I crossed [\ldots] to the dining window nook.
\end{ledger}
\begin{ledger}[barwarn]{House C · skills/find-a-decorative-plant.md · day-4 version, then tasks 67 and 73, then the day-5 version · full}{Find the specific decorative plant shown in the goal image.}{}
\hlw{Avoid:} the two floor-level plants [\ldots] sit too tight in the corner to frame --- 09-22 t9 left them [\ldots]\par Task 67: \hlm{The old note calling that corner too tight to frame was wrong}; standing a step off the desk frames it fine.\par Task 73: [\ldots] \hlw{I could not save that correction; the note-writing tools were disabled, so the stale advice is still on file.}\par Day 5: The old warning [\ldots] is dead --- stop about a step short of the desk and it frames (09-23 t7, borne out by t9/t13).
\end{ledger}
\begin{ledger}[barwarn]{House C · task 67 · no-consolidation · task record}{The same goal.}{470 steps, one recovery; closed 7.1\,m from the goal, verdict unknown}
\hlw{The matching plant sat on a wooden chest beside a yellow lamp in the bright window parlor past the dining room, and I closed there.} [\ldots] I wasted many steps circling the kitchen and hearth area before trying the far south rooms.
\end{ledger}

\begin{insight}
\textbf{What it shows.} The session uses the stored room description and photograph to
recognize the corner, but rejects the instruction to avoid it because
the new goal explicitly asks for that plant. It records this conflict
in the task account, and consolidation revises the rule. The important
behavior is selective reuse: the room information remains useful even
when the associated instruction is inappropriate for the current goal.
\end{insight}

\paragraph{Case 14. A note that misleads, and a journal that recovers.}
On day 4 the notes of House B mention one mirror, a black-framed one in
the entry hall. The goal image was returned at the start, again before a
recovery and once more right after it, yet the executor followed the note
downstairs and then wrote that it had not seen the photograph in this
session; only after a fourth call did it look at the goal
(``re-reading the actual goal photograph instead of trusting my notes''),
see that it was not that mirror, and find in the same day's journal the
failed attempt at the same goal five tasks earlier, which left an exact
lead. Memory was both the cause and the cure. The starts differ by
1.2\,m, both in the same guest room.

\begin{ledger}[barmem]{House B · task 60 · day 4 · full · task record, as it entered the journal}{Find the specific mirror shown in the goal image.}{500 steps, one recovery, failure 1.21\,m}
I hunted for a wooden-framed mirror from a photo and did not find it. I woke upstairs on the bedroom floor, not the main floor my old notes describe [\ldots] Past the master's door a hardwood corridor runs by a descending staircase to a teal bed guest room, which I marked; I ran out of steps before seeing the wall behind its open door. \hlm{Try that wall first.}
\end{ledger}
\begin{ledger}[barwarn]{House B · task 65 · day 4 · full}{The same goal, five tasks later the same day.}{recovery request at step 241, then the messages that followed}
I still haven't actually seen the goal photograph in this session --- \hlw{I've been reasoning from notes.} [\ldots] the real goal image: a pine-framed square mirror on a slate blue-grey wall [\ldots] not the entry hall (black frame, cream walls).\par Reading the journal: \hlo{Task 9 was this same mirror goal, and it left a precise lead}: `[\ldots] I ran out of steps before seeing the wall behind its open door.' [\ldots] I need to go back up.
\end{ledger}
\begin{ledger}[barok]{House B · task 65 · no-consolidation · task record}{The same goal.}{156 steps, SPL 0.38; it never went downstairs}
\hlo{I found it in the bedroom with the red bookshelf}, hanging on the blue-grey wall next to a small framed print, opposite the open door to the hall --- I stopped facing it and closed complete. [\ldots] Wandering the first room cost steps for nothing.
\end{ledger}

\begin{insight}
\textbf{What it shows.} The session treats an incomplete inventory as exhaustive, following
the only mirror in its notes despite having received the goal image.
After recovery it re-examines that image and retrieves task 60's
unfinished-search lead from the journal. This lead directs the search
that ends in a 468-step success. The full run is nevertheless less
efficient than no-consolidation (SPL 0.08 versus 0.38). Detailed failure records
help repair a search misled by a summary, without showing that another
failure was inevitable in their absence.
\end{insight}

\paragraph{How often, and was it right.}
Every statement in the full run's accounts, messages, claims, recovery
requests and refused writes that declares an inherited item wrong,
unreliable or inapplicable was counted and checked against the simulator
(Figure~\ref{fig:memcmp-declined}). This analysis concerns judgements recorded
by the full runs, rather than the availability of memory in the control.
In all three houses a judgement is right with the wrong cause when the
offset or unusability it names is real but blamed on drift or a wrong
floor. In House A, of
seventeen doubts of an inherited marker, nine were right, four right in
effect with the wrong cause and four wrong; of eight doubts of a note's
rule, route or target, three were right, three wrong and two cannot be
checked. In House B all 21 checkable judgements hold in effect, two with
the wrong cause; in House C, of 17 checkable judgements (19 in all),
eight hold, seven hold with the wrong cause and two are contradicted
(tasks 36 and 72).

\begin{figure}[!htb]
\centering
\includegraphics[width=0.85\linewidth]{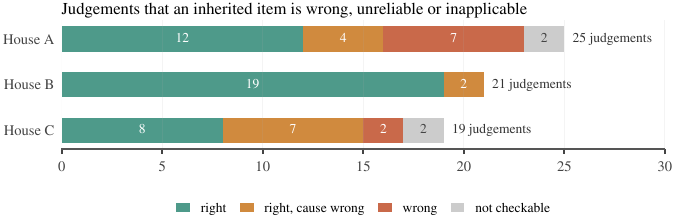}
\caption{Judgements in the full run that an inherited item is wrong,
unreliable or inapplicable, checked against the simulator. Counts describe
the full runs' recorded judgements.}
\label{fig:memcmp-declined}
\end{figure}

Declining an inherited coordinate is almost always right in effect and
almost never right about the cause: every account of a day-start frame
reset calls it drift or a wrong floor. Three wrong judgements doubted an
accurate marker (House A, tasks 3, 40 and 70) and one cost a task (House A,
task 27). Only written judgements are counted; a memory silently not used
leaves no account, and the simulator checks position only.

\paragraph{Memory use across the analyzed cases.}
For each of the forty-eight cases, we manually coded the recorded memory
effect as positive, negative or mixed, or showing no difference.
We also labeled the records involved, allowing multiple labels for notes,
within-day markers, and the ledger, with a separate label for harness or
judge involvement. Figure~\ref{fig:memcmp-credit}
summarizes these case-level interpretations. They describe the analyzed
corpus, not independent ablation effects or rates over all tasks.

\begin{figure}[!htb]
\centering
\includegraphics[width=0.95\linewidth]{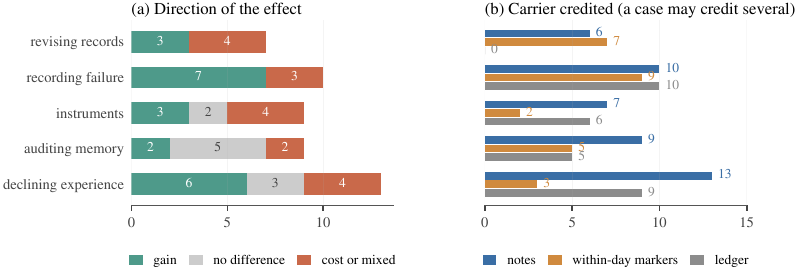}
\caption{Manual coding of the forty-eight analyzed cases, including the
fourteen presented here. (a) The direction of the recorded memory effect.
(b) Memory sources involved in the case assessment, with multiple labels
allowed. Counts describe cases, not task-level component contributions.}
\label{fig:memcmp-credit}
\end{figure}
\paragraph{What the comparison shows.}
Four kinds of note content guide the recorded decisions: a rule that maps a
goal sentence to an object (Cases 5 and 10), a room's inventory and
photographs (Cases 3 and 13), a landmark chain in a skill (Cases 11
and 12), and a spot recorded as certified, which outweighed a wrong skill
(Case 6). These textual cues work alongside the map and the robot's
current location. Within-day
markers carry one kind: waking on the goal, or a marker reachable in the
day's frame (Cases 2, 4 and 6). The ledger is the carrier that turns a
day's observation into the next day's note (Cases 5, 10 and 13), and also
the channel through which a wrong observation becomes a note (Cases 7
and 11). The records also show five risks of reusing memory: inherited
coordinates after a day-start frame reset
(Cases 1, 2 and 3), a spurious floor (Cases 8 and 9), an accepted false
completion written as a certified fact (Case 7), a gap in the notes or a
failed account's lead followed into a dead end (Case 14), and a prohibition
generalized from a ``no need'' (Case 13). These behaviors help interpret
the different profiles in Figure~\ref{fig:memcmp-glance}, without
assigning each profile to a single memory component.
\FloatBarrier
\end{humanconfirmed}

%% file: sections/memcmp_tab_boundary.tex
\begin{table}[!htb]
\caption{Deployment instructions and observed behaviors. A behavior not
explicitly scripted may still be supported by general memory instructions.}
\label{tab:memcmp-boundary}
\centering\small
\begin{apptab}{@{}>{\raggedright\arraybackslash}p{5.6cm}>{\raggedright\arraybackslash}p{4.6cm}l@{}}
\thd{Behaviour} & \thd{Instruction or tool support} & \thd{Relation} \\
\midrule
Report the goal, outcome, and rooms passed & Handover instruction & Requested \\
Report which markers were dropped & Handover instruction & Requested \\
Write down unsuccessful searches & Handover instruction & Requested \\
Move a marker & Marker tool supports re-placement & Tool-supported \\
Question an earlier task account & Warning that accounts may be wrong & Requested \\
Correct a particular marker & General permission to maintain markers & Not scripted \\
Diagnose a particular instrument fault & General reasoning and recovery tools & Not scripted \\
Reject a stored route and navigate by sight & Access to memory and current views & Not scripted \\
\addlinespace
Check a note against current views & Instruction to treat notes as recollection & Requested \\
Identify a particular stale note & General warning about stale notes & Not scripted \\
Attempt to edit inherited notes & Writes outside permitted paths are refused & Unrequested attempt \\
\bottomrule
\end{apptab}
\end{table}

%% file: figures/fig-memcmp-cases-strip.tex
\begin{tikzpicture}[x=0.168cm,y=0.34cm]
\fill[figteal!28] (0,0.00) rectangle (1,1.00);
\fill[figteal!28] (1,0.00) rectangle (2,1.00);
\fill[figteal!28] (2,0.00) rectangle (3,1.00);
\fill[figteal!28] (3,0.00) rectangle (4,1.00);
\fill[figteal!28] (4,0.00) rectangle (5,1.00);
\fill[figteal!28] (5,0.00) rectangle (6,1.00);
\fill[figteal!28] (6,0.00) rectangle (7,1.00);
\fill[figteal!28] (7,0.00) rectangle (8,1.00);
\fill[figred!35] (8,0.00) rectangle (9,1.00);
\fill[figred!35] (9,0.00) rectangle (10,1.00);
\fill[figteal!28] (10,0.00) rectangle (11,1.00);
\fill[figteal!28] (11,0.00) rectangle (12,1.00);
\fill[figteal!28] (12,0.00) rectangle (13,1.00);
\fill[figteal!28] (13,0.00) rectangle (14,1.00);
\fill[figred!35] (14,0.00) rectangle (15,1.00);
\fill[figred!35] (15,0.00) rectangle (16,1.00);
\fill[figred!35] (16,0.00) rectangle (17,1.00);
\fill[figteal!28] (17,0.00) rectangle (18,1.00);
\fill[figteal!28] (18,0.00) rectangle (19,1.00);
\fill[figteal!28] (19,0.00) rectangle (20,1.00);
\fill[figteal!28] (20,0.00) rectangle (21,1.00);
\fill[figteal!28] (21,0.00) rectangle (22,1.00);
\fill[figteal!28] (22,0.00) rectangle (23,1.00);
\fill[figteal!28] (23,0.00) rectangle (24,1.00);
\fill[figteal!28] (24,0.00) rectangle (25,1.00);
\fill[figteal!28] (25,0.00) rectangle (26,1.00);
\fill[figteal!28] (26,0.00) rectangle (27,1.00);
\fill[figteal!28] (27,0.00) rectangle (28,1.00);
\fill[figred!35] (28,0.00) rectangle (29,1.00);
\fill[figteal!28] (29,0.00) rectangle (30,1.00);
\fill[figteal!28] (30,0.00) rectangle (31,1.00);
\fill[figred!35] (31,0.00) rectangle (32,1.00);
\fill[figteal!28] (32,0.00) rectangle (33,1.00);
\fill[figteal!28] (33,0.00) rectangle (34,1.00);
\fill[figteal!28] (34,0.00) rectangle (35,1.00);
\fill[figteal!28] (35,0.00) rectangle (36,1.00);
\fill[figteal!28] (36,0.00) rectangle (37,1.00);
\fill[figteal!28] (37,0.00) rectangle (38,1.00);
\fill[figteal!28] (38,0.00) rectangle (39,1.00);
\fill[figred!35] (39,0.00) rectangle (40,1.00);
\fill[figteal!28] (40,0.00) rectangle (41,1.00);
\fill[figteal!28] (41,0.00) rectangle (42,1.00);
\fill[figteal!28] (42,0.00) rectangle (43,1.00);
\fill[figteal!28] (43,0.00) rectangle (44,1.00);
\fill[figteal!28] (44,0.00) rectangle (45,1.00);
\fill[figteal!28] (45,0.00) rectangle (46,1.00);
\fill[figteal!28] (46,0.00) rectangle (47,1.00);
\fill[figteal!28] (47,0.00) rectangle (48,1.00);
\fill[figteal!28] (48,0.00) rectangle (49,1.00);
\fill[figteal!28] (49,0.00) rectangle (50,1.00);
\fill[figteal!28] (50,0.00) rectangle (51,1.00);
\fill[figteal!28] (51,0.00) rectangle (52,1.00);
\fill[figteal!28] (52,0.00) rectangle (53,1.00);
\fill[figteal!28] (53,0.00) rectangle (54,1.00);
\fill[figteal!28] (54,0.00) rectangle (55,1.00);
\fill[figred!35] (55,0.00) rectangle (56,1.00);
\fill[figteal!28] (56,0.00) rectangle (57,1.00);
\fill[figred!35] (57,0.00) rectangle (58,1.00);
\fill[figteal!28] (58,0.00) rectangle (59,1.00);
\fill[figteal!28] (59,0.00) rectangle (60,1.00);
\fill[figteal!28] (60,0.00) rectangle (61,1.00);
\fill[figteal!28] (61,0.00) rectangle (62,1.00);
\fill[figteal!28] (62,0.00) rectangle (63,1.00);
\fill[figteal!28] (63,0.00) rectangle (64,1.00);
\fill[figteal!28] (64,0.00) rectangle (65,1.00);
\fill[figred!35] (65,0.00) rectangle (66,1.00);
\fill[figred!35] (66,0.00) rectangle (67,1.00);
\fill[figteal!28] (67,0.00) rectangle (68,1.00);
\fill[figteal!28] (68,0.00) rectangle (69,1.00);
\fill[figteal!28] (69,0.00) rectangle (70,1.00);
\fill[figteal!28] (70,0.00) rectangle (71,1.00);
\fill[figteal!28] (71,0.00) rectangle (72,1.00);
\fill[figteal!28] (72,0.00) rectangle (73,1.00);
\fill[figteal!28] (73,0.00) rectangle (74,1.00);
\fill[figteal!28] (74,0.00) rectangle (75,1.00);
\fill[figred!35] (75,0.00) rectangle (76,1.00);
\fill[figteal!28] (76,0.00) rectangle (77,1.00);
\fill[figred!35] (77,0.00) rectangle (78,1.00);
\draw[gray!60,line width=0.3pt] (0,0.00) rectangle (78,1.00);
\fill[figteal!28] (0,1.15) rectangle (1,2.15);
\fill[figteal!28] (1,1.15) rectangle (2,2.15);
\fill[figteal!28] (2,1.15) rectangle (3,2.15);
\fill[figteal!28] (3,1.15) rectangle (4,2.15);
\fill[figteal!28] (4,1.15) rectangle (5,2.15);
\fill[figteal!28] (5,1.15) rectangle (6,2.15);
\fill[figteal!28] (6,1.15) rectangle (7,2.15);
\fill[figteal!28] (7,1.15) rectangle (8,2.15);
\fill[figteal!28] (8,1.15) rectangle (9,2.15);
\fill[figteal!28] (9,1.15) rectangle (10,2.15);
\fill[figteal!28] (10,1.15) rectangle (11,2.15);
\fill[figteal!28] (11,1.15) rectangle (12,2.15);
\fill[figteal!28] (12,1.15) rectangle (13,2.15);
\fill[figteal!28] (13,1.15) rectangle (14,2.15);
\fill[figteal!28] (14,1.15) rectangle (15,2.15);
\fill[figteal!28] (15,1.15) rectangle (16,2.15);
\fill[figteal!28] (16,1.15) rectangle (17,2.15);
\fill[figteal!28] (17,1.15) rectangle (18,2.15);
\fill[figteal!28] (18,1.15) rectangle (19,2.15);
\fill[figteal!28] (19,1.15) rectangle (20,2.15);
\fill[figteal!28] (20,1.15) rectangle (21,2.15);
\fill[figteal!28] (21,1.15) rectangle (22,2.15);
\fill[figteal!28] (22,1.15) rectangle (23,2.15);
\fill[figteal!28] (23,1.15) rectangle (24,2.15);
\fill[figteal!28] (24,1.15) rectangle (25,2.15);
\fill[figteal!28] (25,1.15) rectangle (26,2.15);
\fill[figred!35] (26,1.15) rectangle (27,2.15);
\fill[figteal!28] (27,1.15) rectangle (28,2.15);
\fill[figred!35] (28,1.15) rectangle (29,2.15);
\fill[figteal!28] (29,1.15) rectangle (30,2.15);
\fill[figteal!28] (30,1.15) rectangle (31,2.15);
\fill[figred!35] (31,1.15) rectangle (32,2.15);
\fill[figred!35] (32,1.15) rectangle (33,2.15);
\fill[figteal!28] (33,1.15) rectangle (34,2.15);
\fill[figteal!28] (34,1.15) rectangle (35,2.15);
\fill[figteal!28] (35,1.15) rectangle (36,2.15);
\fill[figteal!28] (36,1.15) rectangle (37,2.15);
\fill[figteal!28] (37,1.15) rectangle (38,2.15);
\fill[figteal!28] (38,1.15) rectangle (39,2.15);
\fill[figteal!28] (39,1.15) rectangle (40,2.15);
\fill[figteal!28] (40,1.15) rectangle (41,2.15);
\fill[figteal!28] (41,1.15) rectangle (42,2.15);
\fill[figteal!28] (42,1.15) rectangle (43,2.15);
\fill[figteal!28] (43,1.15) rectangle (44,2.15);
\fill[figteal!28] (44,1.15) rectangle (45,2.15);
\fill[figteal!28] (45,1.15) rectangle (46,2.15);
\fill[figteal!28] (46,1.15) rectangle (47,2.15);
\fill[figteal!28] (47,1.15) rectangle (48,2.15);
\fill[figteal!28] (48,1.15) rectangle (49,2.15);
\fill[figteal!28] (49,1.15) rectangle (50,2.15);
\fill[figteal!28] (50,1.15) rectangle (51,2.15);
\fill[figteal!28] (51,1.15) rectangle (52,2.15);
\fill[figteal!28] (52,1.15) rectangle (53,2.15);
\fill[figteal!28] (53,1.15) rectangle (54,2.15);
\fill[figteal!28] (54,1.15) rectangle (55,2.15);
\fill[figteal!28] (55,1.15) rectangle (56,2.15);
\fill[figteal!28] (56,1.15) rectangle (57,2.15);
\fill[figteal!28] (57,1.15) rectangle (58,2.15);
\fill[figteal!28] (58,1.15) rectangle (59,2.15);
\fill[figteal!28] (59,1.15) rectangle (60,2.15);
\fill[figteal!28] (60,1.15) rectangle (61,2.15);
\fill[figteal!28] (61,1.15) rectangle (62,2.15);
\fill[figteal!28] (62,1.15) rectangle (63,2.15);
\fill[figteal!28] (63,1.15) rectangle (64,2.15);
\fill[figteal!28] (64,1.15) rectangle (65,2.15);
\fill[figteal!28] (65,1.15) rectangle (66,2.15);
\fill[figteal!28] (66,1.15) rectangle (67,2.15);
\fill[figteal!28] (67,1.15) rectangle (68,2.15);
\fill[figteal!28] (68,1.15) rectangle (69,2.15);
\fill[figteal!28] (69,1.15) rectangle (70,2.15);
\fill[figteal!28] (70,1.15) rectangle (71,2.15);
\fill[figteal!28] (71,1.15) rectangle (72,2.15);
\fill[figteal!28] (72,1.15) rectangle (73,2.15);
\fill[figteal!28] (73,1.15) rectangle (74,2.15);
\fill[figteal!28] (74,1.15) rectangle (75,2.15);
\fill[figteal!28] (75,1.15) rectangle (76,2.15);
\fill[figteal!28] (76,1.15) rectangle (77,2.15);
\fill[figteal!28] (77,1.15) rectangle (78,2.15);
\draw[gray!60,line width=0.3pt] (0,1.15) rectangle (78,2.15);
\node[anchor=east,font=\fontsize{5}{5}\selectfont,text=gray] at (-0.6,0.50) {No consolidation};
\node[anchor=east,font=\fontsize{5}{5}\selectfont,text=gray] at (-0.6,1.65) {full};
\node[anchor=east,font=\footnotesize] at (-5.2,1.08) {House A};
\draw[gray!75,line width=0.3pt,dash pattern=on 1.2pt off 1pt] (16,-0.10) -- (16,2.30);
\draw[gray!75,line width=0.3pt,dash pattern=on 1.2pt off 1pt] (33,-0.10) -- (33,2.30);
\draw[gray!75,line width=0.3pt,dash pattern=on 1.2pt off 1pt] (49,-0.10) -- (49,2.30);
\draw[gray!75,line width=0.3pt,dash pattern=on 1.2pt off 1pt] (67,-0.10) -- (67,2.30);
\node[font=\fontsize{4.6}{4.6}\selectfont,text=gray] at (8.0,-1.25) {day 1};
\node[font=\fontsize{4.6}{4.6}\selectfont,text=gray] at (24.5,-1.25) {day 2};
\node[font=\fontsize{4.6}{4.6}\selectfont,text=gray] at (41.0,-1.25) {day 3};
\node[font=\fontsize{4.6}{4.6}\selectfont,text=gray] at (58.0,-1.25) {day 4};
\node[font=\fontsize{4.6}{4.6}\selectfont,text=gray] at (72.5,-1.25) {day 5};
\node[font=\fontsize{5}{5}\selectfont,text=gray] at (9.5,-0.55) {10};
\node[font=\fontsize{5}{5}\selectfont,text=gray] at (19.5,-0.55) {20};
\node[font=\fontsize{5}{5}\selectfont,text=gray] at (29.5,-0.55) {30};
\node[font=\fontsize{5}{5}\selectfont,text=gray] at (39.5,-0.55) {40};
\node[font=\fontsize{5}{5}\selectfont,text=gray] at (49.5,-0.55) {50};
\node[font=\fontsize{5}{5}\selectfont,text=gray] at (59.5,-0.55) {60};
\node[font=\fontsize{5}{5}\selectfont,text=gray] at (69.5,-0.55) {70};
\node[circle,fill=figred,text=white,inner sep=0.1pt,minimum size=0.28cm,font=\fontsize{5}{5}\selectfont] at (7.5,2.75) {11};
\node[circle,fill=figochre,text=white,inner sep=0.1pt,minimum size=0.28cm,font=\fontsize{5}{5}\selectfont] at (9.5,2.75) {5};
\node[circle,fill=figblue,text=white,inner sep=0.1pt,minimum size=0.28cm,font=\fontsize{5}{5}\selectfont] at (10.5,3.63) {1};
\node[circle,fill=figblue,text=white,inner sep=0.1pt,minimum size=0.28cm,font=\fontsize{5}{5}\selectfont] at (14.5,2.75) {2};
\node[circle,fill=figred,text=white,inner sep=0.1pt,minimum size=0.28cm,font=\fontsize{5}{5}\selectfont] at (14.5,3.63) {11};
\node[circle,fill=figblue,text=white,inner sep=0.1pt,minimum size=0.28cm,font=\fontsize{5}{5}\selectfont] at (16.5,2.75) {1};
\node[circle,fill=figteal,text=white,inner sep=0.1pt,minimum size=0.28cm,font=\fontsize{5}{5}\selectfont] at (27.5,2.75) {8};
\node[circle,fill=figteal,text=white,inner sep=0.1pt,minimum size=0.28cm,font=\fontsize{5}{5}\selectfont] at (28.5,3.63) {8};
\node[circle,fill=figochre,text=white,inner sep=0.1pt,minimum size=0.28cm,font=\fontsize{5}{5}\selectfont] at (31.5,2.75) {7};
\node[circle,fill=figochre,text=white,inner sep=0.1pt,minimum size=0.28cm,font=\fontsize{5}{5}\selectfont] at (32.5,3.63) {7};
\node[circle,fill=figblue,text=white,inner sep=0.1pt,minimum size=0.28cm,font=\fontsize{5}{5}\selectfont] at (33.5,2.75) {1};
\node[circle,fill=figblue,text=white,inner sep=0.1pt,minimum size=0.28cm,font=\fontsize{5}{5}\selectfont] at (33.5,4.51) {2};
\node[circle,fill=figteal,text=white,inner sep=0.1pt,minimum size=0.28cm,font=\fontsize{5}{5}\selectfont] at (33.5,5.40) {8};
\node[circle,fill=figteal,text=white,inner sep=0.1pt,minimum size=0.28cm,font=\fontsize{5}{5}\selectfont] at (34.5,3.63) {8};
\node[circle,fill=figblue,text=white,inner sep=0.1pt,minimum size=0.28cm,font=\fontsize{5}{5}\selectfont] at (39.5,2.75) {2};
\node[circle,fill=figred,text=white,inner sep=0.1pt,minimum size=0.28cm,font=\fontsize{5}{5}\selectfont] at (39.5,3.63) {11};
\node[circle,fill=figochre,text=white,inner sep=0.1pt,minimum size=0.28cm,font=\fontsize{5}{5}\selectfont] at (42.5,2.75) {5};
\node[circle,fill=figblue,text=white,inner sep=0.1pt,minimum size=0.28cm,font=\fontsize{5}{5}\selectfont] at (43.5,3.63) {2};
\node[circle,fill=figblue,text=white,inner sep=0.1pt,minimum size=0.28cm,font=\fontsize{5}{5}\selectfont] at (49.5,2.75) {1};
\node[circle,fill=figochre,text=white,inner sep=0.1pt,minimum size=0.28cm,font=\fontsize{5}{5}\selectfont] at (55.5,2.75) {5};
\node[circle,fill=figred,text=white,inner sep=0.1pt,minimum size=0.28cm,font=\fontsize{5}{5}\selectfont] at (57.5,2.75) {11};
\node[circle,fill=figred,text=white,inner sep=0.1pt,minimum size=0.28cm,font=\fontsize{5}{5}\selectfont] at (64.5,2.75) {11};
\node[circle,fill=figochre,text=white,inner sep=0.1pt,minimum size=0.28cm,font=\fontsize{5}{5}\selectfont] at (66.5,2.75) {5};
\node[circle,fill=figblue,text=white,inner sep=0.1pt,minimum size=0.28cm,font=\fontsize{5}{5}\selectfont] at (67.5,3.63) {1};
\node[circle,fill=figochre,text=white,inner sep=0.1pt,minimum size=0.28cm,font=\fontsize{5}{5}\selectfont] at (68.5,2.75) {5};
\node[circle,fill=figred,text=white,inner sep=0.1pt,minimum size=0.28cm,font=\fontsize{5}{5}\selectfont] at (68.5,4.51) {11};
\node[circle,fill=figteal,text=white,inner sep=0.1pt,minimum size=0.28cm,font=\fontsize{5}{5}\selectfont] at (69.5,3.63) {8};
\node[circle,fill=black!70,text=white,inner sep=0.1pt,minimum size=0.28cm,font=\fontsize{5}{5}\selectfont] at (69.5,5.40) {12};
\node[circle,fill=figochre,text=white,inner sep=0.1pt,minimum size=0.28cm,font=\fontsize{5}{5}\selectfont] at (77.5,2.75) {7};
\node[circle,fill=figteal,text=white,inner sep=0.1pt,minimum size=0.28cm,font=\fontsize{5}{5}\selectfont] at (77.5,3.63) {8};
\draw[gray!55,line width=0.25pt] (7.5,2.20) -- (7.5,2.34);
\draw[gray!55,line width=0.25pt] (9.5,2.20) -- (9.5,2.34);
\draw[gray!55,line width=0.25pt] (10.5,2.20) -- (10.5,3.22);
\draw[gray!55,line width=0.25pt] (14.5,2.20) -- (14.5,2.34);
\draw[gray!55,line width=0.25pt] (16.5,2.20) -- (16.5,2.34);
\draw[gray!55,line width=0.25pt] (27.5,2.20) -- (27.5,2.34);
\draw[gray!55,line width=0.25pt] (28.5,2.20) -- (28.5,3.22);
\draw[gray!55,line width=0.25pt] (31.5,2.20) -- (31.5,2.34);
\draw[gray!55,line width=0.25pt] (32.5,2.20) -- (32.5,3.22);
\draw[gray!55,line width=0.25pt] (33.5,2.20) -- (33.5,2.34);
\draw[gray!55,line width=0.25pt] (34.5,2.20) -- (34.5,3.22);
\draw[gray!55,line width=0.25pt] (39.5,2.20) -- (39.5,2.34);
\draw[gray!55,line width=0.25pt] (42.5,2.20) -- (42.5,2.34);
\draw[gray!55,line width=0.25pt] (43.5,2.20) -- (43.5,3.22);
\draw[gray!55,line width=0.25pt] (49.5,2.20) -- (49.5,2.34);
\draw[gray!55,line width=0.25pt] (55.5,2.20) -- (55.5,2.34);
\draw[gray!55,line width=0.25pt] (57.5,2.20) -- (57.5,2.34);
\draw[gray!55,line width=0.25pt] (64.5,2.20) -- (64.5,2.34);
\draw[gray!55,line width=0.25pt] (66.5,2.20) -- (66.5,2.34);
\draw[gray!55,line width=0.25pt] (67.5,2.20) -- (67.5,3.22);
\draw[gray!55,line width=0.25pt] (68.5,2.20) -- (68.5,2.34);
\draw[gray!55,line width=0.25pt] (69.5,2.20) -- (69.5,3.22);
\draw[gray!55,line width=0.25pt] (77.5,2.20) -- (77.5,2.34);
\fill[figteal!28] (0,-8.06) rectangle (1,-7.06);
\fill[figteal!28] (1,-8.06) rectangle (2,-7.06);
\fill[figteal!28] (2,-8.06) rectangle (3,-7.06);
\fill[figteal!28] (3,-8.06) rectangle (4,-7.06);
\fill[figteal!28] (4,-8.06) rectangle (5,-7.06);
\fill[figteal!28] (5,-8.06) rectangle (6,-7.06);
\fill[figteal!28] (6,-8.06) rectangle (7,-7.06);
\fill[figteal!28] (7,-8.06) rectangle (8,-7.06);
\fill[figteal!28] (8,-8.06) rectangle (9,-7.06);
\fill[figteal!28] (9,-8.06) rectangle (10,-7.06);
\fill[figteal!28] (10,-8.06) rectangle (11,-7.06);
\fill[figteal!28] (11,-8.06) rectangle (12,-7.06);
\fill[figteal!28] (12,-8.06) rectangle (13,-7.06);
\fill[figteal!28] (13,-8.06) rectangle (14,-7.06);
\fill[figteal!28] (14,-8.06) rectangle (15,-7.06);
\fill[figteal!28] (15,-8.06) rectangle (16,-7.06);
\fill[figteal!28] (16,-8.06) rectangle (17,-7.06);
\fill[figteal!28] (17,-8.06) rectangle (18,-7.06);
\fill[figteal!28] (18,-8.06) rectangle (19,-7.06);
\fill[figteal!28] (19,-8.06) rectangle (20,-7.06);
\fill[figteal!28] (20,-8.06) rectangle (21,-7.06);
\fill[figteal!28] (21,-8.06) rectangle (22,-7.06);
\fill[figteal!28] (22,-8.06) rectangle (23,-7.06);
\fill[figteal!28] (23,-8.06) rectangle (24,-7.06);
\fill[figteal!28] (24,-8.06) rectangle (25,-7.06);
\fill[figteal!28] (25,-8.06) rectangle (26,-7.06);
\fill[figteal!28] (26,-8.06) rectangle (27,-7.06);
\fill[figteal!28] (27,-8.06) rectangle (28,-7.06);
\fill[figteal!28] (28,-8.06) rectangle (29,-7.06);
\fill[figteal!28] (29,-8.06) rectangle (30,-7.06);
\fill[figteal!28] (30,-8.06) rectangle (31,-7.06);
\fill[figteal!28] (31,-8.06) rectangle (32,-7.06);
\fill[figteal!28] (32,-8.06) rectangle (33,-7.06);
\fill[figteal!28] (33,-8.06) rectangle (34,-7.06);
\fill[figteal!28] (34,-8.06) rectangle (35,-7.06);
\fill[figred!35] (35,-8.06) rectangle (36,-7.06);
\fill[figteal!28] (36,-8.06) rectangle (37,-7.06);
\fill[figteal!28] (37,-8.06) rectangle (38,-7.06);
\fill[figteal!28] (38,-8.06) rectangle (39,-7.06);
\fill[figteal!28] (39,-8.06) rectangle (40,-7.06);
\fill[figteal!28] (40,-8.06) rectangle (41,-7.06);
\fill[figteal!28] (41,-8.06) rectangle (42,-7.06);
\fill[figteal!28] (42,-8.06) rectangle (43,-7.06);
\fill[figteal!28] (43,-8.06) rectangle (44,-7.06);
\fill[figteal!28] (44,-8.06) rectangle (45,-7.06);
\fill[figteal!28] (45,-8.06) rectangle (46,-7.06);
\fill[figteal!28] (46,-8.06) rectangle (47,-7.06);
\fill[figteal!28] (47,-8.06) rectangle (48,-7.06);
\fill[figteal!28] (48,-8.06) rectangle (49,-7.06);
\fill[figteal!28] (49,-8.06) rectangle (50,-7.06);
\fill[figteal!28] (50,-8.06) rectangle (51,-7.06);
\fill[figteal!28] (51,-8.06) rectangle (52,-7.06);
\fill[figteal!28] (52,-8.06) rectangle (53,-7.06);
\fill[figteal!28] (53,-8.06) rectangle (54,-7.06);
\fill[figteal!28] (54,-8.06) rectangle (55,-7.06);
\fill[figteal!28] (55,-8.06) rectangle (56,-7.06);
\fill[figteal!28] (56,-8.06) rectangle (57,-7.06);
\fill[figteal!28] (57,-8.06) rectangle (58,-7.06);
\fill[figteal!28] (58,-8.06) rectangle (59,-7.06);
\fill[figred!35] (59,-8.06) rectangle (60,-7.06);
\fill[figteal!28] (60,-8.06) rectangle (61,-7.06);
\fill[figred!35] (61,-8.06) rectangle (62,-7.06);
\fill[figteal!28] (62,-8.06) rectangle (63,-7.06);
\fill[figteal!28] (63,-8.06) rectangle (64,-7.06);
\fill[figteal!28] (64,-8.06) rectangle (65,-7.06);
\fill[figteal!28] (65,-8.06) rectangle (66,-7.06);
\draw[gray!60,line width=0.3pt] (0,-8.06) rectangle (66,-7.06);
\fill[figteal!28] (0,-6.91) rectangle (1,-5.91);
\fill[figteal!28] (1,-6.91) rectangle (2,-5.91);
\fill[figteal!28] (2,-6.91) rectangle (3,-5.91);
\fill[figteal!28] (3,-6.91) rectangle (4,-5.91);
\fill[figteal!28] (4,-6.91) rectangle (5,-5.91);
\fill[figteal!28] (5,-6.91) rectangle (6,-5.91);
\fill[figteal!28] (6,-6.91) rectangle (7,-5.91);
\fill[figteal!28] (7,-6.91) rectangle (8,-5.91);
\fill[figteal!28] (8,-6.91) rectangle (9,-5.91);
\fill[figteal!28] (9,-6.91) rectangle (10,-5.91);
\fill[figteal!28] (10,-6.91) rectangle (11,-5.91);
\fill[figteal!28] (11,-6.91) rectangle (12,-5.91);
\fill[figteal!28] (12,-6.91) rectangle (13,-5.91);
\fill[figteal!28] (13,-6.91) rectangle (14,-5.91);
\fill[figteal!28] (14,-6.91) rectangle (15,-5.91);
\fill[figteal!28] (15,-6.91) rectangle (16,-5.91);
\fill[figteal!28] (16,-6.91) rectangle (17,-5.91);
\fill[figteal!28] (17,-6.91) rectangle (18,-5.91);
\fill[figteal!28] (18,-6.91) rectangle (19,-5.91);
\fill[figteal!28] (19,-6.91) rectangle (20,-5.91);
\fill[figteal!28] (20,-6.91) rectangle (21,-5.91);
\fill[figteal!28] (21,-6.91) rectangle (22,-5.91);
\fill[figteal!28] (22,-6.91) rectangle (23,-5.91);
\fill[figteal!28] (23,-6.91) rectangle (24,-5.91);
\fill[figteal!28] (24,-6.91) rectangle (25,-5.91);
\fill[figred!35] (25,-6.91) rectangle (26,-5.91);
\fill[figteal!28] (26,-6.91) rectangle (27,-5.91);
\fill[figteal!28] (27,-6.91) rectangle (28,-5.91);
\fill[figteal!28] (28,-6.91) rectangle (29,-5.91);
\fill[figteal!28] (29,-6.91) rectangle (30,-5.91);
\fill[figteal!28] (30,-6.91) rectangle (31,-5.91);
\fill[figteal!28] (31,-6.91) rectangle (32,-5.91);
\fill[figteal!28] (32,-6.91) rectangle (33,-5.91);
\fill[figteal!28] (33,-6.91) rectangle (34,-5.91);
\fill[figteal!28] (34,-6.91) rectangle (35,-5.91);
\fill[figteal!28] (35,-6.91) rectangle (36,-5.91);
\fill[figteal!28] (36,-6.91) rectangle (37,-5.91);
\fill[figteal!28] (37,-6.91) rectangle (38,-5.91);
\fill[figteal!28] (38,-6.91) rectangle (39,-5.91);
\fill[figteal!28] (39,-6.91) rectangle (40,-5.91);
\fill[figteal!28] (40,-6.91) rectangle (41,-5.91);
\fill[figteal!28] (41,-6.91) rectangle (42,-5.91);
\fill[figteal!28] (42,-6.91) rectangle (43,-5.91);
\fill[figteal!28] (43,-6.91) rectangle (44,-5.91);
\fill[figteal!28] (44,-6.91) rectangle (45,-5.91);
\fill[figteal!28] (45,-6.91) rectangle (46,-5.91);
\fill[figteal!28] (46,-6.91) rectangle (47,-5.91);
\fill[figteal!28] (47,-6.91) rectangle (48,-5.91);
\fill[figteal!28] (48,-6.91) rectangle (49,-5.91);
\fill[figteal!28] (49,-6.91) rectangle (50,-5.91);
\fill[figteal!28] (50,-6.91) rectangle (51,-5.91);
\fill[figteal!28] (51,-6.91) rectangle (52,-5.91);
\fill[figteal!28] (52,-6.91) rectangle (53,-5.91);
\fill[figteal!28] (53,-6.91) rectangle (54,-5.91);
\fill[figteal!28] (54,-6.91) rectangle (55,-5.91);
\fill[figteal!28] (55,-6.91) rectangle (56,-5.91);
\fill[figteal!28] (56,-6.91) rectangle (57,-5.91);
\fill[figteal!28] (57,-6.91) rectangle (58,-5.91);
\fill[figteal!28] (58,-6.91) rectangle (59,-5.91);
\fill[figred!35] (59,-6.91) rectangle (60,-5.91);
\fill[figteal!28] (60,-6.91) rectangle (61,-5.91);
\fill[figteal!28] (61,-6.91) rectangle (62,-5.91);
\fill[figteal!28] (62,-6.91) rectangle (63,-5.91);
\fill[figteal!28] (63,-6.91) rectangle (64,-5.91);
\fill[figteal!28] (64,-6.91) rectangle (65,-5.91);
\fill[figteal!28] (65,-6.91) rectangle (66,-5.91);
\draw[gray!60,line width=0.3pt] (0,-6.91) rectangle (66,-5.91);
\node[anchor=east,font=\fontsize{5}{5}\selectfont,text=gray] at (-0.6,-7.56) {No consolidation};
\node[anchor=east,font=\fontsize{5}{5}\selectfont,text=gray] at (-0.6,-6.41) {full};
\node[anchor=east,font=\footnotesize] at (-5.2,-6.98) {House B};
\draw[gray!75,line width=0.3pt,dash pattern=on 1.2pt off 1pt] (19,-8.16) -- (19,-5.76);
\draw[gray!75,line width=0.3pt,dash pattern=on 1.2pt off 1pt] (35,-8.16) -- (35,-5.76);
\draw[gray!75,line width=0.3pt,dash pattern=on 1.2pt off 1pt] (51,-8.16) -- (51,-5.76);
\node[font=\fontsize{4.6}{4.6}\selectfont,text=gray] at (9.5,-9.31) {day 1};
\node[font=\fontsize{4.6}{4.6}\selectfont,text=gray] at (27.0,-9.31) {day 2};
\node[font=\fontsize{4.6}{4.6}\selectfont,text=gray] at (43.0,-9.31) {day 3};
\node[font=\fontsize{4.6}{4.6}\selectfont,text=gray] at (58.5,-9.31) {day 4};
\node[font=\fontsize{5}{5}\selectfont,text=gray] at (9.5,-8.61) {10};
\node[font=\fontsize{5}{5}\selectfont,text=gray] at (19.5,-8.61) {20};
\node[font=\fontsize{5}{5}\selectfont,text=gray] at (29.5,-8.61) {30};
\node[font=\fontsize{5}{5}\selectfont,text=gray] at (39.5,-8.61) {40};
\node[font=\fontsize{5}{5}\selectfont,text=gray] at (49.5,-8.61) {50};
\node[font=\fontsize{5}{5}\selectfont,text=gray] at (59.5,-8.61) {60};
\node[circle,fill=figochre,text=white,inner sep=0.1pt,minimum size=0.28cm,font=\fontsize{5}{5}\selectfont] at (1.5,-5.31) {6};
\node[circle,fill=figochre,text=white,inner sep=0.1pt,minimum size=0.28cm,font=\fontsize{5}{5}\selectfont] at (7.5,-5.31) {6};
\node[circle,fill=figred,text=white,inner sep=0.1pt,minimum size=0.28cm,font=\fontsize{5}{5}\selectfont] at (12.5,-5.31) {10};
\node[circle,fill=figred,text=white,inner sep=0.1pt,minimum size=0.28cm,font=\fontsize{5}{5}\selectfont] at (18.5,-5.31) {10};
\node[circle,fill=figteal,text=white,inner sep=0.1pt,minimum size=0.28cm,font=\fontsize{5}{5}\selectfont] at (19.5,-4.43) {9};
\node[circle,fill=figred,text=white,inner sep=0.1pt,minimum size=0.28cm,font=\fontsize{5}{5}\selectfont] at (20.5,-5.31) {10};
\node[circle,fill=figteal,text=white,inner sep=0.1pt,minimum size=0.28cm,font=\fontsize{5}{5}\selectfont] at (25.5,-5.31) {9};
\node[circle,fill=figred,text=white,inner sep=0.1pt,minimum size=0.28cm,font=\fontsize{5}{5}\selectfont] at (35.5,-5.31) {10};
\node[circle,fill=figteal,text=white,inner sep=0.1pt,minimum size=0.28cm,font=\fontsize{5}{5}\selectfont] at (51.5,-5.31) {9};
\node[circle,fill=figteal,text=white,inner sep=0.1pt,minimum size=0.28cm,font=\fontsize{5}{5}\selectfont] at (53.5,-5.31) {9};
\node[circle,fill=black!70,text=white,inner sep=0.1pt,minimum size=0.28cm,font=\fontsize{5}{5}\selectfont] at (59.5,-5.31) {14};
\node[circle,fill=figochre,text=white,inner sep=0.1pt,minimum size=0.28cm,font=\fontsize{5}{5}\selectfont] at (60.5,-4.43) {6};
\node[circle,fill=figochre,text=white,inner sep=0.1pt,minimum size=0.28cm,font=\fontsize{5}{5}\selectfont] at (61.5,-5.31) {6};
\node[circle,fill=black!70,text=white,inner sep=0.1pt,minimum size=0.28cm,font=\fontsize{5}{5}\selectfont] at (64.5,-5.31) {14};
\draw[gray!55,line width=0.25pt] (1.5,-5.86) -- (1.5,-5.72);
\draw[gray!55,line width=0.25pt] (7.5,-5.86) -- (7.5,-5.72);
\draw[gray!55,line width=0.25pt] (12.5,-5.86) -- (12.5,-5.72);
\draw[gray!55,line width=0.25pt] (18.5,-5.86) -- (18.5,-5.72);
\draw[gray!55,line width=0.25pt] (19.5,-5.86) -- (19.5,-4.84);
\draw[gray!55,line width=0.25pt] (20.5,-5.86) -- (20.5,-5.72);
\draw[gray!55,line width=0.25pt] (25.5,-5.86) -- (25.5,-5.72);
\draw[gray!55,line width=0.25pt] (35.5,-5.86) -- (35.5,-5.72);
\draw[gray!55,line width=0.25pt] (51.5,-5.86) -- (51.5,-5.72);
\draw[gray!55,line width=0.25pt] (53.5,-5.86) -- (53.5,-5.72);
\draw[gray!55,line width=0.25pt] (59.5,-5.86) -- (59.5,-5.72);
\draw[gray!55,line width=0.25pt] (60.5,-5.86) -- (60.5,-4.84);
\draw[gray!55,line width=0.25pt] (61.5,-5.86) -- (61.5,-5.72);
\draw[gray!55,line width=0.25pt] (64.5,-5.86) -- (64.5,-5.72);
\fill[figred!35] (0,-14.35) rectangle (1,-13.35);
\fill[figteal!28] (1,-14.35) rectangle (2,-13.35);
\fill[figteal!28] (2,-14.35) rectangle (3,-13.35);
\fill[figteal!28] (3,-14.35) rectangle (4,-13.35);
\fill[figteal!28] (4,-14.35) rectangle (5,-13.35);
\fill[figred!35] (5,-14.35) rectangle (6,-13.35);
\fill[figteal!28] (6,-14.35) rectangle (7,-13.35);
\fill[figteal!28] (7,-14.35) rectangle (8,-13.35);
\fill[figred!35] (8,-14.35) rectangle (9,-13.35);
\fill[figred!35] (9,-14.35) rectangle (10,-13.35);
\fill[figteal!28] (10,-14.35) rectangle (11,-13.35);
\fill[figteal!28] (11,-14.35) rectangle (12,-13.35);
\fill[figred!35] (12,-14.35) rectangle (13,-13.35);
\fill[figred!35] (13,-14.35) rectangle (14,-13.35);
\fill[figred!35] (14,-14.35) rectangle (15,-13.35);
\fill[figteal!28] (15,-14.35) rectangle (16,-13.35);
\fill[figteal!28] (16,-14.35) rectangle (17,-13.35);
\fill[figred!35] (17,-14.35) rectangle (18,-13.35);
\fill[figteal!28] (18,-14.35) rectangle (19,-13.35);
\fill[figteal!28] (19,-14.35) rectangle (20,-13.35);
\fill[figteal!28] (20,-14.35) rectangle (21,-13.35);
\fill[figred!35] (21,-14.35) rectangle (22,-13.35);
\fill[figteal!28] (22,-14.35) rectangle (23,-13.35);
\fill[figteal!28] (23,-14.35) rectangle (24,-13.35);
\fill[figred!35] (24,-14.35) rectangle (25,-13.35);
\fill[figred!35] (25,-14.35) rectangle (26,-13.35);
\fill[figteal!28] (26,-14.35) rectangle (27,-13.35);
\fill[figteal!28] (27,-14.35) rectangle (28,-13.35);
\fill[figteal!28] (28,-14.35) rectangle (29,-13.35);
\fill[figteal!28] (29,-14.35) rectangle (30,-13.35);
\fill[figred!35] (30,-14.35) rectangle (31,-13.35);
\fill[figteal!28] (31,-14.35) rectangle (32,-13.35);
\fill[figteal!28] (32,-14.35) rectangle (33,-13.35);
\fill[figteal!28] (33,-14.35) rectangle (34,-13.35);
\fill[figred!35] (34,-14.35) rectangle (35,-13.35);
\fill[figteal!28] (35,-14.35) rectangle (36,-13.35);
\fill[figteal!28] (36,-14.35) rectangle (37,-13.35);
\fill[figteal!28] (37,-14.35) rectangle (38,-13.35);
\fill[figteal!28] (38,-14.35) rectangle (39,-13.35);
\fill[figteal!28] (39,-14.35) rectangle (40,-13.35);
\fill[figteal!28] (40,-14.35) rectangle (41,-13.35);
\fill[figteal!28] (41,-14.35) rectangle (42,-13.35);
\fill[figteal!28] (42,-14.35) rectangle (43,-13.35);
\fill[figteal!28] (43,-14.35) rectangle (44,-13.35);
\fill[figred!35] (44,-14.35) rectangle (45,-13.35);
\fill[figteal!28] (45,-14.35) rectangle (46,-13.35);
\fill[figteal!28] (46,-14.35) rectangle (47,-13.35);
\fill[figteal!28] (47,-14.35) rectangle (48,-13.35);
\fill[figteal!28] (48,-14.35) rectangle (49,-13.35);
\fill[figred!35] (49,-14.35) rectangle (50,-13.35);
\fill[figteal!28] (50,-14.35) rectangle (51,-13.35);
\fill[figteal!28] (51,-14.35) rectangle (52,-13.35);
\fill[figteal!28] (52,-14.35) rectangle (53,-13.35);
\fill[figteal!28] (53,-14.35) rectangle (54,-13.35);
\fill[figteal!28] (54,-14.35) rectangle (55,-13.35);
\fill[figteal!28] (55,-14.35) rectangle (56,-13.35);
\fill[figteal!28] (56,-14.35) rectangle (57,-13.35);
\fill[figred!35] (57,-14.35) rectangle (58,-13.35);
\fill[figred!35] (58,-14.35) rectangle (59,-13.35);
\fill[figred!35] (59,-14.35) rectangle (60,-13.35);
\fill[figteal!28] (60,-14.35) rectangle (61,-13.35);
\fill[figteal!28] (61,-14.35) rectangle (62,-13.35);
\fill[figred!35] (62,-14.35) rectangle (63,-13.35);
\fill[figteal!28] (63,-14.35) rectangle (64,-13.35);
\fill[figred!35] (64,-14.35) rectangle (65,-13.35);
\fill[figteal!28] (65,-14.35) rectangle (66,-13.35);
\fill[figred!35] (66,-14.35) rectangle (67,-13.35);
\fill[figteal!28] (67,-14.35) rectangle (68,-13.35);
\fill[figteal!28] (68,-14.35) rectangle (69,-13.35);
\fill[figteal!28] (69,-14.35) rectangle (70,-13.35);
\fill[figred!35] (70,-14.35) rectangle (71,-13.35);
\fill[figteal!28] (71,-14.35) rectangle (72,-13.35);
\fill[figteal!28] (72,-14.35) rectangle (73,-13.35);
\fill[figteal!28] (73,-14.35) rectangle (74,-13.35);
\fill[figred!35] (74,-14.35) rectangle (75,-13.35);
\fill[figteal!28] (75,-14.35) rectangle (76,-13.35);
\fill[figred!35] (76,-14.35) rectangle (77,-13.35);
\fill[figteal!28] (77,-14.35) rectangle (78,-13.35);
\fill[figteal!28] (78,-14.35) rectangle (79,-13.35);
\draw[gray!60,line width=0.3pt] (0,-14.35) rectangle (79,-13.35);
\fill[figteal!28] (0,-13.20) rectangle (1,-12.20);
\fill[figteal!28] (1,-13.20) rectangle (2,-12.20);
\fill[figteal!28] (2,-13.20) rectangle (3,-12.20);
\fill[figteal!28] (3,-13.20) rectangle (4,-12.20);
\fill[figteal!28] (4,-13.20) rectangle (5,-12.20);
\fill[figteal!28] (5,-13.20) rectangle (6,-12.20);
\fill[figred!35] (6,-13.20) rectangle (7,-12.20);
\fill[figred!35] (7,-13.20) rectangle (8,-12.20);
\fill[figteal!28] (8,-13.20) rectangle (9,-12.20);
\fill[figred!35] (9,-13.20) rectangle (10,-12.20);
\fill[figteal!28] (10,-13.20) rectangle (11,-12.20);
\fill[figteal!28] (11,-13.20) rectangle (12,-12.20);
\fill[figred!35] (12,-13.20) rectangle (13,-12.20);
\fill[figteal!28] (13,-13.20) rectangle (14,-12.20);
\fill[figred!35] (14,-13.20) rectangle (15,-12.20);
\fill[figteal!28] (15,-13.20) rectangle (16,-12.20);
\fill[figteal!28] (16,-13.20) rectangle (17,-12.20);
\fill[figred!35] (17,-13.20) rectangle (18,-12.20);
\fill[figteal!28] (18,-13.20) rectangle (19,-12.20);
\fill[figteal!28] (19,-13.20) rectangle (20,-12.20);
\fill[figteal!28] (20,-13.20) rectangle (21,-12.20);
\fill[figred!35] (21,-13.20) rectangle (22,-12.20);
\fill[figteal!28] (22,-13.20) rectangle (23,-12.20);
\fill[figteal!28] (23,-13.20) rectangle (24,-12.20);
\fill[figteal!28] (24,-13.20) rectangle (25,-12.20);
\fill[figteal!28] (25,-13.20) rectangle (26,-12.20);
\fill[figteal!28] (26,-13.20) rectangle (27,-12.20);
\fill[figteal!28] (27,-13.20) rectangle (28,-12.20);
\fill[figteal!28] (28,-13.20) rectangle (29,-12.20);
\fill[figteal!28] (29,-13.20) rectangle (30,-12.20);
\fill[figred!35] (30,-13.20) rectangle (31,-12.20);
\fill[figteal!28] (31,-13.20) rectangle (32,-12.20);
\fill[figteal!28] (32,-13.20) rectangle (33,-12.20);
\fill[figteal!28] (33,-13.20) rectangle (34,-12.20);
\fill[figteal!28] (34,-13.20) rectangle (35,-12.20);
\fill[figteal!28] (35,-13.20) rectangle (36,-12.20);
\fill[figteal!28] (36,-13.20) rectangle (37,-12.20);
\fill[figteal!28] (37,-13.20) rectangle (38,-12.20);
\fill[figteal!28] (38,-13.20) rectangle (39,-12.20);
\fill[figteal!28] (39,-13.20) rectangle (40,-12.20);
\fill[figteal!28] (40,-13.20) rectangle (41,-12.20);
\fill[figteal!28] (41,-13.20) rectangle (42,-12.20);
\fill[figteal!28] (42,-13.20) rectangle (43,-12.20);
\fill[figteal!28] (43,-13.20) rectangle (44,-12.20);
\fill[figred!35] (44,-13.20) rectangle (45,-12.20);
\fill[figteal!28] (45,-13.20) rectangle (46,-12.20);
\fill[figteal!28] (46,-13.20) rectangle (47,-12.20);
\fill[figteal!28] (47,-13.20) rectangle (48,-12.20);
\fill[figteal!28] (48,-13.20) rectangle (49,-12.20);
\fill[figred!35] (49,-13.20) rectangle (50,-12.20);
\fill[figteal!28] (50,-13.20) rectangle (51,-12.20);
\fill[figteal!28] (51,-13.20) rectangle (52,-12.20);
\fill[figteal!28] (52,-13.20) rectangle (53,-12.20);
\fill[figteal!28] (53,-13.20) rectangle (54,-12.20);
\fill[figteal!28] (54,-13.20) rectangle (55,-12.20);
\fill[figteal!28] (55,-13.20) rectangle (56,-12.20);
\fill[figteal!28] (56,-13.20) rectangle (57,-12.20);
\fill[figteal!28] (57,-13.20) rectangle (58,-12.20);
\fill[figred!35] (58,-13.20) rectangle (59,-12.20);
\fill[figred!35] (59,-13.20) rectangle (60,-12.20);
\fill[figteal!28] (60,-13.20) rectangle (61,-12.20);
\fill[figteal!28] (61,-13.20) rectangle (62,-12.20);
\fill[figred!35] (62,-13.20) rectangle (63,-12.20);
\fill[figteal!28] (63,-13.20) rectangle (64,-12.20);
\fill[figred!35] (64,-13.20) rectangle (65,-12.20);
\fill[figteal!28] (65,-13.20) rectangle (66,-12.20);
\fill[figteal!28] (66,-13.20) rectangle (67,-12.20);
\fill[figteal!28] (67,-13.20) rectangle (68,-12.20);
\fill[figteal!28] (68,-13.20) rectangle (69,-12.20);
\fill[figteal!28] (69,-13.20) rectangle (70,-12.20);
\fill[figred!35] (70,-13.20) rectangle (71,-12.20);
\fill[figred!35] (71,-13.20) rectangle (72,-12.20);
\fill[figteal!28] (72,-13.20) rectangle (73,-12.20);
\fill[figteal!28] (73,-13.20) rectangle (74,-12.20);
\fill[figteal!28] (74,-13.20) rectangle (75,-12.20);
\fill[figteal!28] (75,-13.20) rectangle (76,-12.20);
\fill[figred!35] (76,-13.20) rectangle (77,-12.20);
\fill[figteal!28] (77,-13.20) rectangle (78,-12.20);
\fill[figteal!28] (78,-13.20) rectangle (79,-12.20);
\draw[gray!60,line width=0.3pt] (0,-13.20) rectangle (79,-12.20);
\node[anchor=east,font=\fontsize{5}{5}\selectfont,text=gray] at (-0.6,-13.85) {No consolidation};
\node[anchor=east,font=\fontsize{5}{5}\selectfont,text=gray] at (-0.6,-12.70) {full};
\node[anchor=east,font=\footnotesize] at (-5.2,-13.27) {House C};
\draw[gray!75,line width=0.3pt,dash pattern=on 1.2pt off 1pt] (14,-14.45) -- (14,-12.05);
\draw[gray!75,line width=0.3pt,dash pattern=on 1.2pt off 1pt] (33,-14.45) -- (33,-12.05);
\draw[gray!75,line width=0.3pt,dash pattern=on 1.2pt off 1pt] (47,-14.45) -- (47,-12.05);
\draw[gray!75,line width=0.3pt,dash pattern=on 1.2pt off 1pt] (66,-14.45) -- (66,-12.05);
\node[font=\fontsize{4.6}{4.6}\selectfont,text=gray] at (7.0,-15.60) {day 1};
\node[font=\fontsize{4.6}{4.6}\selectfont,text=gray] at (23.5,-15.60) {day 2};
\node[font=\fontsize{4.6}{4.6}\selectfont,text=gray] at (40.0,-15.60) {day 3};
\node[font=\fontsize{4.6}{4.6}\selectfont,text=gray] at (56.5,-15.60) {day 4};
\node[font=\fontsize{4.6}{4.6}\selectfont,text=gray] at (72.5,-15.60) {day 5};
\node[font=\fontsize{5}{5}\selectfont,text=gray] at (9.5,-14.90) {10};
\node[font=\fontsize{5}{5}\selectfont,text=gray] at (19.5,-14.90) {20};
\node[font=\fontsize{5}{5}\selectfont,text=gray] at (29.5,-14.90) {30};
\node[font=\fontsize{5}{5}\selectfont,text=gray] at (39.5,-14.90) {40};
\node[font=\fontsize{5}{5}\selectfont,text=gray] at (49.5,-14.90) {50};
\node[font=\fontsize{5}{5}\selectfont,text=gray] at (59.5,-14.90) {60};
\node[font=\fontsize{5}{5}\selectfont,text=gray] at (69.5,-14.90) {70};
\node[circle,fill=figblue,text=white,inner sep=0.1pt,minimum size=0.28cm,font=\fontsize{5}{5}\selectfont] at (16.5,-11.60) {3};
\node[circle,fill=figblue,text=white,inner sep=0.1pt,minimum size=0.28cm,font=\fontsize{5}{5}\selectfont] at (28.5,-11.60) {4};
\node[circle,fill=figblue,text=white,inner sep=0.1pt,minimum size=0.28cm,font=\fontsize{5}{5}\selectfont] at (33.5,-11.60) {4};
\node[circle,fill=figblue,text=white,inner sep=0.1pt,minimum size=0.28cm,font=\fontsize{5}{5}\selectfont] at (34.5,-10.72) {3};
\node[circle,fill=figblue,text=white,inner sep=0.1pt,minimum size=0.28cm,font=\fontsize{5}{5}\selectfont] at (35.5,-11.60) {4};
\node[circle,fill=figblue,text=white,inner sep=0.1pt,minimum size=0.28cm,font=\fontsize{5}{5}\selectfont] at (39.5,-11.60) {3};
\node[circle,fill=figblue,text=white,inner sep=0.1pt,minimum size=0.28cm,font=\fontsize{5}{5}\selectfont] at (41.5,-11.60) {3};
\node[circle,fill=figblue,text=white,inner sep=0.1pt,minimum size=0.28cm,font=\fontsize{5}{5}\selectfont] at (53.5,-11.60) {3};
\node[circle,fill=black!70,text=white,inner sep=0.1pt,minimum size=0.28cm,font=\fontsize{5}{5}\selectfont] at (55.5,-11.60) {13};
\node[circle,fill=black!70,text=white,inner sep=0.1pt,minimum size=0.28cm,font=\fontsize{5}{5}\selectfont] at (66.5,-11.60) {13};
\node[circle,fill=figblue,text=white,inner sep=0.1pt,minimum size=0.28cm,font=\fontsize{5}{5}\selectfont] at (71.5,-11.60) {4};
\node[circle,fill=black!70,text=white,inner sep=0.1pt,minimum size=0.28cm,font=\fontsize{5}{5}\selectfont] at (72.5,-10.72) {13};
\node[circle,fill=figblue,text=white,inner sep=0.1pt,minimum size=0.28cm,font=\fontsize{5}{5}\selectfont] at (73.5,-11.60) {4};
\node[circle,fill=figblue,text=white,inner sep=0.1pt,minimum size=0.28cm,font=\fontsize{5}{5}\selectfont] at (77.5,-11.60) {4};
\draw[gray!55,line width=0.25pt] (16.5,-12.15) -- (16.5,-12.01);
\draw[gray!55,line width=0.25pt] (28.5,-12.15) -- (28.5,-12.01);
\draw[gray!55,line width=0.25pt] (33.5,-12.15) -- (33.5,-12.01);
\draw[gray!55,line width=0.25pt] (34.5,-12.15) -- (34.5,-11.13);
\draw[gray!55,line width=0.25pt] (35.5,-12.15) -- (35.5,-12.01);
\draw[gray!55,line width=0.25pt] (39.5,-12.15) -- (39.5,-12.01);
\draw[gray!55,line width=0.25pt] (41.5,-12.15) -- (41.5,-12.01);
\draw[gray!55,line width=0.25pt] (53.5,-12.15) -- (53.5,-12.01);
\draw[gray!55,line width=0.25pt] (55.5,-12.15) -- (55.5,-12.01);
\draw[gray!55,line width=0.25pt] (66.5,-12.15) -- (66.5,-12.01);
\draw[gray!55,line width=0.25pt] (71.5,-12.15) -- (71.5,-12.01);
\draw[gray!55,line width=0.25pt] (72.5,-12.15) -- (72.5,-11.13);
\draw[gray!55,line width=0.25pt] (73.5,-12.15) -- (73.5,-12.01);
\draw[gray!55,line width=0.25pt] (77.5,-12.15) -- (77.5,-12.01);
\end{tikzpicture}

%% file: sections/app_limitations.tex
\section{Limitations and Future Directions}
\label{app:limitations}

The deployment findings suggest several directions for extending NavHarness,
from faster local control to memory access and changing environments.

\begin{humanprompted}
\paragraph{Scope of the deployment evidence.}
The continuous-deployment comparison covers finite task sequences in
multiple static simulated houses. It isolates run-end consolidation while
holding map and task-record retention, recovery, and completion checks fixed.
The benchmark interventions separately test map and ledger reuse.
Paired goal sequences need
not produce identical starting poses after earlier trajectories diverge.
The route analysis therefore additionally examines tasks completed by
both runs from nearby starts. The case studies explain recorded decisions
and memory revisions, rather than estimating how frequently every
mechanism would occur in a wider population of homes. Extending the study
to changing environments and longer deployments remains important.
\end{humanprompted}

\subsection{Ownership of the control loop}

Navigation is paced by repeated model decisions. In deployments A--C,
the full system takes 38--45 model turns per task on average.
A turn can issue tool calls containing several primitive actions, so
model turns, tool calls, and movement steps are different quantities.
A stronger and faster embodied policy could instead
own the local rollout while a larger model maintains goals, memory, and task
boundaries. This requires a local controller that can both navigate reliably
and return observations that support subsequent reasoning and verification.
Separating the executor from the harness allows it to be replaced without
changing how persistent state and its sources are recorded.

\subsection{Execution speed and physical time}

Repeated model inference and tool interaction introduce substantial
latency even when actions are batched. Task boundaries, recovery, and memory consolidation are
event-triggered, so their scheduling can also incorporate elapsed time
without changing the archive format.

\subsection{Scope of real-world evaluation}

The evaluations reported here use simulation. Sensor-based SLAM removes
dependence on simulator pose, but physical deployment also requires robust
communication, low-latency obstacle avoidance, and independently measured
trajectories. Collecting and checking a full day's real-world observations
and outcomes remains a substantial evaluation burden.

\subsection{Spatial representation}

The current spatial state uses an ORB-SLAM3 map, named markers, and small
place-photo sets. It cannot represent object motion, temporal change, or the
state of manipulable objects, and its metric frame is never re-anchored to
ground truth, so drift and coordinate misalignment can persist. Moreover,
tracker loop-closure corrections are not propagated into previously
written occupancy cells.
Recognizing a familiar place and recovering its metric pose are distinct
requirements. Place recognition supports reuse of room-level knowledge,
whereas coordinate-based navigation also requires a reliable alignment
with the stored map. Richer backends could provide dense 3D geometry
or change-aware 4D memory, as in SuperMap \citep{supermap}, which tracks
object appearance, disappearance, and relocation. Integrating these
representations would require adapting the map-query and update tools.
The harness specifies which state
to retain, when to update it, and how to record its source without requiring
a particular map representation.

\subsection{Archive access over longer deployments}

The archive operates throughout the multi-tour deployments, with refreshed
house overviews entering later sessions. Consolidation organizes task
experience into house, room, and skill notes with source references.
These overviews form part of the navigator's subsequent input. The
consolidation comparison shows gains beyond retained maps and task records.
Longer deployments make the access policy increasingly
important, including when to retrieve a detailed account and when to revise
an older belief. The current design exposes the house overview and supports
on-demand recall of individual notes. As experience accumulates, retrieval
must remain selective enough to keep relevant information accessible without
overloading the active session. Handling stale or conflicting accounts is
equally important, as the deployment cases show.

\subsection{Static environments and deployment horizon}

Our evaluations use static scanned environments from GOAT-Bench and
IR2R-CE. The extended deployments test memory reuse across finite task
sequences and prescribed relocations, but do not evaluate adaptation to
moved objects, changing layouts, or people. Although agents encounter
incomplete or incorrect records, revising these records does not establish
that the system can detect and track changes in the environment itself.
Evaluating memory revision under environmental change and over longer
deployments remains future work.

\subsection{Capacity of smaller models}
\label{app:small-models}

With Qwen3.8-27B held fixed, the harness raises GOAT task success from
41.4\% to 71.7\%. This supports its value beyond the largest frontier
models, while still requiring an executor capable of multi-round multimodal
navigation. In our tests, Qwen3.5-4B and Qwen3.5-9B struggled with
multi-round multimodal navigation and had low task-completion rates.
Using these models as the common backbone would test a reasoning-core
limitation as well as the harness. We therefore compare harness and
independent-session configurations within each of the four capable
executors. Adapting the interface to smaller models remains future work.

\paragraph{Process records for future training.}
The runs align goals, frames, execution records, claims, checking verdicts,
and memory updates. After independent validation of the fallible judge
labels, these records could support long-horizon post-training or process
supervision. The current study does not update model parameters.

\subsection{Placement of supervision}

Both benchmark evaluations and continuous deployments check the first
stopping request and assess the completed rollout. Other designs could
inspect every fixed number of steps or monitor an executor
continuously. Frequent checks may interrupt useful work and multiply judge
calls, while continuous supervision requires a full-duplex interface and a
low-latency verifier, neither of which the present transports provide. The
choice also interacts with loop ownership: when a local policy owns the
rollout, the boundary at which control returns is the natural place for a
check, and the three schedules become three trigger rules for the same pre-stop verification mechanism.
The verification schedule can be changed independently of the executor
and the rules for retaining memory.

\subsection{Simulation and model-based judgement}

A benchmark scene identifier selects the house archive, while a real robot would need cross-house place
recognition. Finally, the learned judge can hallucinate and shares model-family
biases with the acting policy. Although its claimed-completion accuracy
exceeds accept-all, the four-view checker still accepts false claims on
5.2\% and rejects true claims on 4.1\% of claimed completions
(Appendix~\ref{app:closure-diag}). Goal verification also leaves spatial
and procedural claims in memory unvalidated. Reported navigation scores come
from environment truth. We
map failed or unparsable judgements to \UNRES{}, and avoid treating the
continuous-deployment results as benchmark leaderboard scores. Real-robot operation, place
recognition, and independently trained judges remain necessary tests.

%% file: refs.bib
@misc{alldaynav,
  title = {{AllDayNav}: Lifelong Navigation via Real-World Reinforcement Learning},
  author = {Yin, Hang and Liang, Yinan and Zhang, Jiazhao and Liu, Jiahang and
            Li, Minghan and Zhang, Zhizheng and Wang, He},
  year = {2026},
  eprint = {2606.10927},
  archivePrefix = {arXiv},
  primaryClass = {cs.RO},
  url = {https://arxiv.org/abs/2606.10927}
}

@misc{contextengineering,
  title = {Effective Context Engineering for {AI} Agents},
  author = {Rajasekaran, Prithvi and Dixon, Ethan and Ryan, Carly and Hadfield, Jeremy},
  year = {2025},
  howpublished = {Anthropic Engineering},
  url = {https://www.anthropic.com/engineering/effective-context-engineering-for-ai-agents}
}

@misc{longrunningharness,
  title = {Effective Harnesses for Long-Running Agents},
  author = {Young, Justin},
  year = {2025},
  howpublished = {Anthropic Engineering},
  url = {https://www.anthropic.com/engineering/effective-harnesses-for-long-running-agents}
}

@misc{appharness,
  title = {Harness Design for Long-Running Application Development},
  author = {Rajasekaran, Prithvi},
  year = {2026},
  howpublished = {Anthropic Engineering},
  url = {https://www.anthropic.com/engineering/harness-design-long-running-apps}
}

@inproceedings{openhands,
  title = {{OpenHands}: An Open Platform for {AI} Software Developers as Generalist Agents},
  author = {Wang, Xingyao and Li, Boxuan and Song, Yufan and Xu, Frank F. and Tang, Xiangru and Zhuge, Mingchen and Pan, Jiayi and Song, Yueqi and Li, Bowen and Singh, Jaskirat and Tran, Hoang H. and Li, Fuqiang and Ma, Ren and Zheng, Mingzhang and Qian, Bill and Shao, Yanjun and Muennighoff, Niklas and Zhang, Yizhe and Hui, Binyuan and Lin, Junyang and Brennan, Robert and Peng, Hao and Ji, Heng and Neubig, Graham},
  booktitle = {International Conference on Learning Representations},
  year = {2025},
  url = {https://arxiv.org/abs/2407.16741}
}

@inproceedings{amem,
  title = {{A-MEM}: Agentic Memory for {LLM} Agents},
  author = {Xu, Wujiang and Liang, Zujie and Mei, Kai and Gao, Hang and Tan, Juntao and Zhang, Yongfeng},
  booktitle = {Advances in Neural Information Processing Systems},
  year = {2025},
  url = {https://arxiv.org/abs/2502.12110}
}

@misc{ace,
  title = {Agentic Context Engineering: Evolving Contexts for Self-Improving Language Models},
  author = {Zhang, Qizheng and Hu, Changran and Upasani, Shubhangi and Ma, Boyuan and Hong, Fenglu and Kamanuru, Vamsidhar and Rainton, Jay and Wu, Chen and Ji, Mengmeng and Li, Hanchen and Thakker, Urmish and Zou, James and Olukotun, Kunle},
  year = {2025},
  eprint = {2510.04618},
  archivePrefix = {arXiv},
  url = {https://arxiv.org/abs/2510.04618}
}

@misc{acm,
  title = {{ACM}: Agentic Context Management for Long Horizon Tasks},
  author = {Li, Xiaochuan and Ming, Ryan and Chu, Meng and Shao, Shuai and Jin, Rong and Xiong, Chenyan},
  year = {2026},
  eprint = {2607.23809},
  archivePrefix = {arXiv},
  url = {https://arxiv.org/abs/2607.23809}
}

@inproceedings{r2r,
  title = {Vision-and-Language Navigation: Interpreting Visually-Grounded Navigation Instructions in Real Environments},
  author = {Anderson, Peter and Wu, Qi and Teney, Damien and Bruce, Jake and Johnson, Mark and S{\"u}nderhauf, Niko and Reid, Ian and Gould, Stephen and van den Hengel, Anton},
  booktitle = {Proceedings of the IEEE Conference on Computer Vision and Pattern Recognition},
  pages = {3674--3683},
  year = {2018},
  url = {https://arxiv.org/abs/1711.07280}
}

@misc{harnessvla,
  title = {{Harness VLA}: Steering Frozen {VLAs} into Reliable Manipulation Primitives via Memory-Guided Agents},
  author = {Zhang, Yixian and Zhang, Huanming and Gao, Feng and Li, Xiao and Liu, Zhihao and Zhu, Chunyang and Qiu, Jiaxing and Yan, Yuchen and Liu, Jiyuan and Tang, Wenhao and Fang, Zhengru and Nie, Yi and Wei, Changxu and Wang, Yu and Ding, Wenbo and Yu, Chao},
  year = {2026},
  eprint = {2607.08448},
  archivePrefix = {arXiv},
  url = {https://arxiv.org/abs/2607.08448}
}

@misc{harnessvln,
  title = {{HarnessVLN}: Unifying Training-Free Embodied Navigation through an Agent Harness},
  author = {Chen, Yang and Che, Lirong and Huang, Zhenyu and Fu, Wenbo and Wang, Chuang and Cao, Xu and Liu, Daqi and Yang, Yuzhe and Su, Jian and Guo, Lan-Zhe},
  year = {2026},
  eprint = {2609.15195},
  archivePrefix = {arXiv},
  url = {https://arxiv.org/abs/2609.15195}
}

@misc{showharness,
  title = {{Show-Harness}: Just a {VLM} Agent Can Play Robots},
  author = {Chen, Yanzhe and Bai, Zechen and Cao, Zhijun and Zeng, Wenzheng and Lin, Kevin Qinghong and Lin, Yiqi and Liang, Guoqiang and Ma, Kevin Yuchen and Huang, Qiming and Shou, Mike Zheng},
  year = {2026},
  eprint = {2609.10522},
  archivePrefix = {arXiv},
  url = {https://arxiv.org/abs/2609.10522}
}

@misc{zetta,
  title = {{Zetta} $\zeta$: An Efficient Closed-Loop Embodied Harness for Self-Evolving Physical Intelligence},
  author = {Ding, Xin and Mi, Liang and Huang, Mingzhe and Wang, Zixuan and Zhang, Chao and Hao, Zixu and Chen, Fu and Li, Xiangyu and Zheng, Yikai and Guo, Yaoyu and Wang, Weijun and Li, Kun and Wu, Hao and Liu, Yunxin and Cao, Ting},
  year = {2026},
  eprint = {2608.16590},
  archivePrefix = {arXiv},
  url = {https://arxiv.org/abs/2608.16590}
}

@misc{mip,
  title = {Embodied Agents Take Control: Minimal-Interface Zero-Shot Agents Rival Industrial-Scale Policies in Vision-and-Language Navigation},
  author = {Zhou, Jian and Zhao, Xunyi and Zhou, Gengze and Li, Zerui and Lin, Sihao and Liu, Jiajun and Wu, Qi},
  year = {2026},
  eprint = {2607.26148},
  archivePrefix = {arXiv},
  url = {https://arxiv.org/abs/2607.26148}
}

@article{orbslam3,
  title = {{ORB-SLAM3}: An Accurate Open-Source Library for Visual, Visual-Inertial, and Multimap {SLAM}},
  author = {Campos, Carlos and Elvira, Richard and G{\'o}mez Rodr{\'i}guez, Juan J. and Montiel, Jos{\'e} M. M. and Tard{\'o}s, Juan D.},
  journal = {IEEE Transactions on Robotics},
  volume = {37}, number = {6}, pages = {1874--1890}, year = {2021},
  doi = {10.1109/TRO.2021.3075644}
}

@misc{agentharness,
  title        = {Agent Harness for Large Language Model Agents: A Survey},
  author       = {Meng, Qianyu and Wang, Yanan and Chen, Liyi and Li, Yihang and Wu, Wei and Jiang, Wenyuan and Wang, Qimeng and Lu, Chengqiang and Gao, Yan and Wu, Yi and Hu, Yao},
  year         = {2026},
  howpublished = {Preprints.org},
  doi          = {10.20944/preprints202604.0428.v3},
  url          = {https://doi.org/10.20944/preprints202604.0428.v3},
  note         = {Version 3}
}

@article{voyager,
  title     = {Voyager: An Open-Ended Embodied Agent with Large Language Models},
  author    = {Wang, Guanzhi and Xie, Yuqi and Jiang, Yunfan and Mandlekar, Ajay and Xiao, Chaowei and Zhu, Yuke and Fan, Linxi and Anandkumar, Anima},
  journal   = {Transactions on Machine Learning Research},
  year      = {2024},
  url       = {https://openreview.net/forum?id=pAMNKGwja6}
}

@inproceedings{ivln,
  title     = {Iterative Vision-and-Language Navigation},
  author    = {Krantz, Jacob and Banerjee, Shurjo and Zhu, Wang and Corso, Jason
               and Anderson, Peter and Lee, Stefan and Thomason, Jesse},
  booktitle = {Proceedings of the IEEE/CVF Conference on Computer Vision and
               Pattern Recognition (CVPR)},
  pages     = {14921--14930},
  year      = {2023},
  url = {https://openaccess.thecvf.com/content/CVPR2023/html/Krantz_Iterative_Vision-and-Language_Navigation_CVPR_2023_paper.html}
}

@inproceedings{goatbench,
  title     = {{GOAT-Bench}: A Benchmark for Multi-Modal Lifelong Navigation},
  author    = {Khanna, Mukul and Ramrakhya, Ram and Chhablani, Gunjan and
               Yenamandra, Sriram and Gervet, Theophile and Chang, Matthew and
               Kira, Zsolt and Chaplot, Devendra Singh and Batra, Dhruv and
               Mottaghi, Roozbeh},
  booktitle = {Proceedings of the IEEE/CVF Conference on Computer Vision and
               Pattern Recognition (CVPR)},
  year      = {2024},
  url = {https://openaccess.thecvf.com/content/CVPR2024/html/Khanna_GOAT-Bench_A_Benchmark_for_Multi-Modal_Lifelong_Navigation_CVPR_2024_paper.html}
}

@misc{navmcp,
  title         = {Scaffolding Foundation Models into Physical-World Agents
                   Pushes the Frontier of Long-Horizon Navigation},
  author        = {Lei, Zixing and Zhou, Gengze and Chen, Xiong-Hui and
                   Zhang, Jiazhao and Huang, Yiyang and Yin, Hang and
                   Yuan, Haoqi and Wu, Qi and Li, Weixin and Chen, Siheng},
  year          = {2026},
  eprint        = {2608.30396},
  archivePrefix = {arXiv},
  primaryClass  = {cs.AI},
  url = {https://arxiv.org/abs/2608.30396}
}

@misc{agenticnav,
  title         = {{AgenticNav}: Zero-Shot Vision-and-Language Navigation as a
                   Tool-Calling Harness},
  author        = {Li, Yijian and Li, Changze and Shi, Hantian and Luo, Jiaying
                   and Cai, Jiyuan and Yang, Ming and Qin, Tong},
  year          = {2026},
  eprint        = {2606.10577},
  archivePrefix = {arXiv},
  primaryClass  = {cs.RO},
  url = {https://arxiv.org/abs/2606.10577}
}

@misc{spacevln,
  title         = {{SpaceVLN}: A Zero-Shot Vision-and-Language Navigation Agent
                   with Online Spatial Cognitive Memory and Reasoning},
  author        = {Deng, Yucheng and Lai, Pingrui and Li, Xinhai and Bai, Chenjia
                   and Deng, Xiaoheng and Sun, Chengnuo and Li, Xuelong and
                   Yang, Hua},
  year          = {2026},
  eprint        = {2606.08992},
  archivePrefix = {arXiv},
  primaryClass  = {cs.RO},
  url = {https://arxiv.org/abs/2606.08992}
}

@misc{anygoal,
  title         = {{AnyGoal}: Vision-Language Guided Multi-Agent Exploration for
                   Training-Free Lifelong Navigation},
  author        = {James, MoniJesu and Fernando, Marcelino Julio and
                   Cabrera, Miguel Altamirano and Tsetserukou, Dzmitry},
  year          = {2026},
  eprint        = {2606.13878},
  archivePrefix = {arXiv},
  primaryClass  = {cs.RO},
  url = {https://arxiv.org/abs/2606.13878}
}

@misc{ssmgnav,
  title         = {{SSMG-Nav}: Enhancing Lifelong Object Navigation with
                   Semantic Skeleton Memory Graph},
  author        = {Niu, Haochen and Zhang, Lantao and Ji, Xingwu and
                   Ying, Rendong and Liu, Peilin and Wen, Fei},
  year          = {2026},
  eprint        = {2603.01813},
  archivePrefix = {arXiv},
  primaryClass  = {cs.RO},
  url = {https://arxiv.org/abs/2603.01813}
}

@misc{gsmem,
  title         = {{GSMem}: {3D} Gaussian Splatting as Persistent Spatial Memory
                   for Zero-Shot Embodied Exploration and Reasoning},
  author        = {Lu, Yiren and Du, Yi and Liu, Disheng and Zhou, Yunlai and
                   Wang, Chen and Yin, Yu},
  year          = {2026},
  eprint        = {2603.19137},
  archivePrefix = {arXiv},
  primaryClass  = {cs.CV},
  url = {https://arxiv.org/abs/2603.19137}
}

@misc{lmee,
  title         = {Explore with Long-term Memory: A Benchmark and Multimodal
                   {LLM}-based Reinforcement Learning Framework for Embodied
                   Exploration},
  author        = {Wang, Sen and Liu, Bangwei and Gao, Zhenkun and Ma, Lizhuang
                   and Wang, Xuhong and Xie, Yuan and Tan, Xin},
  year          = {2026},
  eprint        = {2601.10744},
  archivePrefix = {arXiv},
  primaryClass  = {cs.CV},
  note          = {CVPR 2026},
  url = {https://arxiv.org/abs/2601.10744}
}

@inproceedings{uniwalker,
  title     = {Lifelong Embodied Navigation Learning},
  author    = {Wang, Xudong and Dong, Jiahua and Liu, Baichen and Lyu, Qi and
               Liu, Lianqing and Han, Zhi},
  booktitle = {International Conference on Learning Representations (ICLR)},
  year      = {2026},
  url = {https://openreview.net/forum?id=PaYo96rjij}
}

@misc{memgpt,
  title         = {{MemGPT}: Towards {LLM}s as Operating Systems},
  author        = {Packer, Charles and Wooders, Sarah and Lin, Kevin and
                   Fang, Vivian and Patil, Shishir G. and Stoica, Ion and
                   Gonzalez, Joseph E.},
  year          = {2023},
  eprint        = {2310.08560},
  archivePrefix = {arXiv},
  url = {https://arxiv.org/abs/2310.08560}
}

@article{coala,
  title = {Cognitive Architectures for Language Agents},
  author = {Sumers, Theodore R. and Yao, Shunyu and Narasimhan, Karthik and Griffiths, Thomas L.},
  journal = {Transactions on Machine Learning Research},
  year = {2024},
  url = {https://arxiv.org/abs/2309.02427}
}

@misc{roboharness,
  title = {{RoboHarness}: Memory-Driven Orchestration of Heterogeneous Robot Policies for Long-Horizon Planning},
  author = {Huang, Jinbang and Hu, Yuanzhao and Li, Zhiyuan and Qi, Ran and Xiao, Yixin and Zhang, Zhanguang and Coates, Mark and Cao, Tongtong and Zhang, Yingxue},
  year = {2026},
  eprint = {2607.18060},
  archivePrefix = {arXiv},
  url = {https://arxiv.org/abs/2607.18060}
}

@inproceedings{sweagent,
  title     = {{SWE-agent}: Agent-Computer Interfaces Enable Automated Software
               Engineering},
  author    = {Yang, John and Jimenez, Carlos E. and Wettig, Alexander and
               Lieret, Kilian and Yao, Shunyu and Narasimhan, Karthik and Press, Ofir},
  booktitle = {Advances in Neural Information Processing Systems},
  year      = {2024},
  url = {https://arxiv.org/abs/2405.15793}
}

@misc{reflexion,
  title         = {Reflexion: Language Agents with Verbal Reinforcement Learning},
  author        = {Shinn, Noah and Cassano, Federico and Berman, Edward and
                   Gopinath, Ashwin and Narasimhan, Karthik and Yao, Shunyu},
  year          = {2023},
  eprint        = {2303.11366},
  archivePrefix = {arXiv},
  primaryClass  = {cs.AI},
  url = {https://arxiv.org/abs/2303.11366}
}

@misc{generativeagents,
  title         = {Generative Agents: Interactive Simulacra of Human Behavior},
  author        = {Park, Joon Sung and O'Brien, Joseph C. and Cai, Carrie J. and
                   Morris, Meredith Ringel and Liang, Percy and Bernstein, Michael S.},
  year          = {2023},
  eprint        = {2304.03442},
  archivePrefix = {arXiv},
  primaryClass  = {cs.HC},
  url = {https://arxiv.org/abs/2304.03442}
}

@misc{shaper,
  title = {Self-Evolving Embodied Agents via Skill-Harness Evolution},
  author = {Wang, Peidong and Ma, Zhiming and Chang, Ying and Luo, Xufang and Zhang, Yiqun and Wang, Zihan and Yang, Xiaocui and Feng, Shi and Yang, Yuqing and Li, Dongsheng},
  year = {2026},
  eprint = {2608.11350},
  archivePrefix = {arXiv},
  url = {https://arxiv.org/abs/2608.11350}
}

@misc{vlmnav,
  title         = {End-to-End Navigation with Vision Language Models: Transforming Spatial Reasoning into Question-Answering},
  author        = {Goetting, Dylan and Singh, Himanshu Gaurav and Loquercio, Antonio},
  year          = {2024},
  eprint        = {2411.05755},
  archivePrefix = {arXiv},
  primaryClass  = {cs.RO},
  url = {https://arxiv.org/abs/2411.05755}
}

@misc{dynavlm,
  title         = {{DyNaVLM}: Zero-Shot Vision-Language Navigation System with Dynamic Viewpoints and Self-Refining Graph Memory},
  author        = {Ji, Zihe and Lin, Huangxuan and Gao, Yue},
  year          = {2025},
  eprint        = {2506.15096},
  archivePrefix = {arXiv},
  primaryClass  = {cs.RO},
  url = {https://arxiv.org/abs/2506.15096}
}

@inproceedings{tango,
  title     = {{TANGO}: Training-free Embodied {AI} Agents for Open-world Tasks},
  author    = {Ziliotto, Filippo and Campari, Tommaso and Serafini, Luciano and Ballan, Lamberto},
  booktitle = {Proceedings of the IEEE/CVF Conference on Computer Vision and Pattern Recognition (CVPR)},
  year      = {2025}
}

@inproceedings{mtu3d,
  title     = {Move to Understand a {3D} Scene: Bridging Visual Grounding and Exploration for Efficient and Versatile Embodied Navigation},
  author    = {Zhu, Ziyu and Wang, Xilin and Li, Yixuan and Zhang, Zhuofan and Ma, Xiaojian and Chen, Yixin and Jia, Baoxiong and Liang, Wei and Yu, Qian and Deng, Zhidong and Huang, Siyuan and Li, Qing},
  booktitle = {Proceedings of the IEEE/CVF International Conference on Computer Vision (ICCV)},
  year      = {2025}
}

@inproceedings{threedmem,
  title     = {{3D-Mem}: {3D} Scene Memory for Embodied Exploration and Reasoning},
  author    = {Yang, Yuncong and Yang, Han and Zhou, Jiachen and Chen, Peihao and Zhang, Hongxin and Du, Yilun and Gan, Chuang},
  booktitle = {Proceedings of the IEEE/CVF Conference on Computer Vision and Pattern Recognition (CVPR)},
  year      = {2025}
}

@inproceedings{cow,
  title     = {{CoWs} on Pasture: Baselines and Benchmarks for Language-Driven Zero-Shot Object Navigation},
  author    = {Gadre, Samir Yitzhak and Wortsman, Mitchell and Ilharco, Gabriel and Schmidt, Ludwig and Song, Shuran},
  booktitle = {Proceedings of the IEEE/CVF Conference on Computer Vision and Pattern Recognition (CVPR)},
  year      = {2023}
}

@misc{agentsystemsurvey,
  title         = {From Question Answering to Task Completion: A Survey on Agent System and Harness Design},
  author        = {Guo, Jianyuan and Hao, Zhiwei and Wang, Chengcheng and Fan, Cheng and Luo, Tingzhang and Li, Hongguang and Gao, Ying and Mei, Hefei and Peng, Jiankun and Xu, Rongjian and Dong, Minjing and Wu, Han and Zheng, Mengyu and Han, Kai and Wang, Shiqi and Xu, Chang and Wang, Yunhe},
  year          = {2026},
  eprint        = {2606.20683},
  archivePrefix = {arXiv},
  primaryClass  = {cs.AI},
  url = {https://arxiv.org/abs/2606.20683}
}

@inproceedings{acon,
  title     = {{ACON}: Optimizing Context Compression for Long-horizon {LLM} Agents},
  author    = {Kang, Minki and Chen, Wei-Ning and Han, Dongge and Inan, Huseyin A. and Wutschitz, Lukas and Chen, Yanzhi and Sim, Robert and Rajmohan, Saravan},
  booktitle = {International Conference on Machine Learning (ICML)},
  year      = {2026},
  url = {https://arxiv.org/abs/2510.00615}
}

@misc{beyondcompaction,
  title         = {Beyond Compaction: Structured Context Eviction for Long-Horizon Agents},
  author        = {Semenov, Andrew and Dorofeev, Svyatoslav},
  year          = {2026},
  eprint        = {2606.11213},
  archivePrefix = {arXiv},
  primaryClass  = {cs.AI},
  url = {https://arxiv.org/abs/2606.11213}
}

@misc{selfgc,
  title         = {{Self-GC}: Self-Governing Context for Long-Horizon {LLM} Agents},
  author        = {Hao, Xubin and Meng, Hongjin and Yin, Xin and Zhu, Jiawei and Cao, Chenpeng},
  year          = {2026},
  eprint        = {2607.00692},
  archivePrefix = {arXiv},
  primaryClass  = {cs.AI},
  url = {https://arxiv.org/abs/2607.00692}
}

@inproceedings{habitat,
  title     = {Habitat: A Platform for Embodied {AI} Research},
  author    = {Savva, Manolis and Kadian, Abhishek and Maksymets, Oleksandr and Zhao, Yili and Wijmans, Erik and Jain, Bhavana and Straub, Julian and Liu, Jia and Koltun, Vladlen and Malik, Jitendra and Parikh, Devi and Batra, Dhruv},
  booktitle = {Proceedings of the IEEE/CVF International Conference on Computer Vision (ICCV)},
  year      = {2019}
}

@inproceedings{hm3d,
  title     = {Habitat-Matterport {3D} Dataset ({HM3D}): 1000 Large-scale {3D} Environments for Embodied {AI}},
  author    = {Ramakrishnan, Santhosh Kumar and Gokaslan, Aaron and Wijmans, Erik and Maksymets, Oleksandr and Clegg, Alexander and Turner, John and Undersander, Eric and Galuba, Wojciech and Westbury, Andrew and Chang, Angel X. and Savva, Manolis and Zhao, Yili and Batra, Dhruv},
  booktitle = {NeurIPS Datasets and Benchmarks Track},
  year      = {2021}
}

@inproceedings{mp3d,
  title     = {{Matterport3D}: Learning from {RGB-D} Data in Indoor Environments},
  author    = {Chang, Angel and Dai, Angela and Funkhouser, Thomas and Halber, Maciej and Niessner, Matthias and Savva, Manolis and Song, Shuran and Zeng, Andy and Zhang, Yinda},
  booktitle = {International Conference on 3D Vision (3DV)},
  year      = {2017}
}

@inproceedings{vlnce,
  title     = {Beyond the Nav-Graph: Vision-and-Language Navigation in Continuous Environments},
  author    = {Krantz, Jacob and Wijmans, Erik and Majumdar, Arjun and Batra, Dhruv and Lee, Stefan},
  booktitle = {European Conference on Computer Vision (ECCV)},
  year      = {2020}
}

@misc{spl,
  title         = {On Evaluation of Embodied Navigation Agents},
  author        = {Anderson, Peter and Chang, Angel and Chaplot, Devendra Singh and Dosovitskiy, Alexey and Gupta, Saurabh and Koltun, Vladlen and Kosecka, Jana and Malik, Jitendra and Mottaghi, Roozbeh and Savva, Manolis and Zamir, Amir R.},
  year          = {2018},
  eprint        = {1807.06757},
  archivePrefix = {arXiv},
  primaryClass  = {cs.AI},
  url = {https://arxiv.org/abs/1807.06757}
}

@misc{ndtw,
  title         = {General Evaluation for Instruction Conditioned Navigation using Dynamic Time Warping},
  author        = {Ilharco, Gabriel and Jain, Vihan and Ku, Alexander and Ie, Eugene and Baldridge, Jason},
  year          = {2019},
  eprint        = {1907.05446},
  archivePrefix = {arXiv},
  primaryClass  = {cs.AI},
  url = {https://arxiv.org/abs/1907.05446}
}

@misc{claudeopus5,
  title        = {Claude {Opus} 5 System Card},
  author       = {{Anthropic}},
  year         = {2026},
  howpublished = {\url{https://www.anthropic.com/transparency}}
}

@misc{qwen38,
  title        = {{Qwen3.8-27B} Model Card},
  author       = {{Qwen Team}},
  year         = {2026},
  howpublished = {\url{https://huggingface.co/Qwen/Qwen3.8-27B}}
}

@misc{gpt6astra,
  title        = {{GPT-6 Astra}: A New Generation of Intelligence},
  author       = {{OpenAI}},
  year         = {2026},
  howpublished = {\url{https://openai.com/index/gpt-6-astra/}}
}

@inproceedings{overnav,
  title = {{OVER-NAV}: Elevating Iterative Vision-and-Language Navigation with Open-Vocabulary Detection and StructurEd Representation},
  author = {Zhao, Ganlong and Li, Guanbin and Chen, Weikai and Yu, Yizhou},
  booktitle = {Proceedings of the IEEE/CVF Conference on Computer Vision and Pattern Recognition},
  year = {2024},
  url = {https://arxiv.org/abs/2403.17334}
}

@inproceedings{seqwalker,
  title = {{SeqWalker}: Sequential-Horizon Vision-and-Language Navigation with Hierarchical Planning},
  author = {Han, Zebin and Wang, Xudong and Liu, Baichen and Lyu, Qi and Shang, Zhenduo and Dong, Jiahua and Liu, Lianqing and Han, Zhi},
  booktitle = {Proceedings of the AAAI Conference on Artificial Intelligence},
  year = {2026},
  url = {https://arxiv.org/abs/2601.04699}
}

@misc{embodiedharness,
  title         = {Towards the Harness of Embodied Agents},
  author        = {Wang, Qi and Wang, Tianyi and Li, Chengyang and Ban, Shikun and Chen, Yurun and Ge, Yizhong and Qin, Jason and Li, Chengtai and Zhu, Wentao},
  year          = {2026},
  eprint        = {2608.11246},
  archivePrefix = {arXiv},
  primaryClass  = {cs.RO},
  url = {https://arxiv.org/abs/2608.11246}
}

@misc{rho,
  title         = {{RHO}: Your Coding Agent is Secretly a Roboticist},
  author        = {Elmaaroufi, Karim and Svegliato, Justin and Kalade, Sarunas and Schelle, Graham and Seshia, Sanjit A. and Zaharia, Matei},
  year          = {2026},
  eprint        = {2606.16458},
  archivePrefix = {arXiv},
  primaryClass  = {cs.RO},
  url = {https://arxiv.org/abs/2606.16458}
}

@misc{reexplore,
  title         = {{ReEXplore}: Improving {MLLMs} for Embodied Exploration with Contextualized Retrospective Experience Replay},
  author        = {Zhang, Gengyuan and Ding, Mingcong and Wu, Jingpei and Liao, Ruotong and Tresp, Volker},
  year          = {2025},
  eprint        = {2511.19033},
  archivePrefix = {arXiv},
  primaryClass  = {cs.CV},
  url = {https://arxiv.org/abs/2511.19033}
}

@misc{himm,
  title         = {{HIMM}: Human-Inspired Long-Term Memory Modeling for Embodied Exploration and Question Answering},
  author        = {Li, Ji and Wang, Bo and Xia, Jing and Li, Mingyi and Hu, Shiyan},
  year          = {2026},
  eprint        = {2602.15513},
  archivePrefix = {arXiv},
  primaryClass  = {cs.CV},
  url = {https://arxiv.org/abs/2602.15513}
}

@misc{metanav,
  title         = {Stop Wandering: Efficient Vision-Language Navigation via Metacognitive Reasoning},
  author        = {Li, Xueying and Lyu, Feng and Wu, Hao and Liu, Mingliu and Liu, Jia-Nan and Liu, Guozi},
  year          = {2026},
  eprint        = {2604.02318},
  archivePrefix = {arXiv},
  primaryClass  = {cs.CV},
  url = {https://arxiv.org/abs/2604.02318}
}

@misc{hgr,
  title         = {Hypothesis Graph Refinement: Hypothesis-Driven Exploration with Cascade Error Correction for Embodied Navigation},
  author        = {Chen, Peixin and Zhang, Guoxi and Ma, Jianwei and Li, Qing},
  year          = {2026},
  eprint        = {2604.04108},
  archivePrefix = {arXiv},
  primaryClass  = {cs.CV},
  url = {https://arxiv.org/abs/2604.04108}
}

@misc{obsgraph,
  title         = {{ObsGraph}: Hierarchical Observation Representation for Embodied Reasoning and Exploration},
  author        = {Lee, Taekbeom and Jang, Youngseok and Heo, Jeonghwa and Choi, Jeongjun and Kim, H. Jin},
  year          = {2026},
  eprint        = {2606.24068},
  archivePrefix = {arXiv},
  primaryClass  = {cs.CV},
  url = {https://arxiv.org/abs/2606.24068}
}

@misc{evomemnav,
  title         = {{EvoMemNav}: Efficient Self-Evolving Fine-Grained Memory for Zero-Shot Embodied Navigation},
  author        = {Ge, Zuhao and Jia, Xiaosong and Wu, Chao and Zhou, Yuchen and Wu, Zuxuan and Jiang, Yu-Gang},
  year          = {2026},
  eprint        = {2606.03509},
  archivePrefix = {arXiv},
  primaryClass  = {cs.CV},
  url = {https://arxiv.org/abs/2606.03509}
}

@misc{astranavmem,
  title         = {{AstraNav-Memory}: Contexts Compression for Long Memory},
  author        = {Ren, Botao and Hu, Junjun and Xue, Xinda and Luo, Minghua and Chen, Jintao and Bai, Haochen and You, Liangliang and Xu, Mu},
  year          = {2025},
  eprint        = {2512.21627},
  archivePrefix = {arXiv},
  primaryClass  = {cs.CV},
  url = {https://arxiv.org/abs/2512.21627}
}

@inproceedings{supermap,
  title = {{SuperMap}: A Spatio-Temporal {SLAM} System for Visual-Language Navigation},
  author = {Zhao, Shibo and Chen, Guofei and Zhu, Honghao and Li, Zhiheng and Yao, Changwei and Zantout, Nader and Kim, Seungchan and Wang, Wenshan and Zhang, Ji and Scherer, Sebastian},
  booktitle = {Proceedings of Robotics: Science and Systems (RSS)},
  year = {2026},
  url = {https://superodometry.com/supermap}
}

@misc{gpt4o,
  title = {{GPT-4o} System Card},
  author = {{OpenAI}},
  year = {2024},
  url = {https://openai.com/index/gpt-4o-system-card/}
}

@misc{stegnav,
  title         = {{STEGNav}: Spatio-Temporal Event Graph Reasoning for Multimodal Lifelong Object Navigation},
  author        = {Chen, Yang and Huang, Zhenyu and Fu, Wenbo and Peng, Danyang and Tian, Shi-Yu and Yu, Kun-Yang and Guo, Lan-Zhe},
  year          = {2026},
  eprint        = {2608.28279},
  archivePrefix = {arXiv},
  primaryClass  = {cs.CV},
  url = {https://arxiv.org/abs/2608.28279}
}
